\documentclass[lettersize,journal]{IEEEtran}
\usepackage{amsmath,amsfonts}
\usepackage{algorithmic}
\usepackage{algorithm}
\usepackage{array}
\usepackage[caption=false,font=normalsize,labelfont=sf,textfont=sf]{subfig}
\usepackage{textcomp}
\usepackage{stfloats}
\usepackage{url}
\usepackage{bm}
\usepackage{multirow}
\usepackage{verbatim}
\usepackage{graphicx}
\usepackage{color}
\usepackage{cite}
\usepackage{colortbl}
\usepackage[table,xcdraw]{xcolor}
\begin{document}
	
	\title{Searching a Compact Architecture for  Robust Multi-Exposure Image Fusion }
	
	\author{Zhu Liu, Jinyuan Liu, Guanyao Wu, Zihang Chen, Xin Fan, \IEEEmembership{Senior Member, IEEE}, Risheng Liu,  \IEEEmembership{Member, IEEE}
		% <-this % stops a space
		\thanks{This work is partially supported by the National Key R\&D Program of China (No. 2022YFA1004101), the National Natural Science Foundation of China (No. U22B2052).}
		\thanks{Zhu Liu  is with the School of Software Technology, Dalian University of
			Technology, Dalian, 116024, China. (e-mail: liuzhu@mail.dlut.edu.cn,).}
		\thanks{Jinyuan Liu is with the School of Mechanical Engineering, Dalian University of
			Technology, Dalian, 116024, China. (e-mail: atlantis918@hotmail.com).}
		\thanks{Guanyao Wu  is with the School of Software Technology, Dalian University of
			Technology, Dalian, 116024, China. (e-mail:  rollingplainko@gmail.com).}
		\thanks{Zihang Chen  is with the School of Software Technology, Dalian University of
			Technology, Dalian, 116024, China. (e-mail:  chenzi\_hang@mail.dlut.edu.cn).}
		\thanks{ Xin Fan is with School of Software Technology, Dalian University of Technology, Dalian, 116024,  China.	(e-mail: xin.fan@dlut.edu.cn).}
		
		\thanks{ Risheng. Liu is  with the School of Software Technology, Dalian University of Technology, Dalian, 116024, China.
			(Corresponding author, e-mail: rsliu@dlut.edu.cn).}
	}
	% <-this % stops a space
	%\thanks{Manuscript received April 19, 2021; revised August 16, 2021.}}

\markboth{Journal of \LaTeX\ Class Files,~Vol.~14, No.~8, August~2021}%
{Shell \MakeLowercase{\textit{et al.}}: A DASMple Article Using IEEEtran.cls for IEEE Journals}


\maketitle

\begin{abstract}
In recent years, learning-based methods have achieved significant advancements in multi-exposure image fusion. However, two major stumbling blocks hinder the development, including pixel misalignment and inefficient inference. Reliance on aligned image pairs in existing methods causes susceptibility to artifacts due to device motion. Additionally, existing techniques often rely on handcrafted architectures with huge network engineering, resulting in redundant parameters, adversely impacting inference efficiency and flexibility. To mitigate these limitations, this study introduces an architecture search-based paradigm incorporating self-alignment and detail repletion modules for robust multi-exposure image fusion.
		Specifically, targeting the extreme discrepancy of exposure, we propose the self-alignment module, leveraging scene relighting to constrain the illumination degree for following alignment and feature extraction. Detail repletion is proposed to enhance the texture details of scenes. Additionally, incorporating a hardware-sensitive constraint, we present the fusion-oriented architecture search to explore compact and efficient networks for fusion. The proposed method outperforms various competitive schemes, achieving a noteworthy 3.19\% improvement in PSNR for general scenarios and an impressive 23.5\%  enhancement in misaligned scenarios. Moreover, it significantly reduces inference time by 69.1\%. The  code will be available at \url{https://github.com/LiuZhu-CV/CRMEF}.

\end{abstract}


\section{Introduction}
High Dynamic Range Imaging (HDRI)~\cite{wu2022ace,wang2021deep,wu2022dmef}, encompassing comprehensive scene content with optimal exposure, has gained considerable attention in recent years. As a pivotal technique for computer vision, HDRI not only delivers visually appealing observations in harmony with the human visual system, but also integrates essential features for a range of downstream vision applications, including object detection ~\cite{liu2023bi,liu2022target,liu2023task}, visual enhancement~\cite{liu2021retinex,liang2021recurrent,zhang2023iid,liu2023optimization} and semantic segmentation~\cite{ma2022toward,liu2023multi,liu2023paif}. 
Unfortunately, due to constraints in photography equipment (\emph{e.g.,} smartphones and single-lens reflex cameras) that capture images with limited dynamic ranges, these images experience varying degrees of luminance degradation, leading to corrupted over/under-exposed regions. As a result, Low Dynamic Range (LDR) images are plagued by color distortion and loss of detail, hindering the accurate portrayal of complete natural scenes. Consequently, the generation of well-exposed HDR images remains both a challenging and significant research topic.

\begin{figure}[t]
	\centering
	\begin{tabular}{c@{\extracolsep{0.1em}}c} 
		\includegraphics[width=0.485\textwidth]{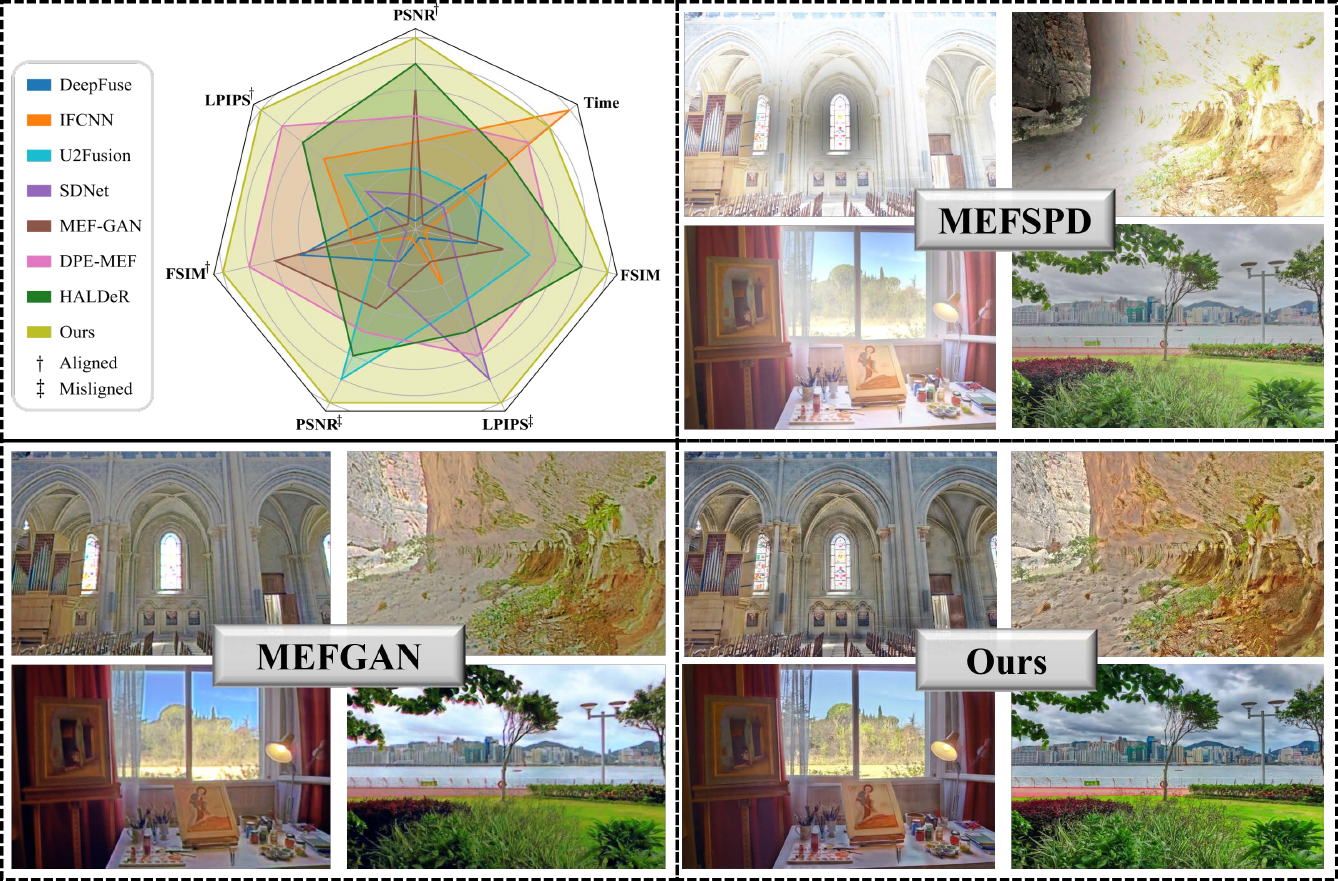}\\
	\end{tabular}
	
	\caption{Visual comparison with representative MEF methods on general and misaligned fusion scenarios. The left figure plots the ranking with these competitors under general and misaligned scenarios. Other figures shows the obvious comparison with patch-based scheme MEFSPD~\cite{li2021detail} and learning-based method MEFGAN~\cite{xu2020mef}. }
	\label{fig:fristfig}
	
\end{figure}

Recently, a growing number of researchers have endeavored to create cutting-edge HDRI hardware devices capable of producing an extensive range of illumination, thereby addressing the limitations inherent to traditional digital cameras~\cite{debevec2008recovering,munkberg2006high,nayar2000high}. Nevertheless, due to elevated production costs and suboptimal efficiency, these intricately designed devices face challenges in achieving widespread adoption in real-world applications. As an alternative, Multi-Exposure Fusion (MEF) offers an efficacious solution for generating HDR images by assimilating characteristic texture information from a collection of LDR images captured under diverse exposures. This strategy adeptly bypasses hardware-specific limitations while maintaining a lower computational cost. Within the existing literature, conventional frameworks~\cite{li2013image,wang2019detail,hayat2019ghost,hu2017exploiting} and learning-based ones~\cite{ram2017deepfuse,liu2022attention,li2022learning} constitute the predominant categories of MEF techniques. Despite these advancements, MEF continues to grapple with certain hurdles that impede its overall efficacy.

In essence, there is an urgent demand for an all-encompassing, robust, and efficient learning approach that not only delivers promising visual realism enhancement but also ensures high efficiency and stability across a wide array of scenes. It is imperative to highlight that current learning-based methods neglect the essential adaptive preservation in multi-exposure image fusion. Specifically, a variety of approaches utilize direct fusion rules for feature aggregation, such as summation~\cite{li2013image} and multiplication~\cite{liu2022attention}. Regrettably, these rudimentary fusion rules fall short in effectively aggregating information from LDR pairs with markedly distinct characteristics, thereby failing to preserve critical information (\emph{e.g.}, pixel intensity and texture details) appropriately. Furthermore, given the considerable variation in multi-exposure image distributions, manually designed architectures face difficulties in flexibly adapting to disparate data distributions. In addition, due to the unavoidable movements and shaking of imaging devices, minor pixel misalignments in LDR pairs are commonplace. Existing methods seldom tackle this issue, leading to fused images characterized by blurred details and compromised structure. 

To be more specific, the second challenge stems from the computational efficiency of existing methods. Present MEF methods, encompassing both conventional and learning-based approaches, rely heavily on handcrafted architectures and operations. In terms of conventional frameworks, various transformations such as wavelet transform~\cite{zhang2018multi}, multi-scale representation~\cite{yang2022multi}, and Laplacian pyramid~\cite{li2011performance} are proposed to enable feature fusion through handcrafted mechanisms. Unfortunately, these manual designs for feature extraction and fusion rules demand substantial fine-tuning and a significant amount of experiential knowledge. Most of these methods utilize numerical optimization, which in turn impacts the inference efficiency and robustness in real-world applications. Furthermore, in recent years, the powerful feature extraction capabilities of CNN-based learning have led to the increasing dominance of end-to-end models in MEF development, considerably improving performance concerning statistical metrics and visual effects. For architectural construction, a variety of learnable mechanisms~\cite{xu2020u2fusion,li2021multiple,han2022multi} have been introduced to forge connections between LDR pairs and HDR outputs. We contend that current neural architectures for MEF largely borrow effective practices from other vision tasks without paying adequate attention to MEF-specific characteristics. As a consequence, these simplistic cascaded architectures, marked by increased width and depth, possess an excessive number of parameters, making them prone to feature redundancy.

\subsection{Contributions}
\textcolor{black}{ It can be observed that current learning-based MEF methods suffer from inflexible detail preservation, especially on the misalignment scenarios and computational efficiency. To partially overcome these limitations, we propose a comprehensive architecture search-based approach for multi-exposure fusion.}

\textcolor{black}{To be specific, we first develop a MEF-oriented hyper-architecture, adhering to two primary principles for robust fusion: self-alignment and detail repletion. Initially, we introduce a scene-relighting technique designed to harmonize illumination among source images, effectively enhancing over/under-exposed details and facilitating improved feature aggregation. Complementing this, our method integrates deformable alignment, ensuring precise feature registration to minimize blurring artifacts in the fused images. Subsequently, we propose the detail repletion module to refine the coarse fusion results, leading to richer texture details. Next, we make the first attempt to investigate the automatic compact architecture design for the MEF task. To contend with hardware latency constraints, we leverage differentiable architecture search, facilitating the automatic discovery of a streamlined and efficient model tailored for image fusion tasks.
	As a result, our method consistently delivers vivid colors, abundant details, and ghosting-free results, as visually demonstrated in Fig.~\ref{fig:fristfig}. In summary, our primary contributions can be outlined as follows:}

\begin{itemize}
	\item \textcolor{black}{ Tackling pixel misalignment and detail enhancement as critical components of multi-exposure image fusion, we propose a comprehensive paradigm  that combines robust registration and detail repletion to  preserve texture, to address diverse  scenarios  while ensuring high efficiency.}

	\item \textcolor{black}{We present  a hardware-sensitive,
		architecture search-based framework to realize
		the effective and fast inference.
		To our knowledge, it is the first time to investigate the automatic light-weight architecture construction  specifically tailored for multi-exposure image fusion.}
	
	\item \textcolor{black}{By applying principles of scene relighting and detail repletion, we decompose multi-exposure image fusion, employing neural architecture search based on principled super-net and search space, which automatically constructs effective modules with flexible adaptability.}
	
	\item A comprehensive array of experiments, encompassing both quantitative and qualitative analyses as well as extensive evaluations, emphatically demonstrate the considerable enhancements our proposed method in terms of both visual quality and inference efficiency.

\end{itemize}

\begin{figure*}[thb]
	\centering
	\includegraphics[width=0.98\textwidth,height=0.19\textheight]{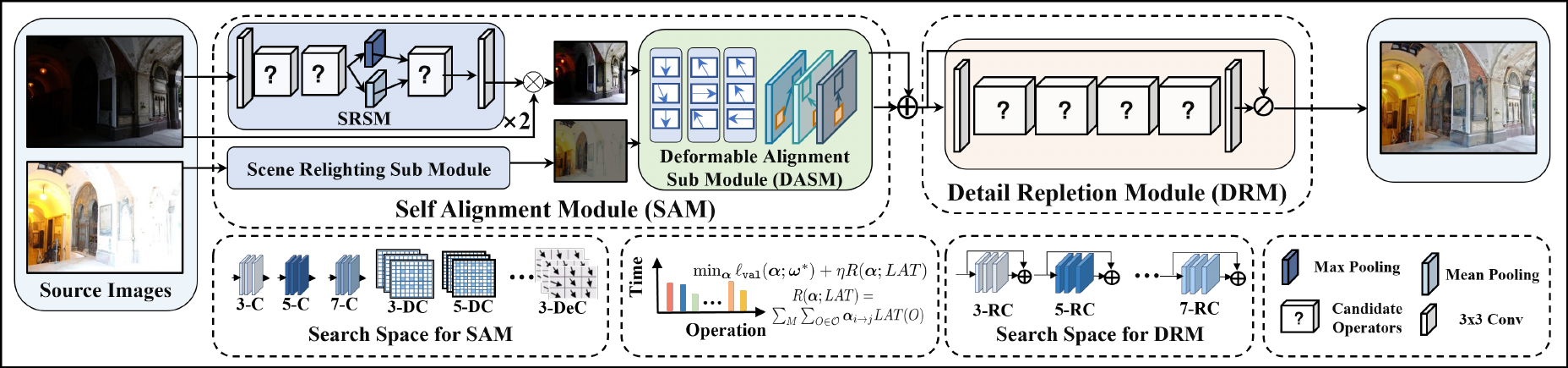}
	\caption{{ 
			Schematic diagram of the proposed architecture. The super-architecture for multi-exposure fusion consists of  Self-Alignment Module (SAM) and Detail Repletion Module (DRM).  Search spaces of SAM  and for DRM are also illustrated   respectively.
	}}
	
	\label{fig:illustration}
\end{figure*}
\section{Related Works}
In this part, we first briefly overview the related literature on multi-exposed image fusion, encompassing two prominent categories of methods: traditional numerical schemes and learning-based networks.. Then we introduce the development of related architecture search methods.
\subsection{Traditional Numerical Schemes}
In the past decades, various handcrafted numerical strategies are proposed to achieve multi-exposure fusion. These schemes can be roughly divided into transform-based, gradient-based, weighting-based, and patch-based methods. In detail, transform-based schemes first extract features and introduce fusion strategies based on the informativeness measurement. Diverse multi-scale transformations are developed to comprehensively utilize the principled features from each scale, such as wavelet transform~\cite{lewis2007pixel}, contourlet transform~\cite{qiguang2006novel}, pyramid transform~\cite{shen2014exposure} and dense invariant  transform~\cite{liu2015dense}. For instance, guided filters~\cite{li2013image} is proposed to decompose images into base and details parts on the spatial domain. By introducing the weighted average technique, this method can fuse the consistent feature comprehensively for various fusion scenarios, including multi-exposure, multi-focus, and multi-modal image fusion. Another representative technique is the Gaussian pyramids transform~\cite{li2017pixel}, which fuse source images to enhance the under/over-exposed regions progressively. 

Patch-based methods~\cite{ma2017robust,ma2015multi,li2020fast} are robust for the fusion scenarios but suffer from artifacts and blurred boundaries. Ma \textit{et al.}~\cite{ma2017robust,ma2015multi} introduced a patch-based method to measure the structural information and use the decomposition strategy to extract the richest features to form the fused images.  Kou \textit{et al.}~\cite{kou2017multi,kou2018edge} present the gradient-domain smoothing to realize edge preservation instead of Gaussian smoothing and avoid the inference of halo artifacts. Furthermore, tone-mapping-based methods are developed to achieve HDR construction with various LDR images. Sparse representation methods~\cite{yan2017high,wang2014exposure} are widely utilized for multi-exposure image fusion. These schemes utilize the overcomplete dictionary to capture the features of source images and fuse the features utilized by the corresponding sparse coefficients.  In this way, traditional schemes cannot adequately perform for challenging fusion scenarios (\textit{e.g.,} extreme exposure variation). Moreover, these schemes are also limited by the huge computation resource and the fusion performance can be reduced drastically when facing large exposure intervals.

\subsection{Learning-based Schemes}
With the flourishing progress of the deep learning paradigm, learning-based methods realize the promising improvement in the quantity and quality of multi-exposure fusion tasks, compared with traditional methods. Supervised by the MEF-SSIM metric, DeepFuse pioneered the first learning framework to aggregate the luminance components, and utilized a weighted fusion strategy to fuse color and 
brightness components. Ma~\textit{et al.} proposed MEF-Net~\cite{ma2019deep} to predict the weighted map by feeding down-sampled images. Zhang~\textit{et al.} presented the network IFCNN~\cite{zhang2020ifcnn} to adopt the element-wise feature fusion based on the features extracted from two independent branches.
The above networks either trained by unsatisfied metric (\textit{e.g.,} MEF-SSIM~\cite{ma2015perceptual}) or based on the local pixel-wise feature fusion, are easy to result in the color distortion and global structural inconsistency. On the other hand,
There are various unified learning-based schemes to uniformly address diverse image fusion tasks. Zhang \textit{et al.} proposes a densenet with the squeeze and decomposition principle, called SDNet~\cite{zhang2021sdnet} to realize the versatile image fusion framework. A multi-decoder-based framework is introduced with a shared encoder to realize the unified fusion~\cite{li2021multiple}. U2Fusion~\cite{xu2020u2fusion} utilizes the feature similarity based on the gradient to measure the differences between source and fusion images. These versatile fusion methods pursuit  discoverer the similarity of tasks, inevitably lacking task-oriented consideration, thus leading to color distortion and structural detail degradation.

Lately, attention mechanisms are widely used for multi-exposure fusion. Liu~\textit{et al.} propose a hierarchical attention module~\cite{liu2021halder,liu2022attention} to investigate the sufficient information on both under/over-exposed images.  Yan~\textit{et al.}~\cite{yan2019attention} introduce the dual spatial attention module to remove the ghosts and misalignments of adjoint frames.
Xu~\textit{et al.}~\cite{xu2020mef} introduce the non-local self-attention block to capture the long-range dependency of all regions with a generative adversarial network. Similarly, Li~\textit{et al.} design various attention modules~\cite{li2022learning} (\textit{e.g.,} coordinate and self-attention) to extract the texture details from source images. Han~\textit{et al.}~\cite{han2022multi} decouple the MEF task into two deep perceptual enhancements including content detail extraction and color correction.
We emphasize that the mainstream learning-based schemes put significant effort into designing attention modules (\textit{e.g.,} hierarchical attention~\cite{liu2021halder}, non-local attention~\cite{xu2020mef}) to realize the visual-pleasant realistic color correction. However, these schemes mostly utilize the heuristic attention architectures, which cannot achieve the adaptive feature extraction among diverse fusion scenes and suffer from the slow inference time with abundant parameters. Thus, 
the texture information from source images cannot be sufficiently investigated to generate reasonable results. \textcolor{black}{Luo~\textit{et.al}~\cite{luo2023multi} propose the deformable self-attention to construct the multi-scale alignment for refining contextual features and details, performing adaptive image fusion. In order to alleviate the modal difference, Wang \textit{et.al}~\cite{wang2022unsupervised} and Xu \textit{et.al}~\cite{xu2022rfnet} leverage the style transfer to convert the one modality into other ones, to guarantee the consistency of modal information for infrared-visible image fusion. Moreover, Xu \textit{et.al}~\cite{xu2023murf} also proposes the shared information extraction to transform diverse modals into shared feature spaces to eliminate the modal variances. In order to solve the domain and spatial misalignment for pseudo-supervision, AlignFormer~\cite{feng2023generating} is proposed to mitigate the discrepancy by the incorporation of geometric cues. In recent years, exposure correction and low-light enhancement schemes have obtained wide attention to recover visual-appealing results from single degraded images. Li~\textit{et.al} propose the zero-reference curve estimation methods~\cite{guo2020zero,li2021learning} based on a   lightweight deep network for dynamic range adjustment. Inheriting this paradigm, CuDi~\cite{li2022cudi} is proposed to speed up the inference and realize the controllable exposure adjustment with a  novel curve distillation. Ren \textit{et.al}~\cite{ren2019low} present the deep hybrid network to integrate the learning of global content and the salient structures for low light image enhancement. Luo \textit{et.al}~\cite{luo2022under} propose the cascaded curve estimation, leveraging attention-aware features for under-display camera image enhancement.}

\textcolor{black}{Compared with existing  schemes~\cite{liu2022attention,li2021detail,li2022learning}, we highlight the several limitation. A critical limitation of these methods lies in their inadequate consideration of detail fusion on misalignment scenarios. The prevalent approaches, whether employing numerical optimization or  learning techniques, heavily rely on handcrafted architectural engineering, easily leading to slow inference and heightened computational complexity. In contrast, our method designs a comprehensive MEF-specific architecture including self-alignment and detail refinement, adapting for diverse scenarios. Furthermore, we design an automatic NAS-based scheme for desired architecture construction.}

\section{Proposed Method}
% In this Section, we develop
%scene-oriented attention mechanisms and detail repletion by exploiting the neural architecture search strategy. Imposing the hardware-latency constraint into the search procedure, we can discover the effective architecture with high efficiency. Another challenge needs to be concerned is to address the misalignment in real imaging conditions.
%We introduce the self-alignment module to achieve the pixel-level alignment.

We first introduce a robust multi-exposure fusion framework to address the misalignment between source images and the visual aesthetics of fused images. Then we introduce a hardware-latency-constrained architecture search with corresponding loss functions to discover the nimble network for fast inference. Concrete components are schematically illustrated in Fig.~\ref{fig:illustration}.
\subsection{Robust Multi-Exposure Fusion Framework}

\subsubsection{Self-alignment Module} 
As aforementioned, in the real world, the misalignment of image pairs caused by device shaking and movement is almost inevitable. On the other hand, due to the extreme exposure intervals of pairs, it is untoward to straightforwardly utilize alignment techniques (\textit{e.g.,} optical flow~\cite{teed2020raft}, registration module~\cite{wang2022unsupervised,huang2022reconet}), which may produce texture artifacts under inexact alignment. Thus, we conquer this obstacle in two steps: scene relighting for illumination correction and deformable aligning for feature registration.

%However, current schemes utilize the registered image pairs  $\mathbf{I}_\mathtt{U}, \mathbf{I}_\mathtt{O}$ with distinct exposure time.

%Given two source images with distinct exposure time,, the most straightforward scheme is to extract informative details (\textit{e.g.,} salient textures and color distribution) from source images based on diverse attention mechanisms.

\textbf{Scene Relighting Sub-Module:}
In essence, based on the Retinex theory, we propose a recurrent adaptative attention mechanism for scene relighting, aiming to push the images into a similar illumination domain.
We introduce two Scene Relighting Sub-Modules (SRSM) for each image to restrain the degree of illumination of source images into a similar domain for following alignment and detail enhancement. Furthermore, instead of targeting to restore the normal-light scene from single sources, SRSM aims to leverage the illumination map to preserve the comprehensive structures.

Denoted the intermediate results as $\mathbf{I}_\mathtt{U}^\mathtt{S}$ and $\mathbf{I}_\mathtt{O}^\mathtt{S}$ and SRSM as $\mathcal{S}$, the illumination correction can be formulated as 
\begin{equation}\label{eq:srsm}
	\mathbf{I}_{i}^\mathtt{S} = \mathbf{I}_{i} \otimes \mathcal{S}(\mathbf{I}_{i}), {i} \in \{\mathtt{U};\mathtt{O}\},
\end{equation}
where $\otimes$ denotes the element-wise multiplication.  $\mathtt{O}$ and $\mathtt{U}$ represent under/over exposed images. Noting that, we exploit a recurrent gradual scheme to cascade SRSM, aiming to realize the progressive illumination correction. The stage-wise attention maps can benefit the procedure of complementary feature learning fully and elaborately.

As shown in the left part of Fig.~\ref{fig:illustration}, rather than utilizing heuristic handcrafted methodology, we leverage the differentiable architecture search to construct this module for fast scene-adaption. In detail, we first utilize one $3\times 3$ convolution to transfer the image into the feature domain. Then we set two candidate operations to extract scene features. Max pooling and average pooling are hierarchically embedded to realize the amplification of salient features for illumination estimation completely. Then we leverage one undetermined convolution layer to 
boost the information richness of features and utilize one $3\times 3$ convolution with sigmoid function to generate a three-channel illumination map with range [0,1].

\textbf{Deformable Aligning Sub-Module:} Few multi-exposure fusion methods consider the misalignment of source images, which are based on pre-registered pairs.
However, in real-world scenes, the misalignment of  
over/under-exposed images would damage the visual quality with serious ghost artifacts, due to the movement of image devices or targets. Moreover, introducing learning-based optical flow methods would lead to the huge computation of pixel motion. The lack of real optical flow 
as ground truth for pre-training limits their performance.
Thus, we introduce the Pyramid, Cascading, and Deformable Convolution (PCD) mechanism~\cite{wang2019edvr}  to establish a Deformable Aligning Sub-Module (DASM) based on the supervision of visual quality metrics. We only consider DASM under the misalignment scenario.

Specifically, DASM first employs diverse stridden convolution to generate pyramid features $\mathbf{F}_\mathtt{U}$ and  $\mathbf{F}_\mathtt{O}$ based on the intermediate results from SRSM, we utilize deformable convolutions to conduct the feature-level alignment by coarse-to-fine manner.
Denoting the DASM as $\mathcal{A}$, we can obtain the comprehensive feature as 
\begin{equation}\label{eq:SRSM}
	\mathbf{F}_\mathtt{A} = \mathcal{A}(\mathbf{F}_\mathtt{U},\mathbf{F}_\mathtt{O}) + \mathbf{F}_\mathtt{O},
\end{equation}
where  $\mathbf{F}_\mathtt{A}$ represents the fused features based on the summation of aligned source features. Similarly, instead of introducing the original PCD network, we employ an architecture search scheme to rebuild the structure (\textit{i.e.,} replacing different kernels of deformable convolutions) to accommodate itself into a multi-exposure fusion task. 

\subsubsection{Detail Repletion Module}
Then we introduce the Detail Repletion Module (DRM) to enhance the textural details of complementary features. \textcolor{black}{We argue that only aggregating the images based on the self-align modules cannot reconstruct the desired illumination and textural details.}
In order to preserve the spatial structures, we utilize successive structures under the same resolution to promote information richness. \textcolor{black}{Specifically, inspired by effective residual learning mechanisms for image restoration tasks (\textit{e.g.,} residual dense blocks~\cite{zhang2020residual} and dilated dense block~\cite{yan2019attention}), we introduce a residual operator-based search space to discover a suitable dense structure. Subsequently,
	we investigate the illumination restoration and global color distortion based on pixel-wise division.} Thus, denoted the network as $\mathcal{R}$ and output as $\mathbf{y}$, we can formulate the optimization procedure as
\begin{equation}\label{eq:drm}
	\mathbf{y} = \mathbf{F}_\mathtt{A} \oslash \mathcal{R}(\mathbf{F}_\mathtt{A}).
\end{equation}
In a word, DRM not only targets to strengthen feature representation of details from the fused features, but also protects the integral normal illumination. Specifically, we employ four candidate operators to composite this module. Lastly, we utilize one $3\times3$ convolution layer with a sigmoid function to estimate the illumination map. In the following, we will discuss the concrete  strategy to search for compact
MEF framework.

\subsection{Automatic Architecture Construction}
In this part, we introduce the detailed search space and strategy for lightweight effective architectures.

\subsubsection{Principle-driven Search Space}
Different from recent NAS-based schemes~\cite{liu2021searching,liu2021retinex}, which introduces the single operators (\textit{e.g.,} one-layer convolution and primitive pooling operations)
to composite the search space, without the deep investigation of principles for module-related characteristics, we construct the principle-driven search space.
As shown in the bottom part of Fig.~\ref{fig:illustration}, normal convolutions (denoted as ``C'') and dilated convolutions (denoted as ``DC'')  with different kernel size $k\times k, k \in \{1,3,5,7\}$ are utilized for the SRSM, which are consisted by three layers of convolutions for feature representation and dimension changing. In order to persevere the sufficient features to recover the complementary information, we add the skip connection to establish the residual learning, which is denoted as ``RConv'' and 
``RDConv'' respectively. Similarly, DASM also can be searched using three kinds of deformable convolutions, denoted as ``3-DeC", ``5-DeC" and ``7-DeC" respectively. 
\subsubsection{Compact Architecture Search}
In this paper, following with the continuous relaxation~\cite{liu2018darts}, we introduce the architecture weight $\bm{\alpha}$ to connect the operators  from search space $\mathcal{O}$ for the super-net construction. The continuous relaxation from layer $i$ to layer $j$ is formulated as:
\begin{equation}\label{eq:cr}
	\mathbf{F}_{j} = \widetilde{O}_{i\rightarrow j}(\mathbf{F}_{i});\quad	\widetilde{O}_{i\rightarrow j} =\sum_{{O}\in \mathcal{O}}\bm{\alpha}_{i\rightarrow j} O(\mathbf{F}_{i}),
\end{equation}
where the relaxation operator is denoted as $\widetilde{O}$ and $\sum_{{O}\in \mathcal{O}}\bm{\alpha}_{i\rightarrow j} = 1$. In order to obtain the desired architecture with high performance and fast inference time, we also establish continuous relaxation with operation latency. In this way, we can obtain the inference time of this super-net:
\begin{equation}\label{eq:cr}
	\text{R}(\bm{\alpha};\text{LAT}) =\sum_{M}\sum_{{O}\in \mathcal{O}}\bm{\alpha}_{i\rightarrow j} \text{LAT}(O),
\end{equation}
where $M$ denotes the number of search blocks.
Thus, we introduce the summation of  operation latency $\text{R}(\bm{\alpha};\text{LAT})$ as the constraint for architecture search objective, which  can be expressed as:
\begin{equation}\label{eq:hs}
	\min_{\bm{\alpha}}\ell_\mathtt{val}(\bm{\alpha};\bm{\omega}^{*})+\eta \text{R}(\bm{\alpha};\text{LAT}) ,
\end{equation}
where $\ell_\mathtt{val}$ and $\bm{\omega}^{*}$ are the validation loss and optimal parameters based on the training data. Introducing the differentiable search strategy, we conducted the  search of whole super-net.

\subsection{Loss Functions}
Focusing on the texture details preservation, color information promotion and global scene consistency, we leverage three categories of loss functions to train the proposed network, including
pixel-intensity loss $\ell_\mathtt{Int}$, gradient loss $\ell_\mathtt{Gra}$ and 
global-adversarial loss $\ell_\mathtt{Dis}$ by supervised learning with ground truth $\mathbf{y}_\mathtt{gt}$. On the whole, the total loss measurement $\ell_\mathtt{Total}$ is denoted as:
\begin{equation}\label{eq:totloss}
	\ell_\mathtt{Total} = \ell_\mathtt{Int} + \beta_{1} \ell_\mathtt{Gra}  +\beta_{2} \ell_\mathtt{Dis}.
\end{equation}
where $\{ \beta_{1},\beta_{2}\}$ are a series of trade-off parameters. 

To ensure the same intensity distribution as the ground truth image (denoted as $\mathbf{y}_\mathtt{gt}$), we impose the $\ell_\mathtt{1}$ distance to measure the discrepancy, which can be formulated as:
\begin{equation}\label{eq:mseloss}
	\ell_\mathtt{Int} = \frac{1}{HW}\|\mathbf{y}-\mathbf{y}_\mathtt{gt}\|_{1},
\end{equation}
where $H,W$ denote the height and width of image.

Due to interference by noises and corrupted exposures, source images lack partial details. We utilize the Sobel operator to preserve the fine-grained texture details.
\begin{equation}\label{eq:graloss}
	\ell_\mathtt{Gra} = \frac{1}{HW}\|\nabla\mathbf{y}-\nabla\mathbf{y}_\mathtt{gt}\|_{2},
\end{equation}
%Except the pixel-level measurement, we also introduce the deep feature-level constraint to minimize the differences, using pretrained VGG-19 network $\phi$. The perception loss can effectively remove the ghosts and provide more natural details. The defination can be formulated as 
%\begin{equation}\label{eq:perloss}
%	\ell_\mathtt{Per} = \sum_{l} \frac{1}{C_{l} H_{l } W_{l}}\|\phi_{l}(\mathbf{y})-\phi_{l}(\mathbf{y}_\mathtt{gt})\|_{2},
%\end{equation}
%where $l$ indicates the index of VGG network and $C$ represents the feature channel. We set 4 layers to measure it. 

To address the deficiency of local region information and achieve global consistency in color distribution, we introduce the discriminator $\mathcal{D}$ from PatchGAN~\cite{liu2019perceptual} to judge the generated results with a global
activation map. By incorporating this constraint, color distribution of whole scene can be guaranteed. We introduce the gradient-penalty wasserstein  training strategy~\cite{gulrajani2017improved} to conduct the generative adversarial learning. $\ell_\mathtt{Dis}$ is formulated as:
\begin{equation}\label{eq:gangp}
	\mathbb{E}_{\tilde{\mathbf{x}}\sim\mathbb{P}_\mathtt{fake}}\mathcal{D}(\mathbf{y}) - \mathbb{E}_{{\mathbf{x}}\sim\mathbb{P}_\mathtt{real}}\mathcal{D}(\mathbf{y}_\mathtt{gt})\\+\eta \mathbb{E}_{\tilde{\mathbf{x}}\sim\mathbb{P}_\mathtt{fake}} [(\|\nabla_{\mathbf{y}}\mathcal{D}(\mathbf{y})\|_{2}-1)^2].
\end{equation} 
\begin{figure*}[thb]
	\centering \begin{tabular}{c@{\extracolsep{0.15em}}c@{\extracolsep{0.15em}}c@{\extracolsep{0.15em}}c@{\extracolsep{0.15em}}c@{\extracolsep{0.15em}}c}
		
		\includegraphics[width=0.164\textwidth]{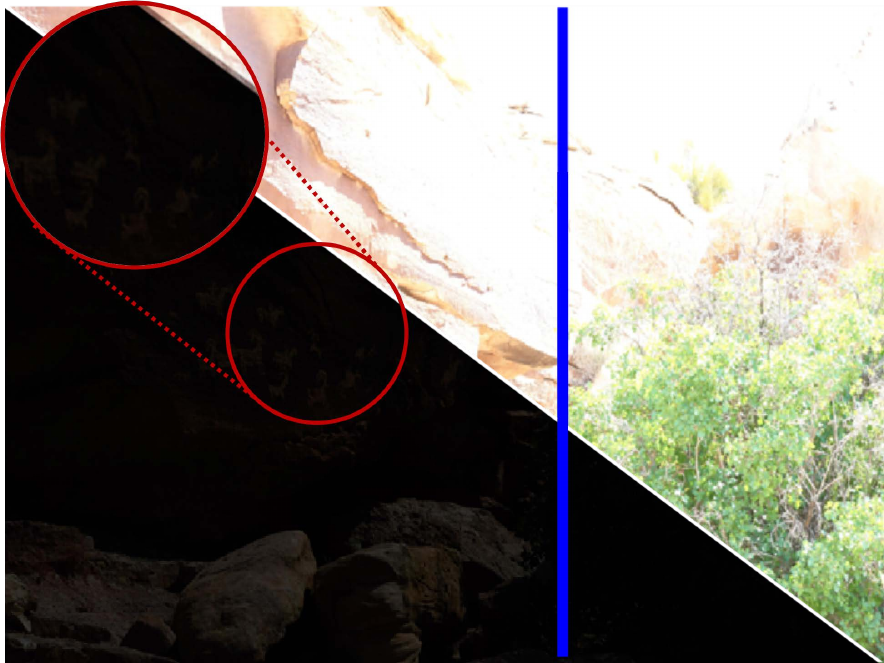}
		&		\includegraphics[width=0.164\textwidth]{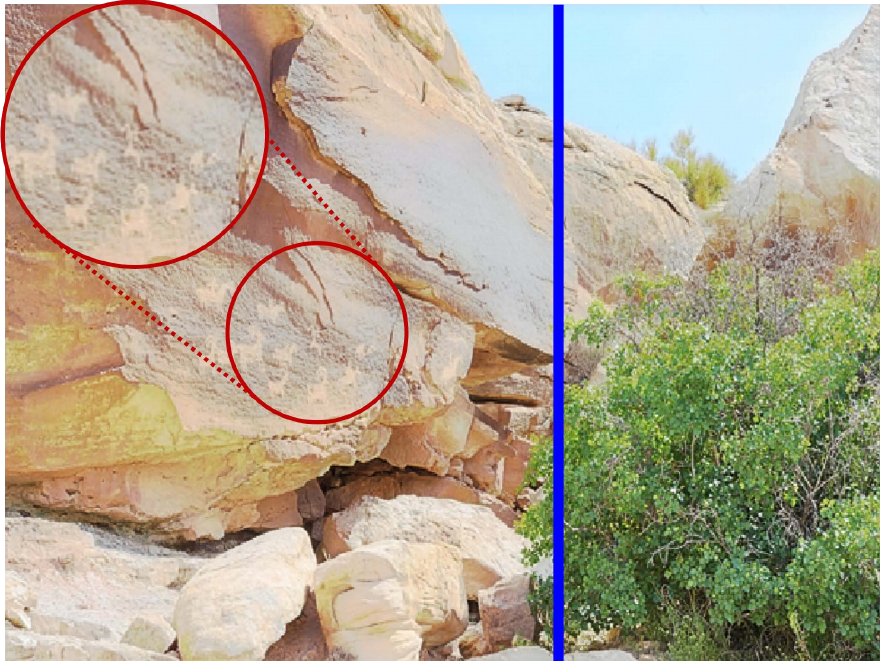}
		
		&		\includegraphics[width=0.164\textwidth]{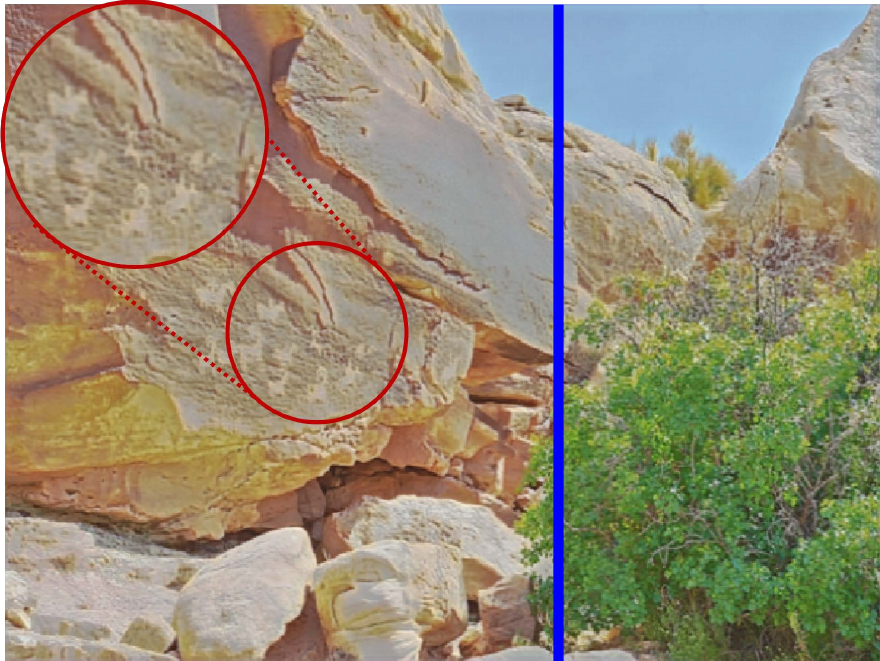}
		&		\includegraphics[width=0.164\textwidth]{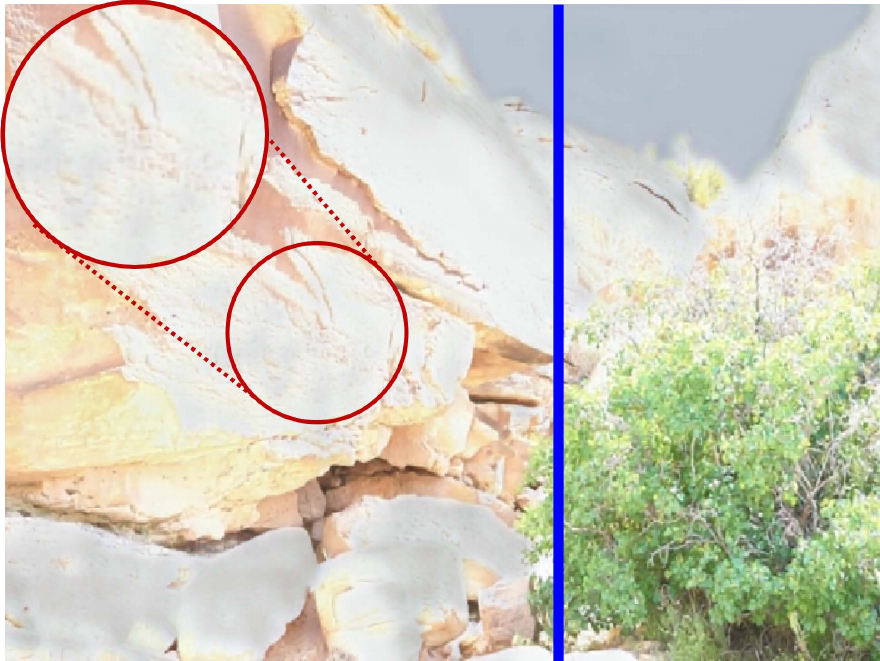}
		&		\includegraphics[width=0.164\textwidth]{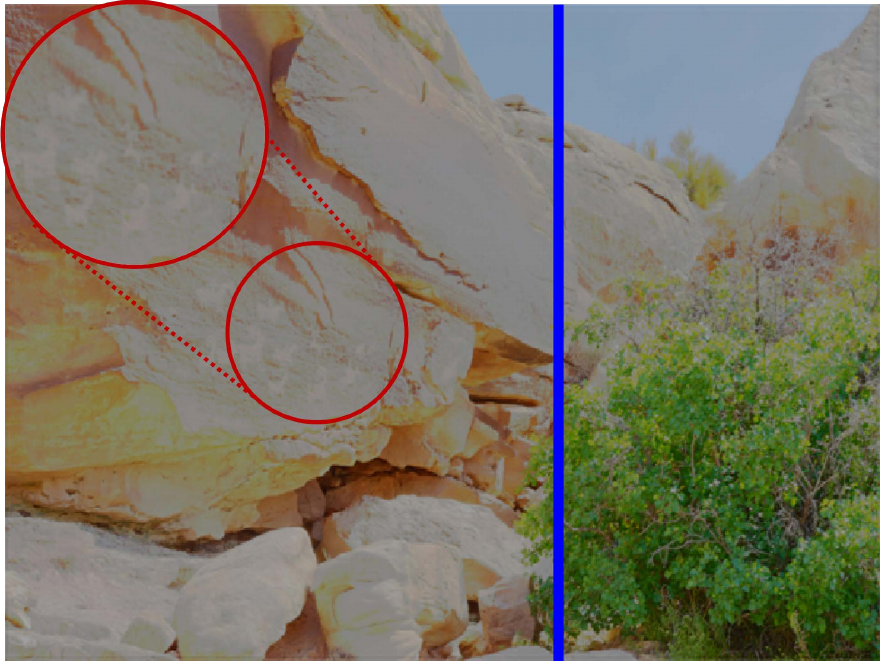}
		&	\includegraphics[width=0.164\textwidth]{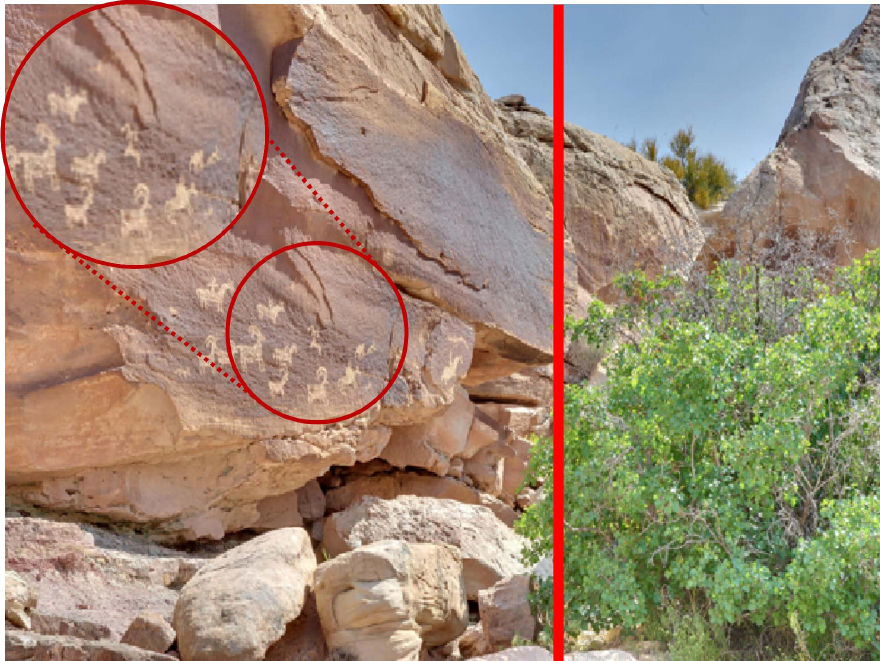}\\
		
		\includegraphics[width=0.164\textwidth]{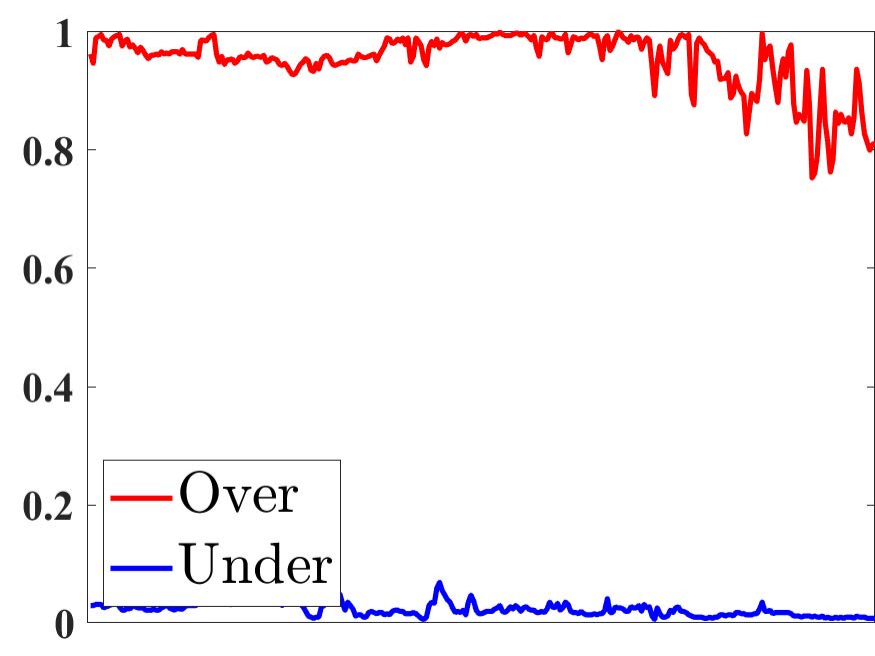}
		&		\includegraphics[width=0.164\textwidth]{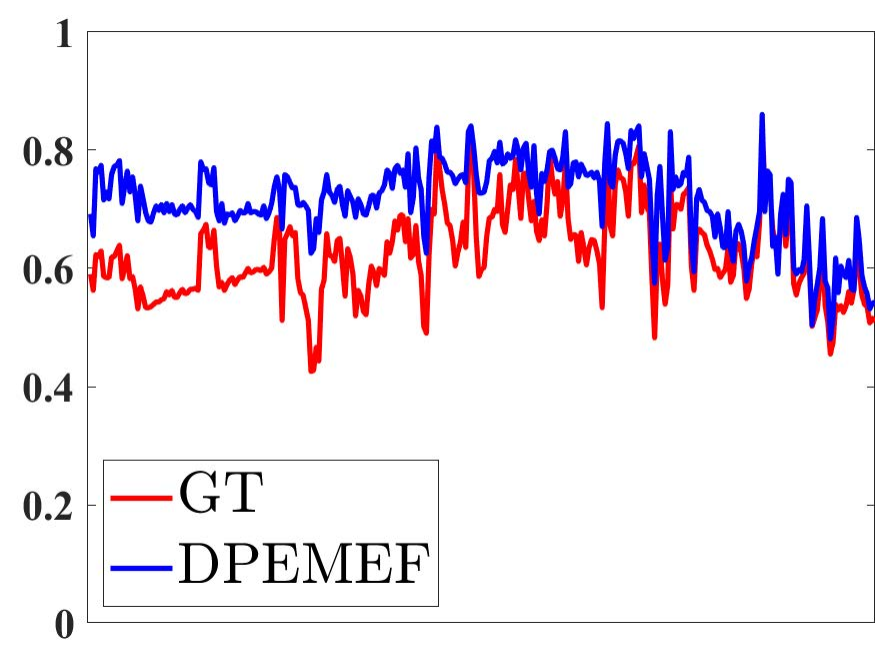}
		&		\includegraphics[width=0.164\textwidth]{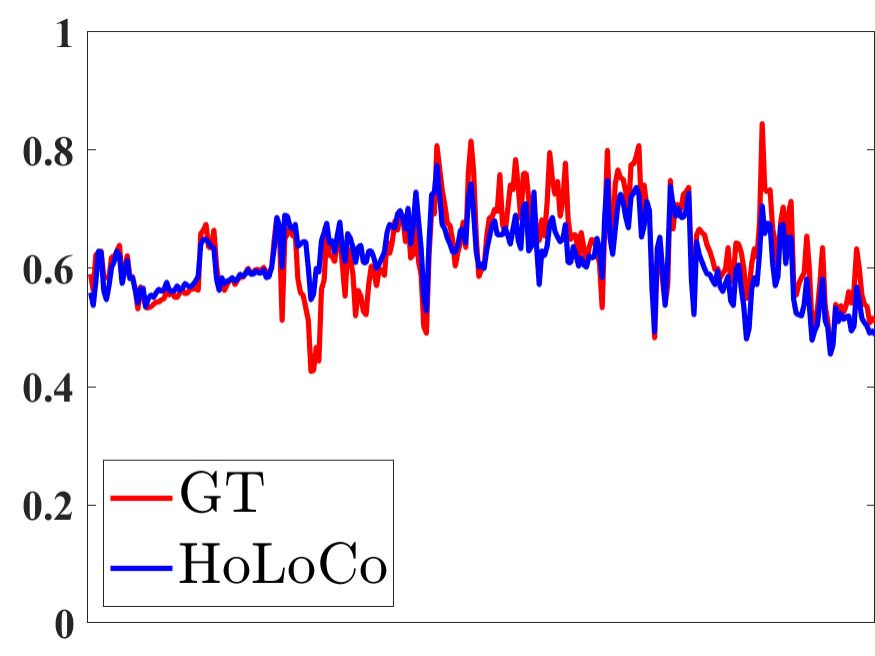}
		&		\includegraphics[width=0.164\textwidth]{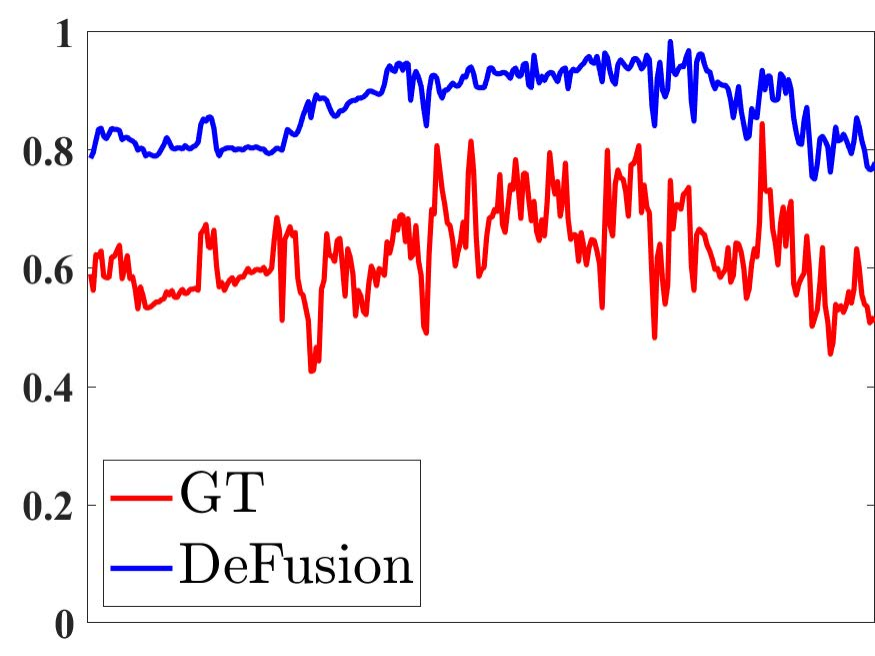}
		&		\includegraphics[width=0.164\textwidth]{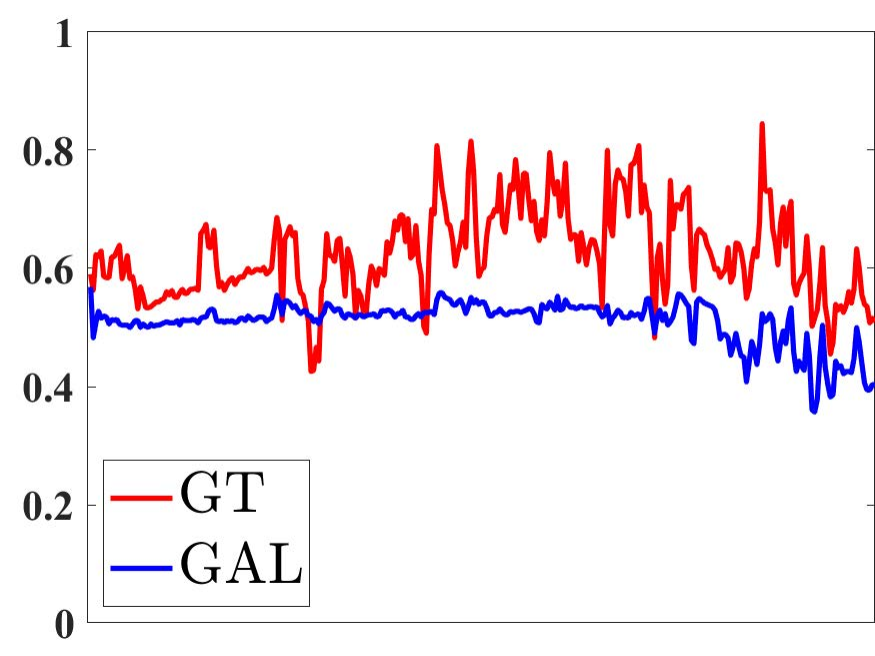}
		&	\includegraphics[width=0.164\textwidth]{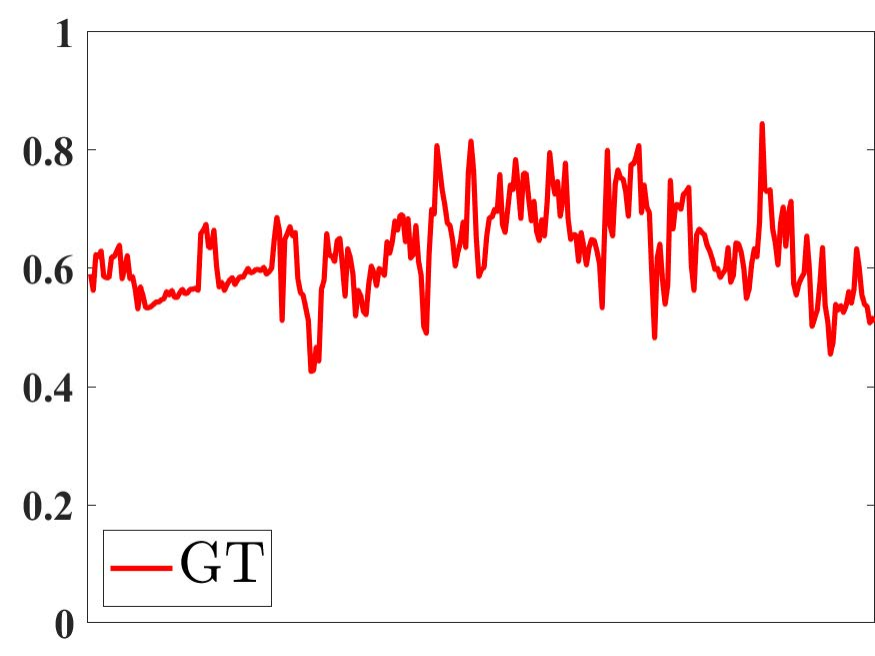}\\
		\footnotesize	Input images & \footnotesize DPEMEF &  \footnotesize HoLoCo &  \footnotesize DeFusion  & \footnotesize GAL & \footnotesize Ground Truth \\
		\includegraphics[width=0.164\textwidth]{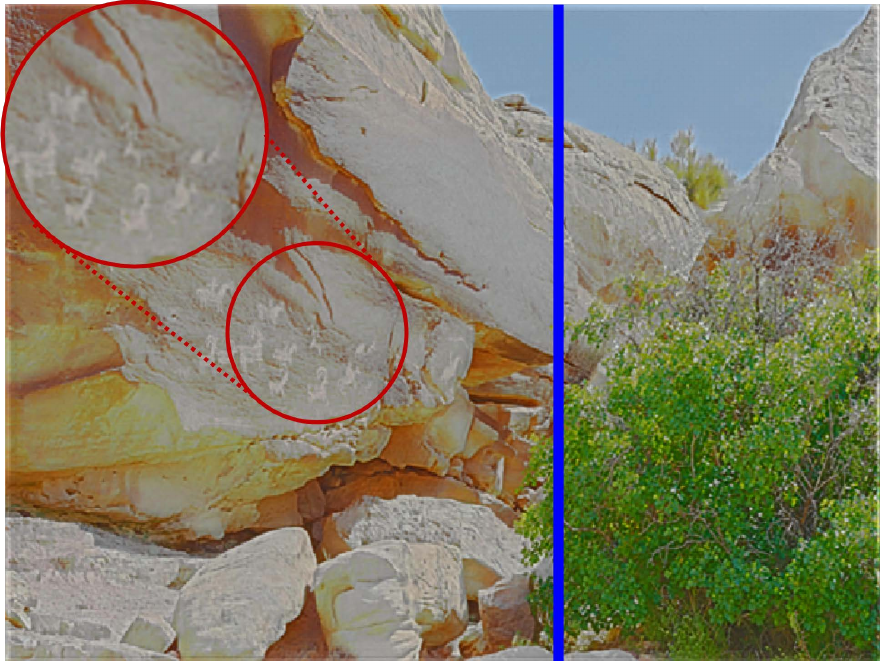}
		&	\includegraphics[width=0.164\textwidth]{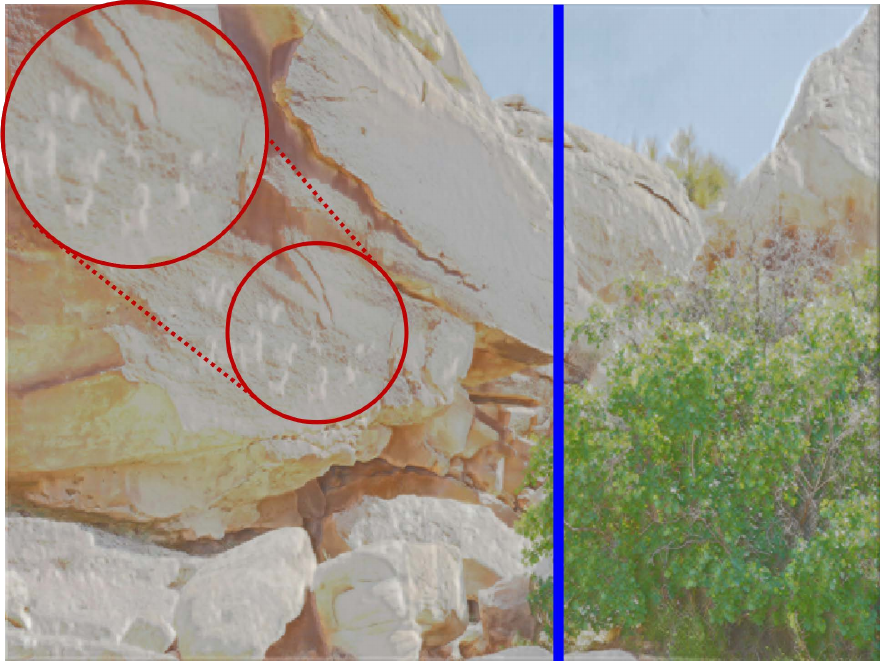}
		&		\includegraphics[width=0.164\textwidth]{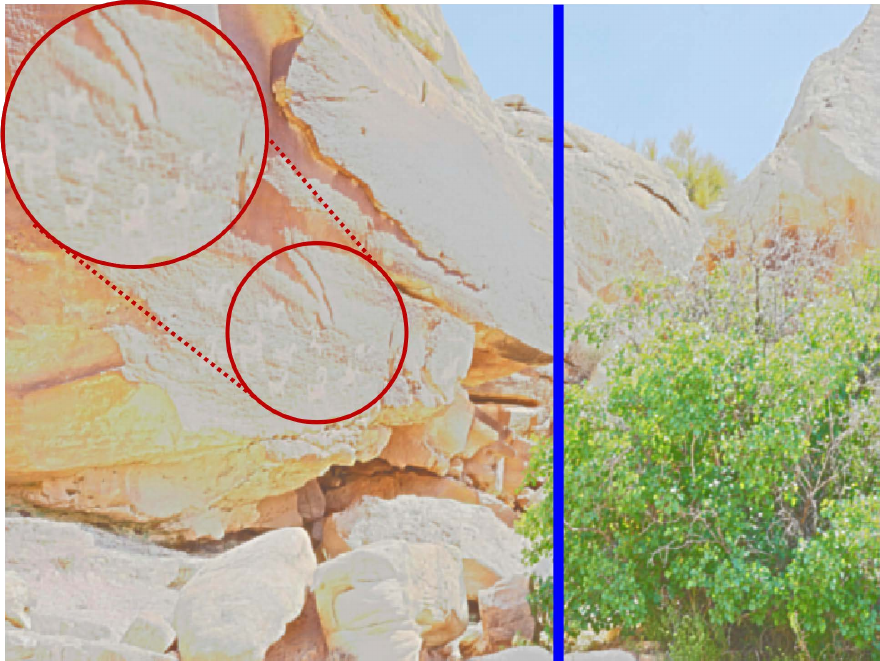}
		
		&		\includegraphics[width=0.164\textwidth]{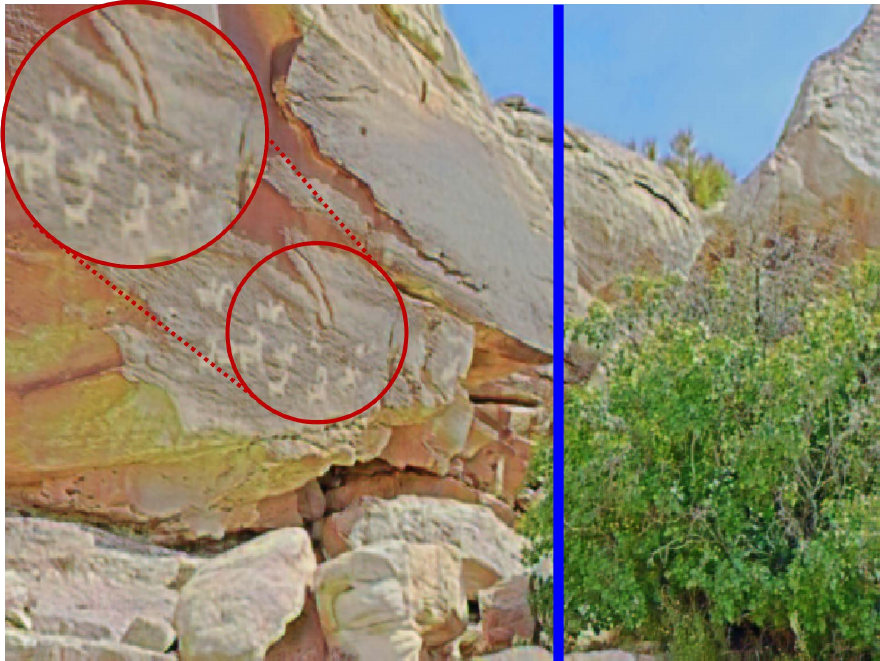}
		&		\includegraphics[width=0.164\textwidth]{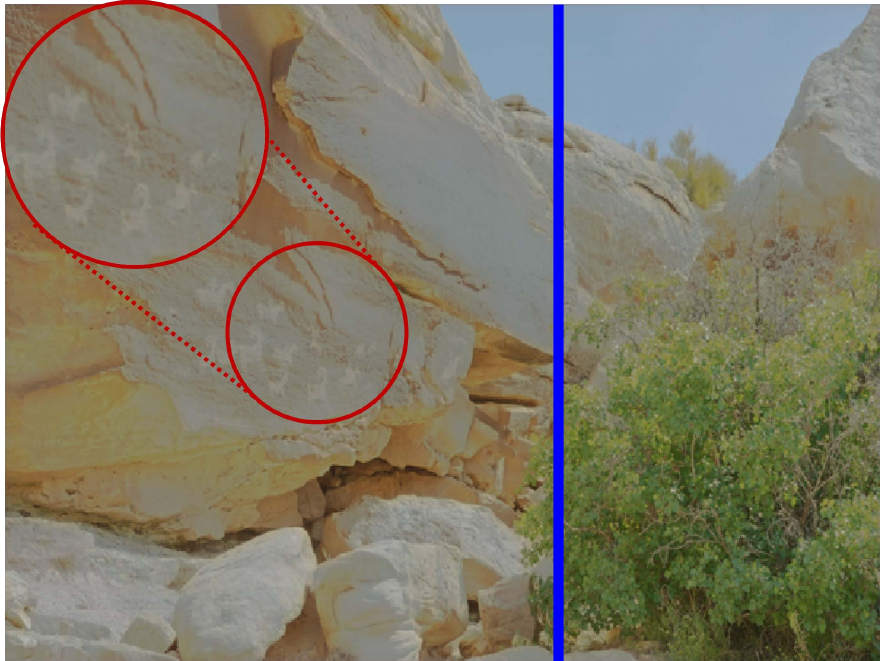}
		&	\includegraphics[width=0.164\textwidth]{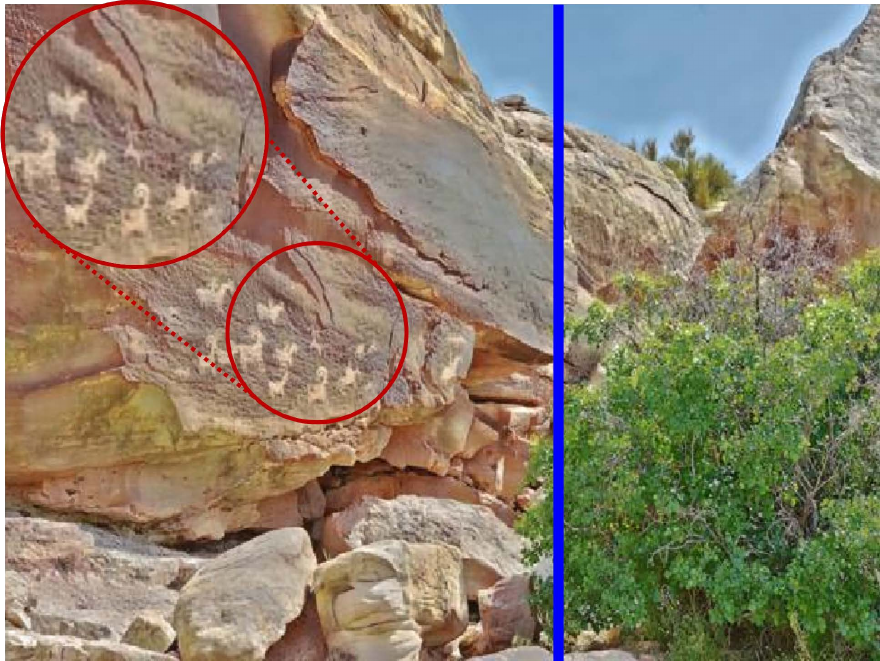}\\
		\includegraphics[width=0.164\textwidth]{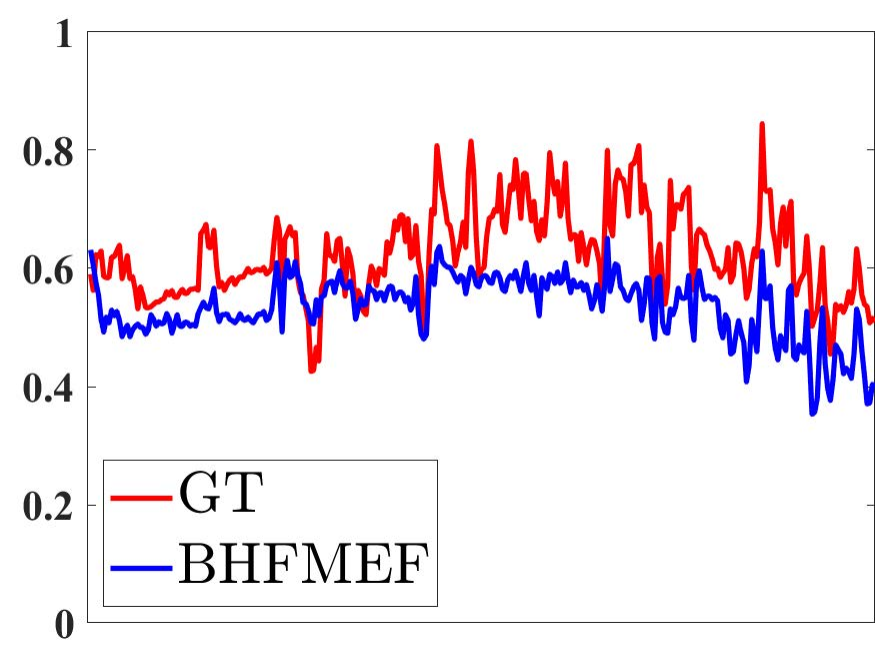}
		&		\includegraphics[width=0.164\textwidth]{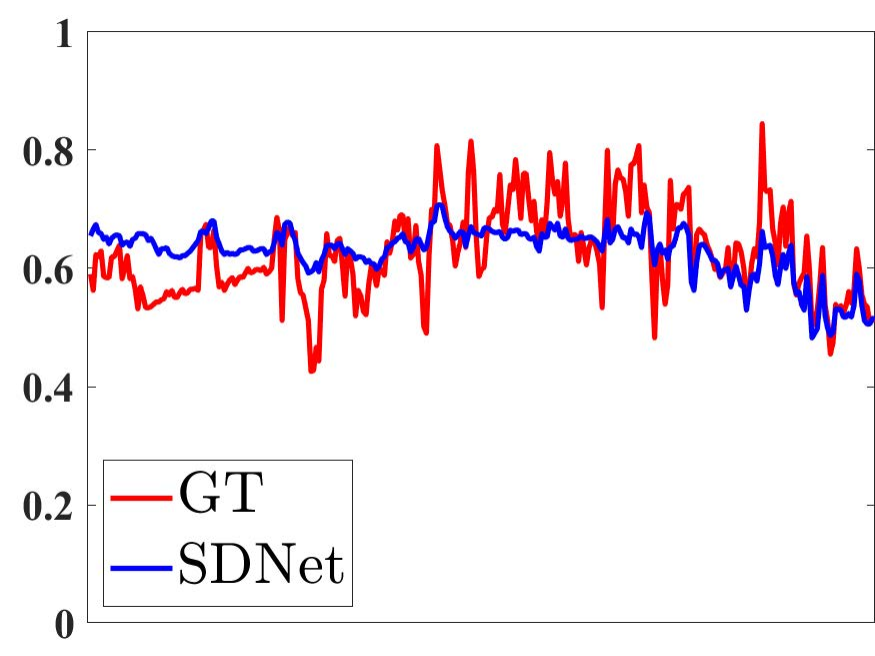}
		&		\includegraphics[width=0.164\textwidth]{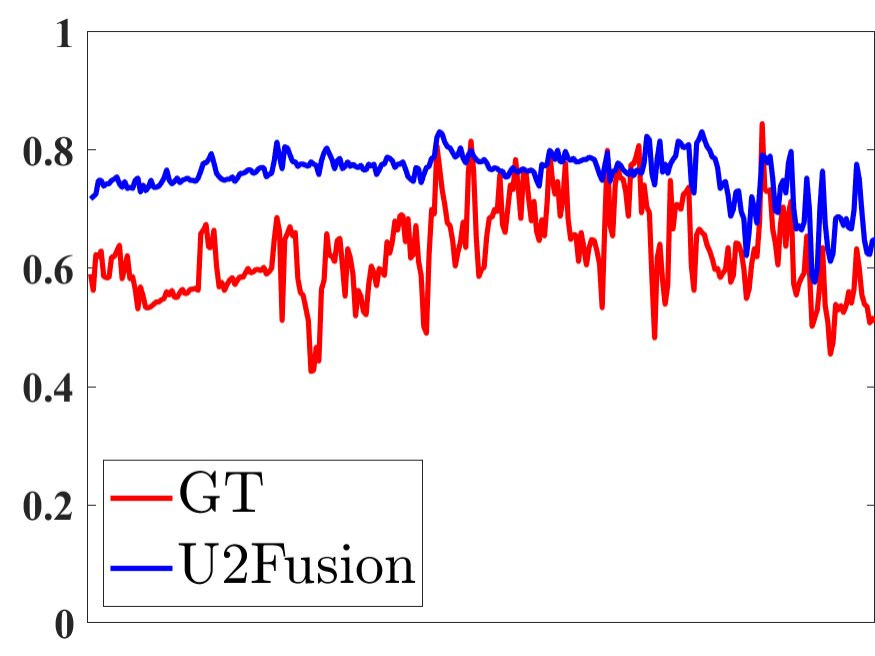}
		&		\includegraphics[width=0.164\textwidth]{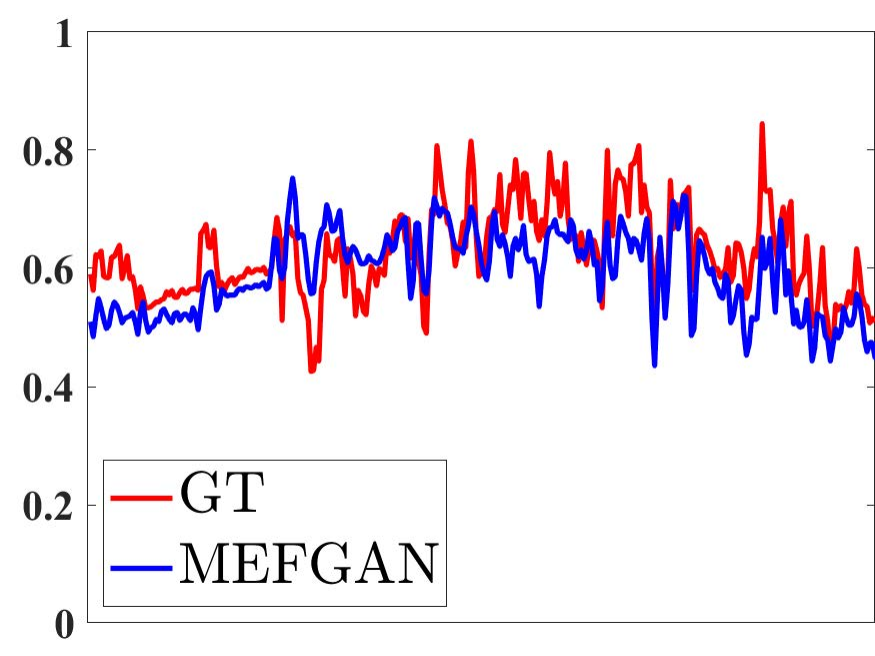}
		&		\includegraphics[width=0.164\textwidth]{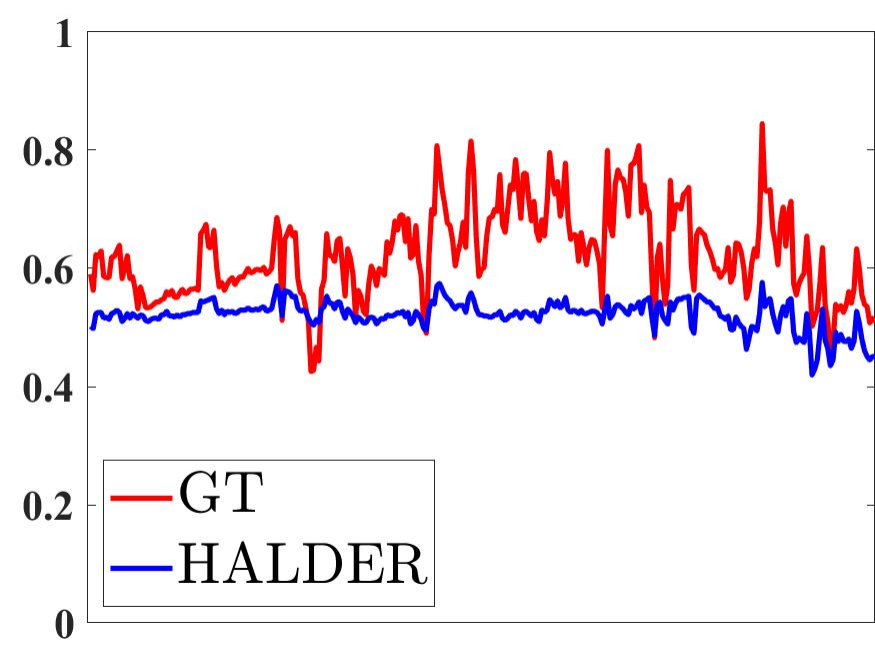}
		&	\includegraphics[width=0.164\textwidth]{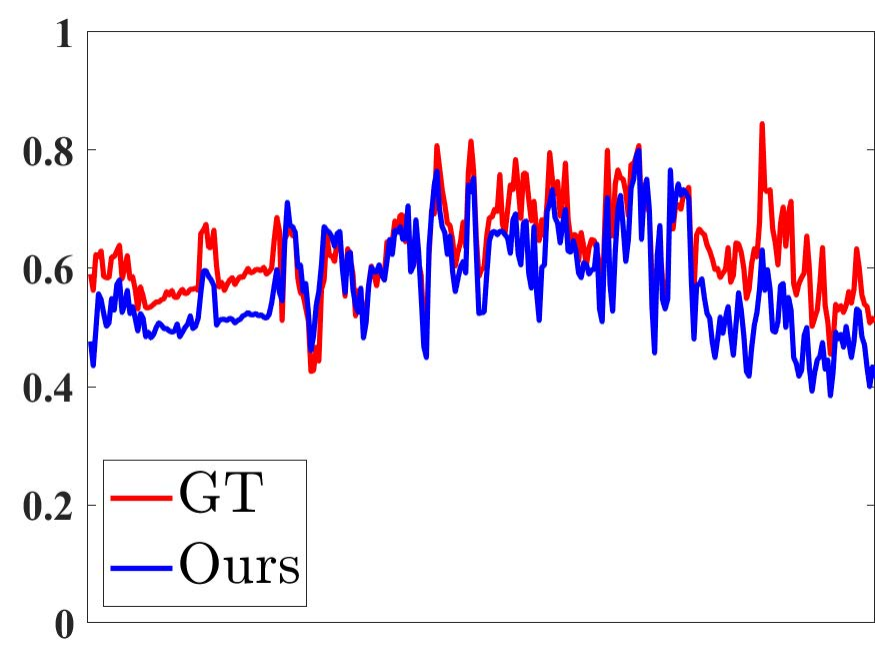}\\
		\footnotesize	 BHFMEF  & \footnotesize SDNet & \footnotesize  U2Fusion & \footnotesize MEFGAN &  \footnotesize HALDER & \footnotesize Ours \\
	\end{tabular}
	\caption{\textcolor{black}{Qualitative comparison with nine state-of-the-art methods. The signal maps provide the differences of pixel intensity with the ground truth.}}
	\label{fig:gmefusion}
\end{figure*}

\begin{figure*}[thb]
	\centering \begin{tabular}{c@{\extracolsep{0.15em}}c@{\extracolsep{0.15em}}c@{\extracolsep{0.15em}}c@{\extracolsep{0.15em}}c@{\extracolsep{0.15em}}c}
		
		\includegraphics[width=0.164\textwidth]{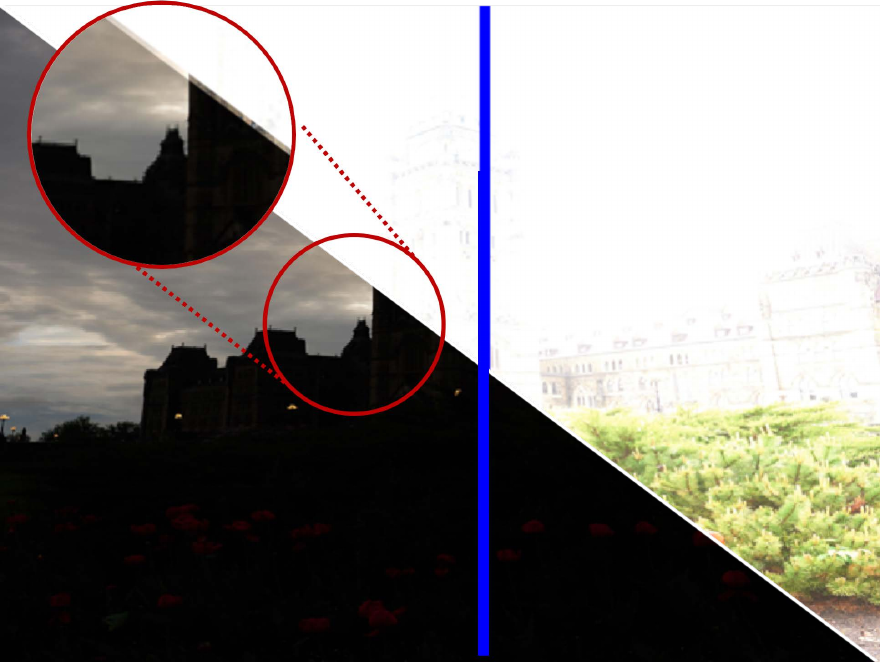}
		&		\includegraphics[width=0.164\textwidth]{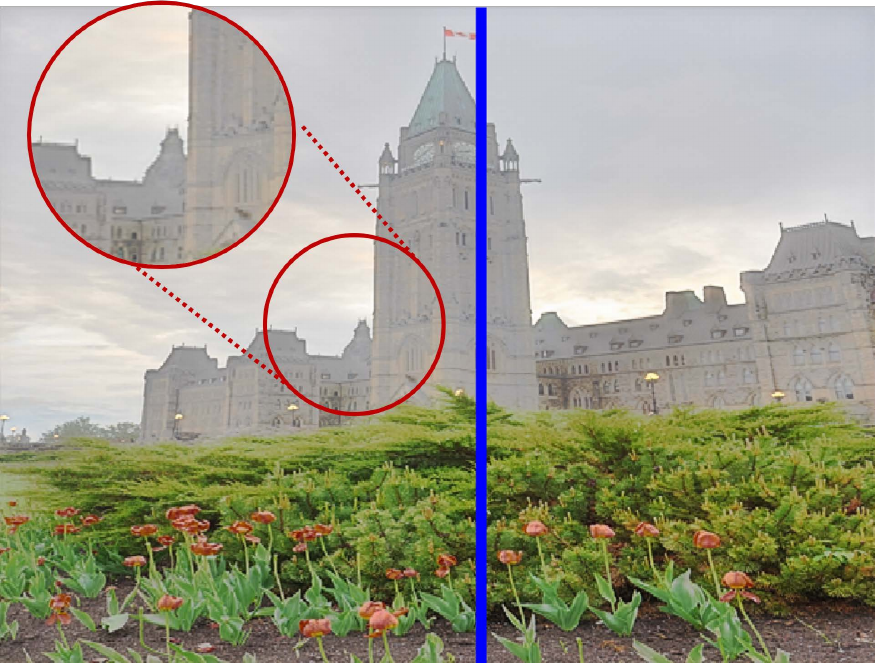}
		
		&		\includegraphics[width=0.164\textwidth]{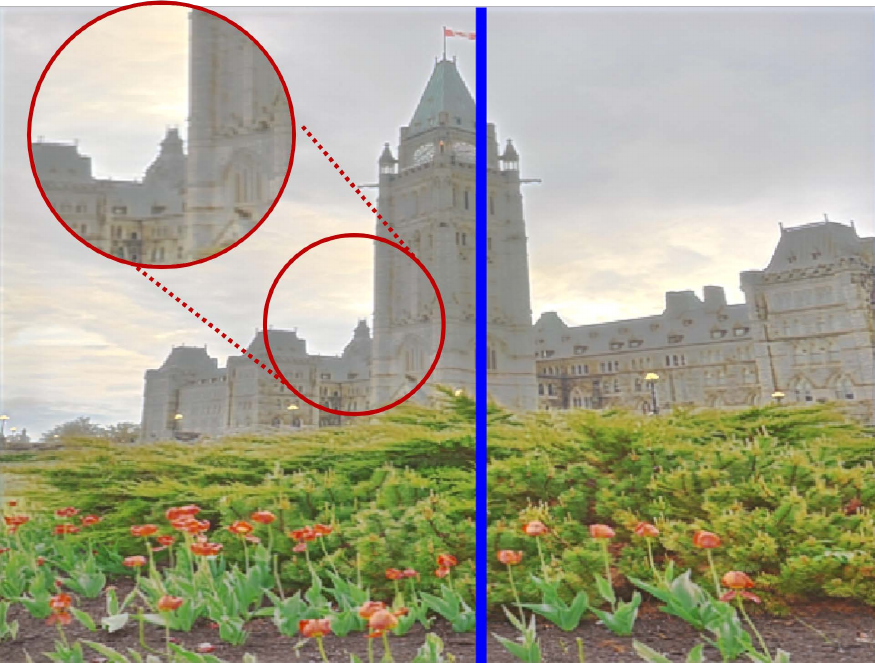}
		&		\includegraphics[width=0.164\textwidth]{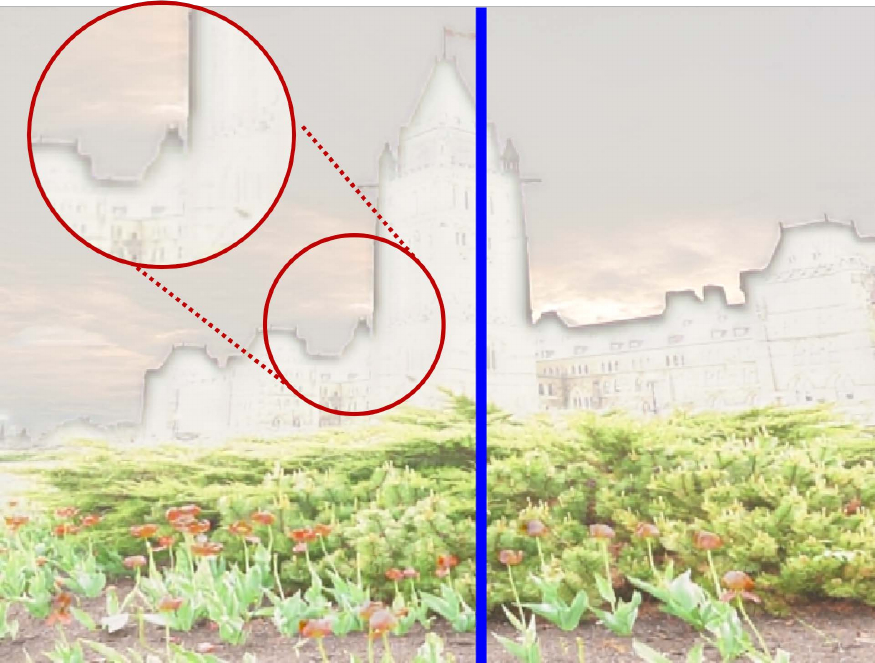}
		&		\includegraphics[width=0.164\textwidth]{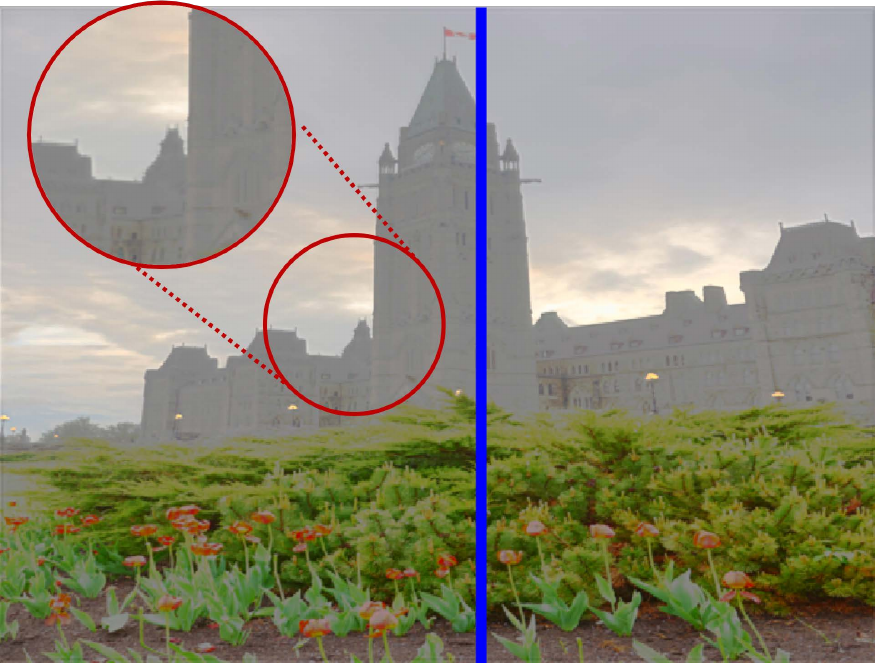}
		&	\includegraphics[width=0.164\textwidth]{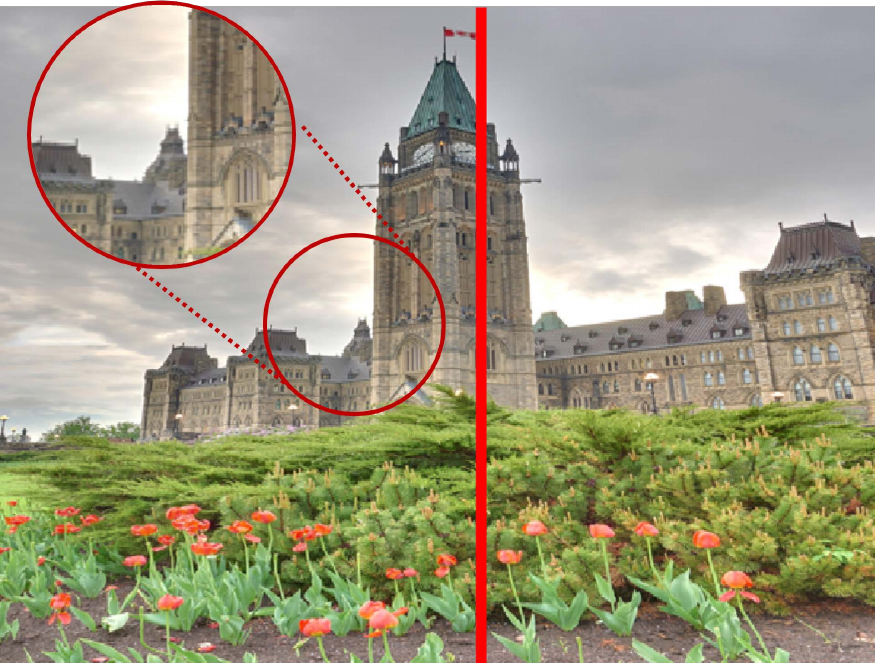}\\
		
		\includegraphics[width=0.164\textwidth]{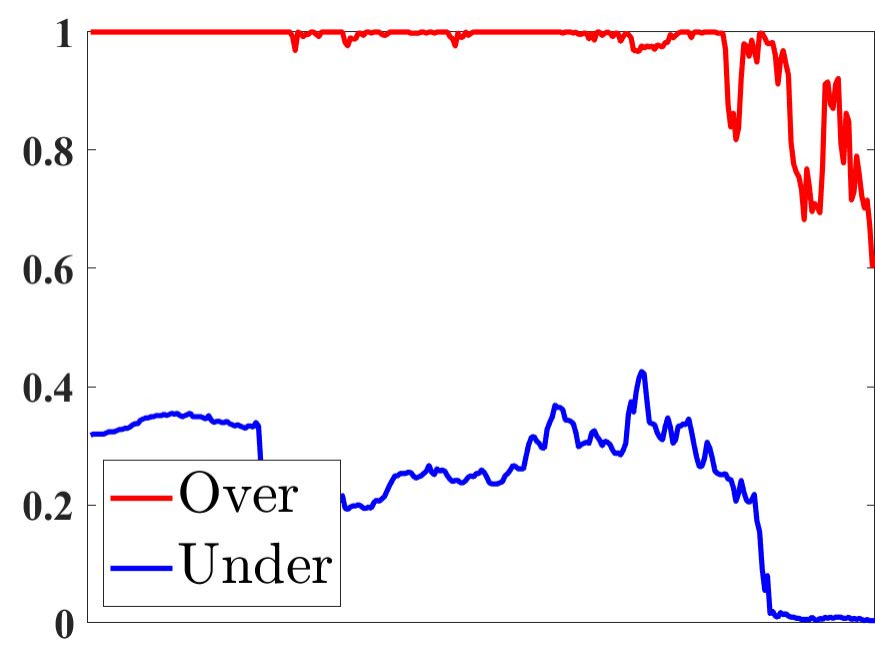}
		&		\includegraphics[width=0.164\textwidth]{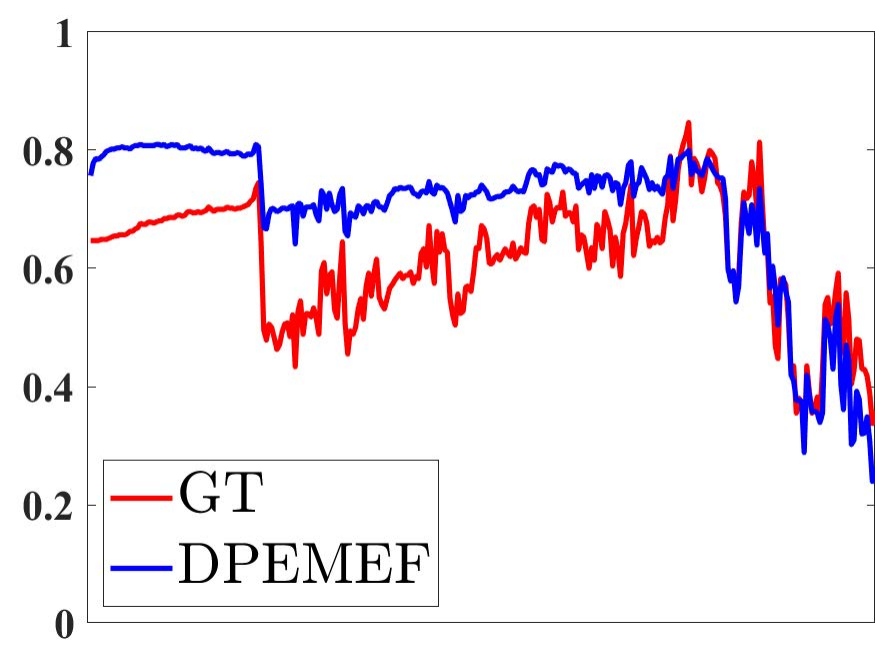}
		&		\includegraphics[width=0.164\textwidth]{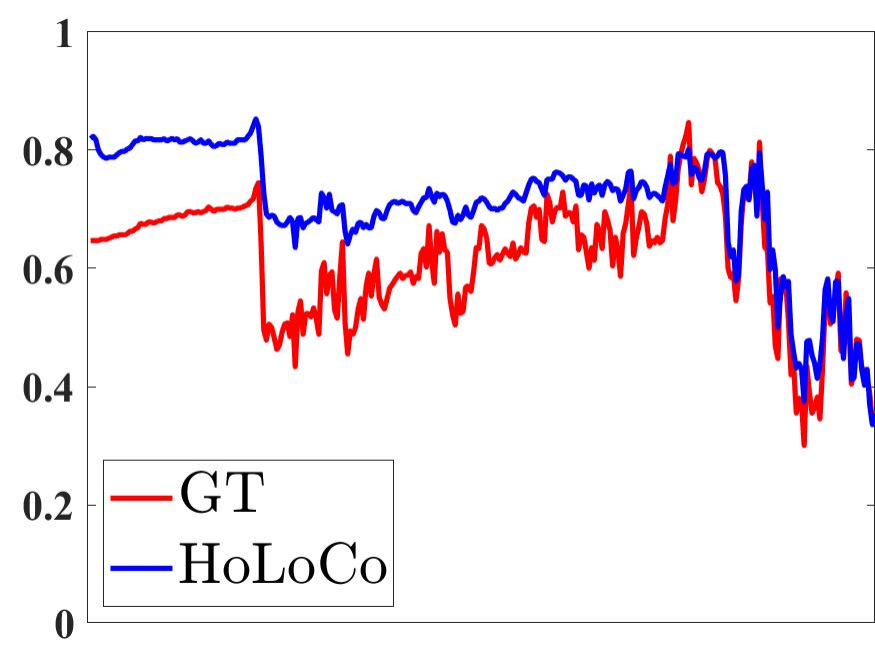}
		&		\includegraphics[width=0.164\textwidth]{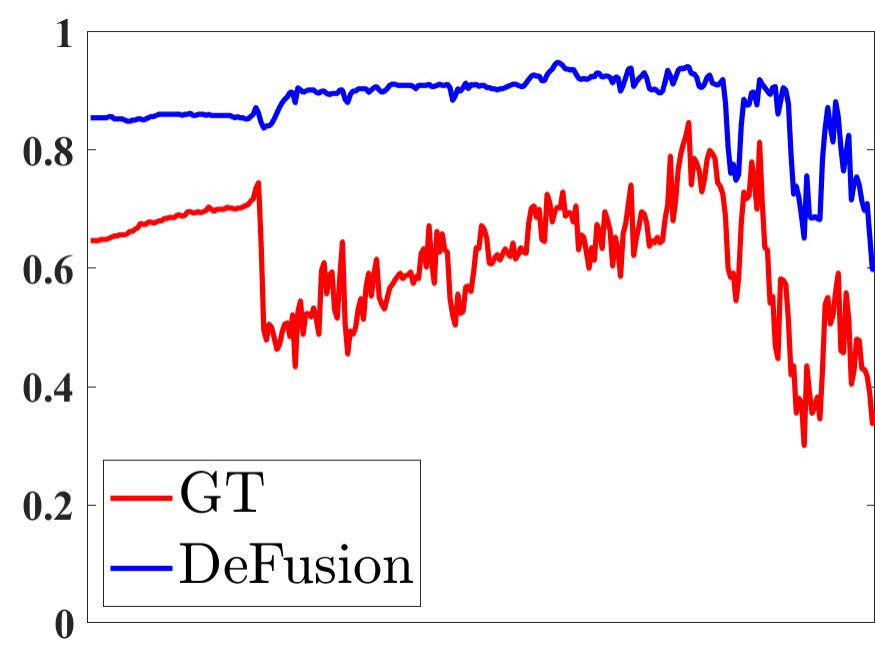}
		&		\includegraphics[width=0.164\textwidth]{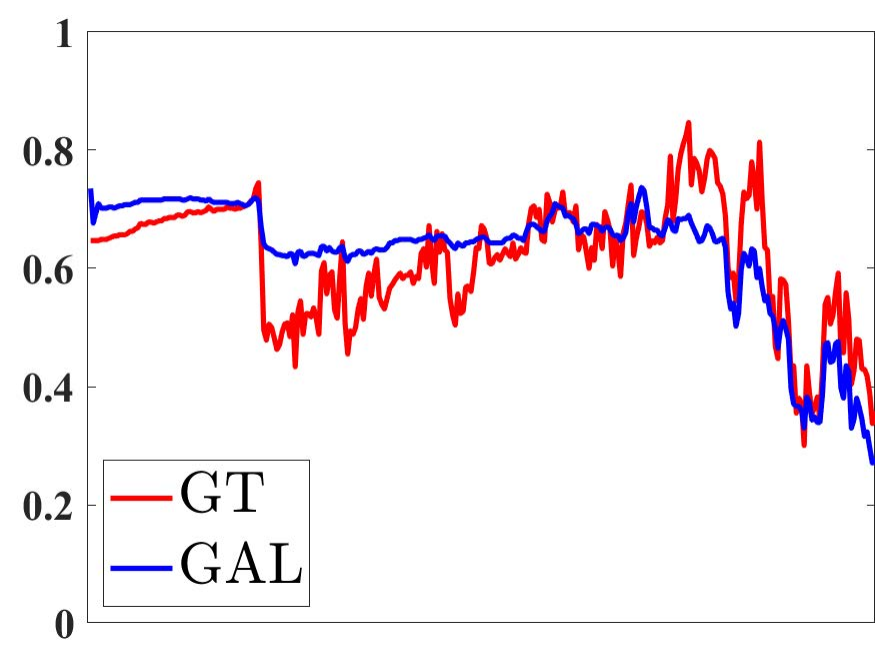}
		&	\includegraphics[width=0.164\textwidth]{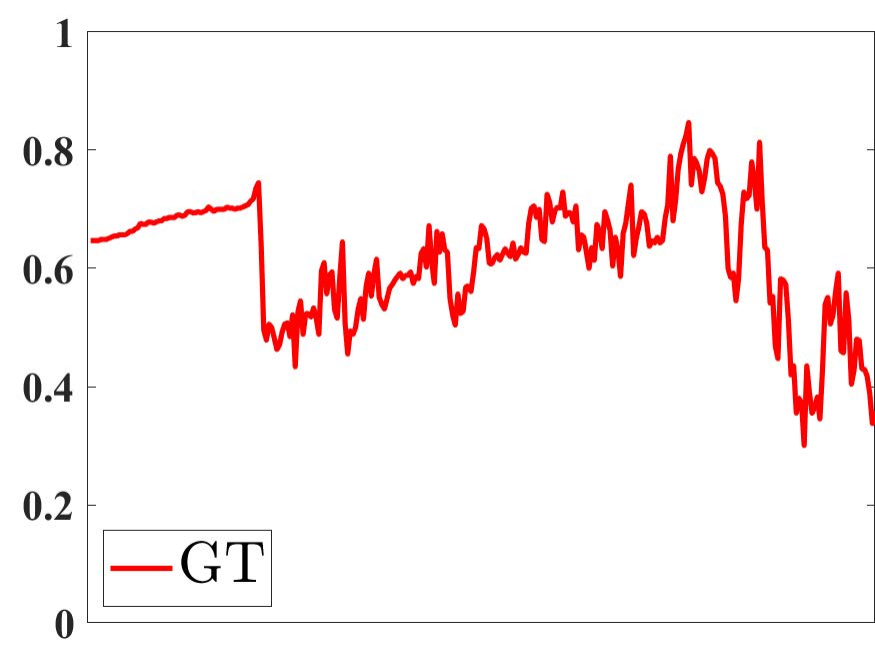}\\
		\footnotesize	Input images & \footnotesize DPEMEF &  \footnotesize HoLoCo &  \footnotesize DeFusion  & \footnotesize GAL & \footnotesize Ground Truth \\
		\includegraphics[width=0.164\textwidth]{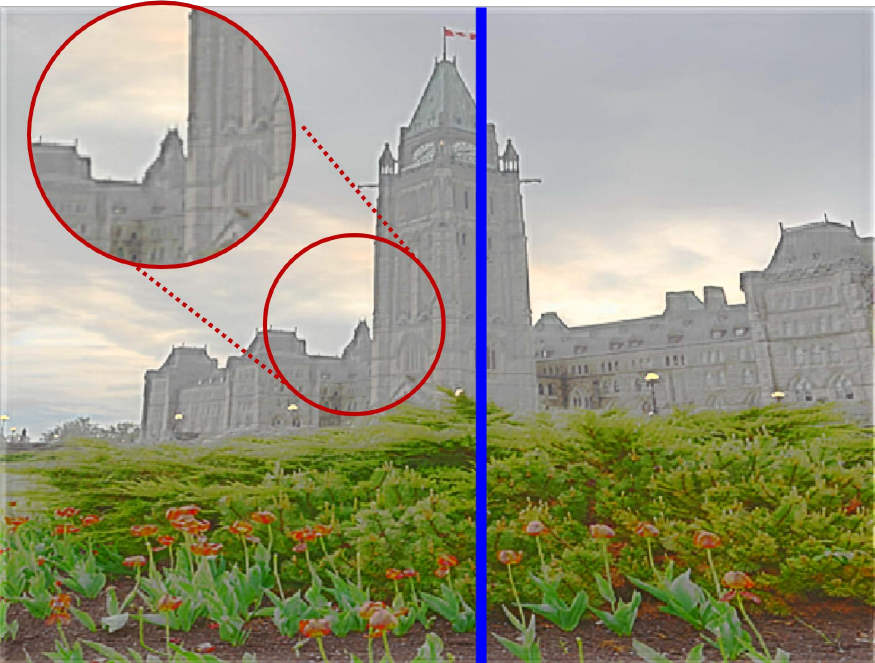}
		&	\includegraphics[width=0.164\textwidth]{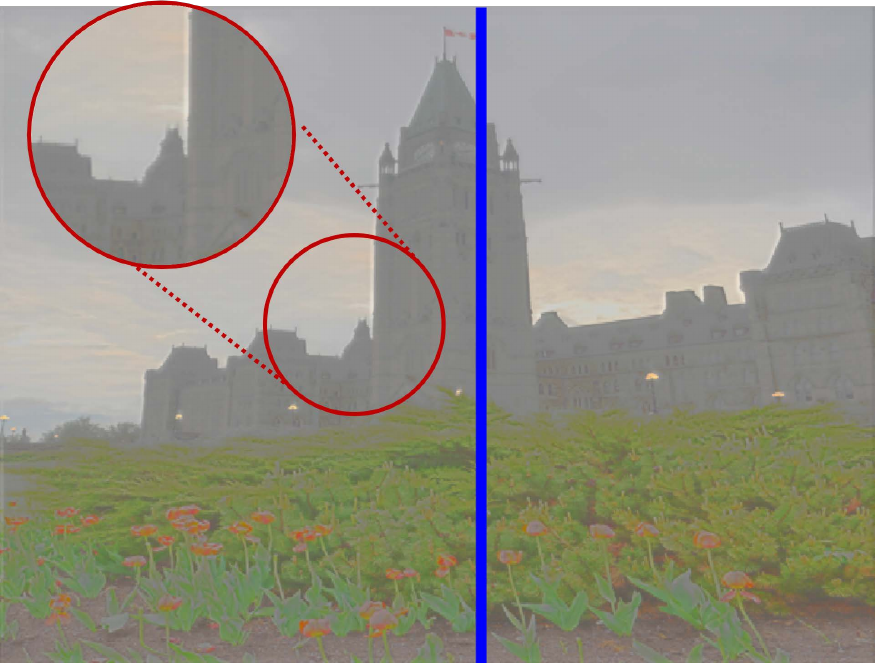}
		&		\includegraphics[width=0.164\textwidth]{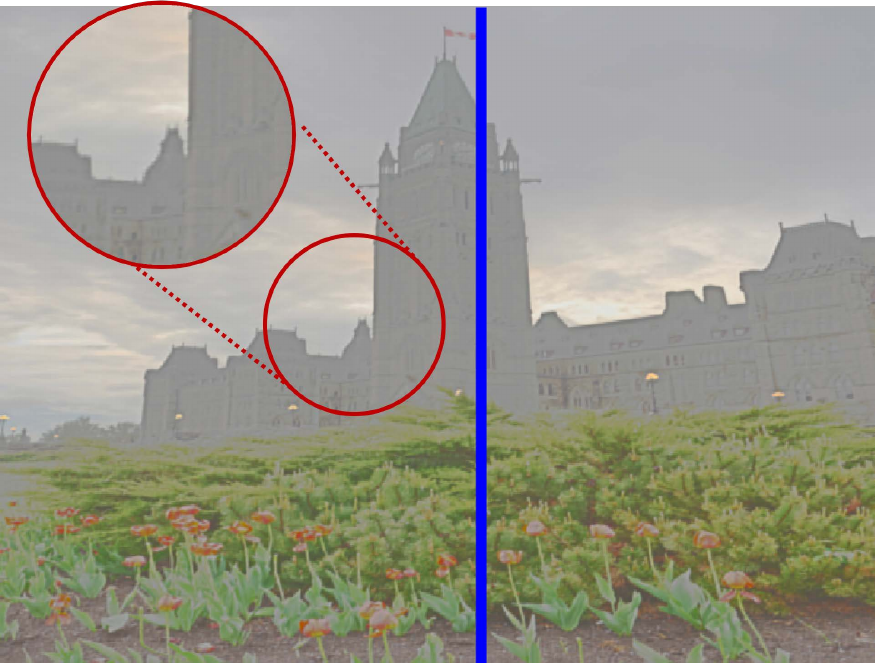}
		
		&		\includegraphics[width=0.164\textwidth]{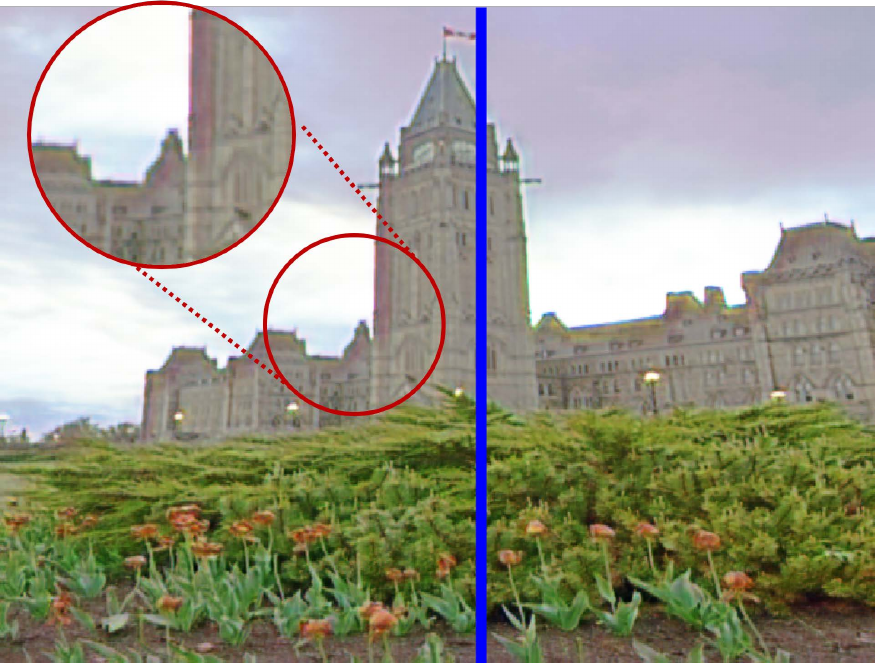}
		&		\includegraphics[width=0.164\textwidth]{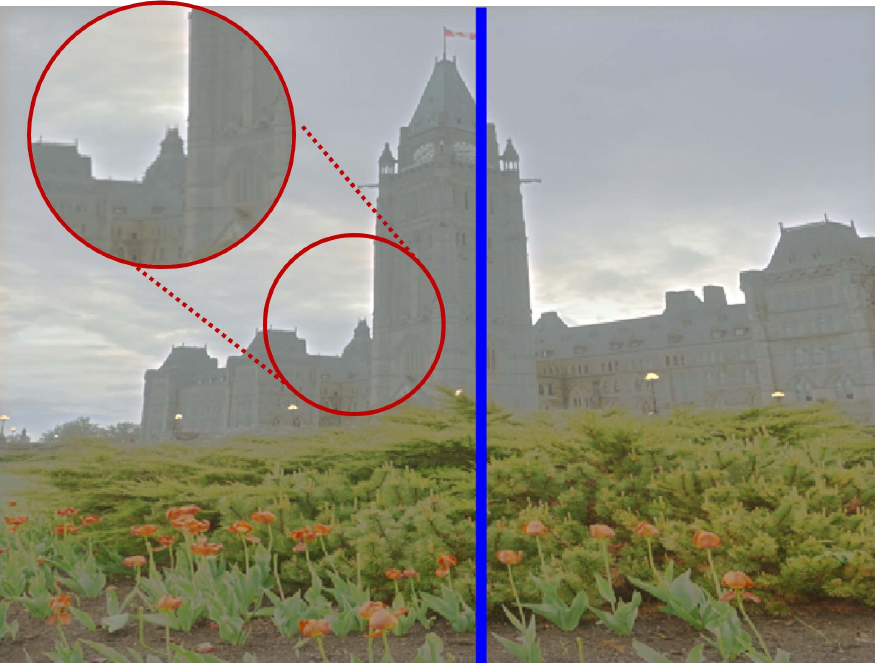}
		&	\includegraphics[width=0.164\textwidth]{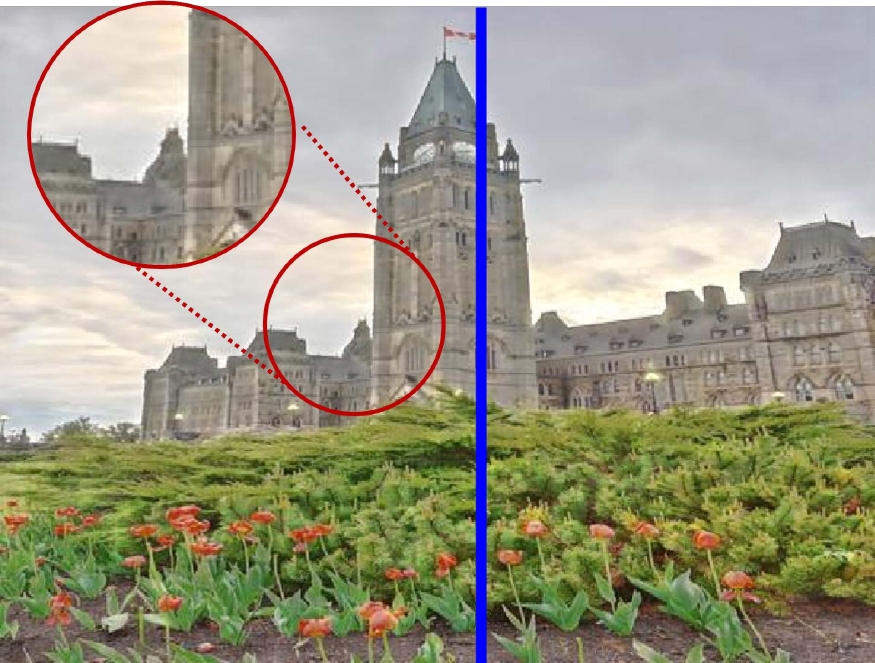}\\
		\includegraphics[width=0.164\textwidth]{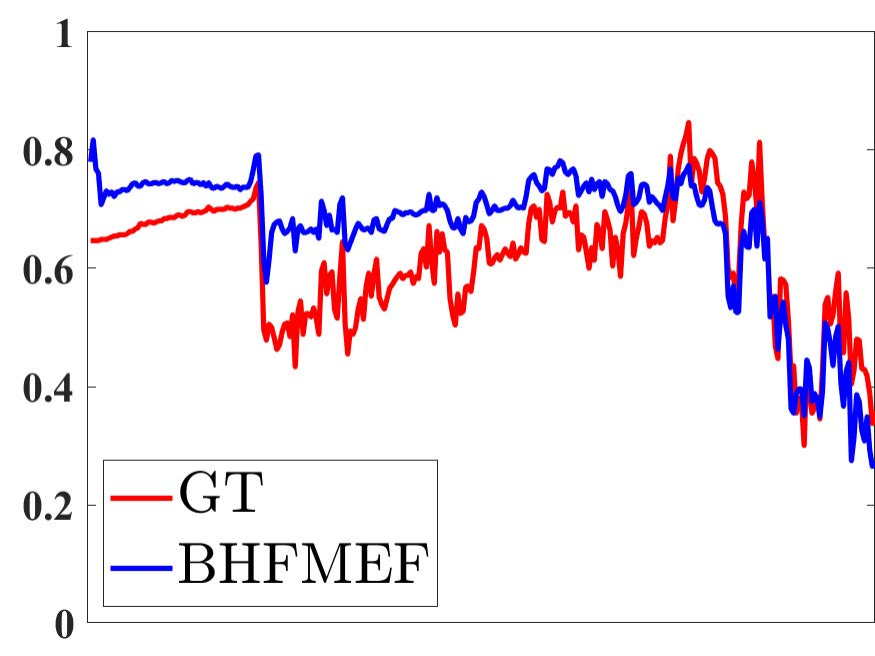}
		&		\includegraphics[width=0.164\textwidth]{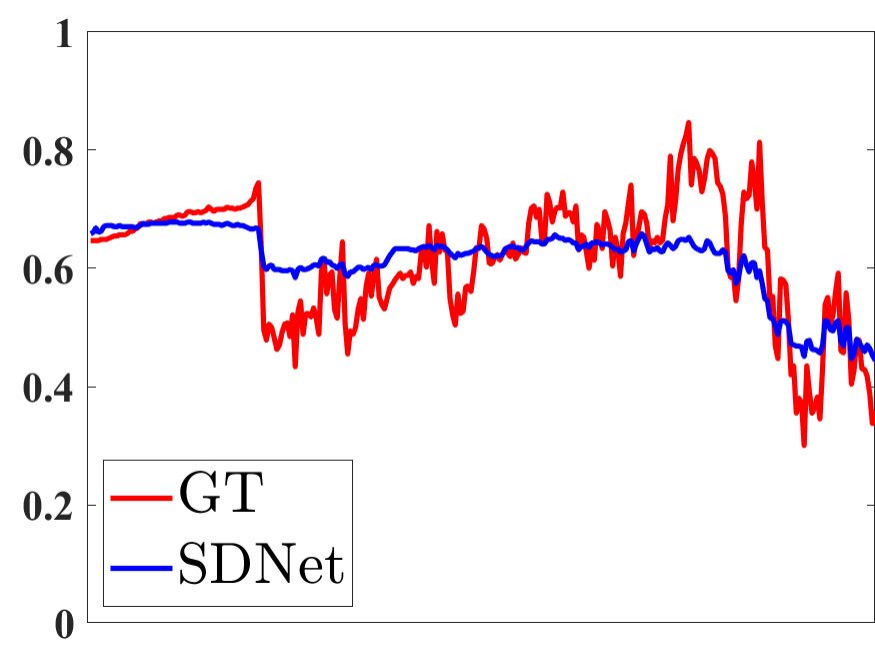}
		&		\includegraphics[width=0.164\textwidth]{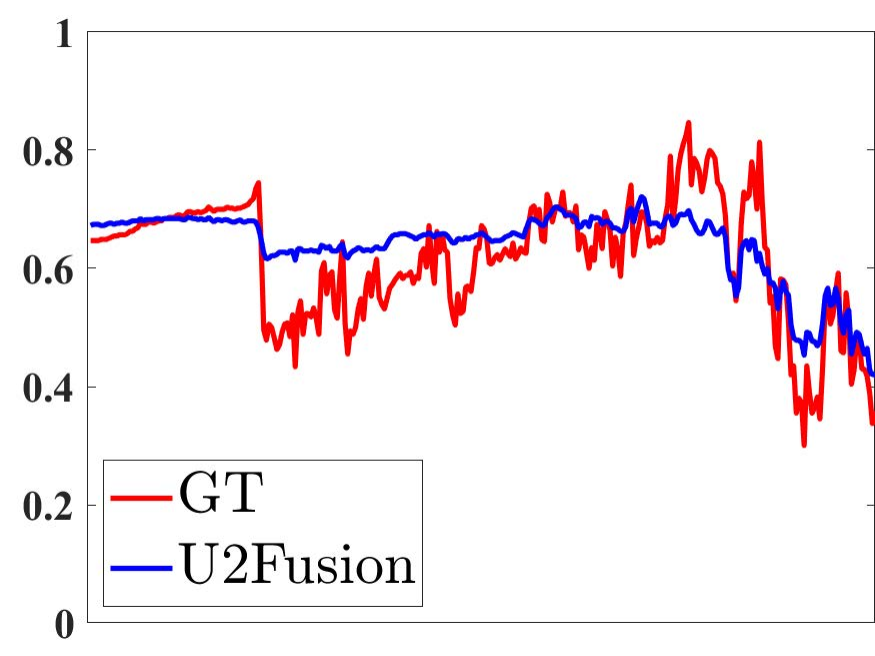}
		&		\includegraphics[width=0.164\textwidth]{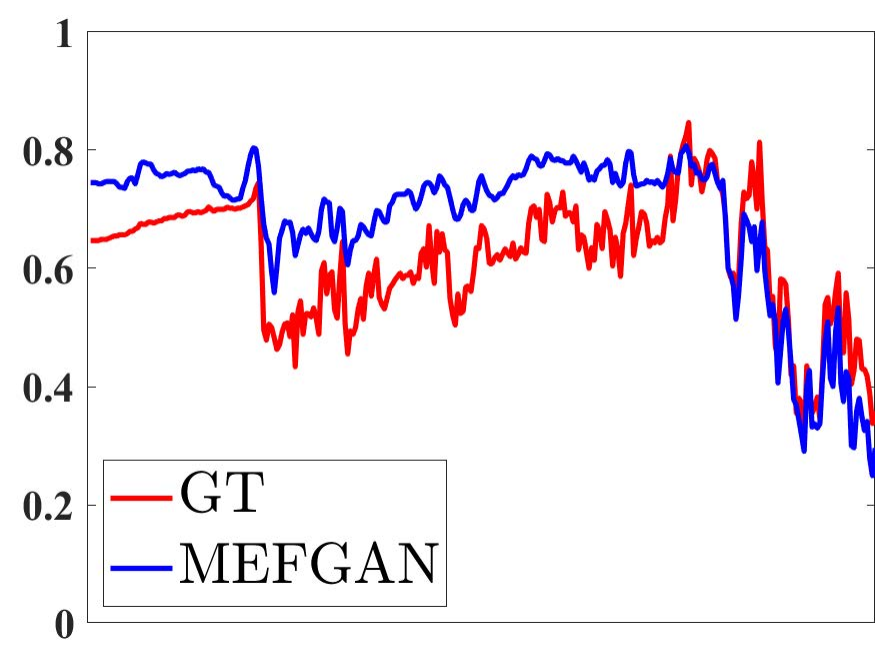}
		&		\includegraphics[width=0.164\textwidth]{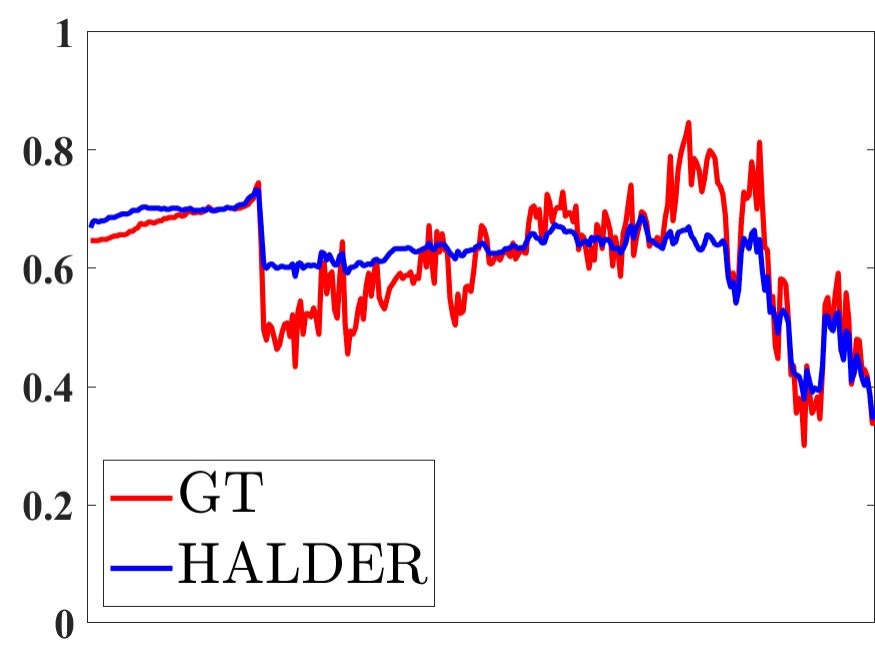}
		&	\includegraphics[width=0.164\textwidth]{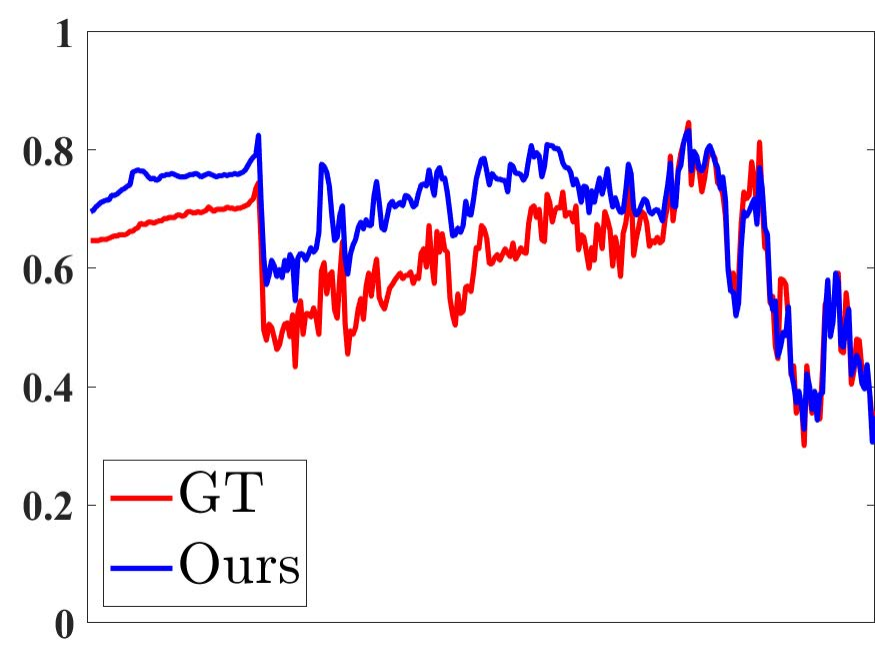}\\
		\footnotesize	 BHFMEF  & \footnotesize SDNet & \footnotesize  U2Fusion & \footnotesize MEFGAN &  \footnotesize HALDER & \footnotesize Ours \\
	\end{tabular}
	\caption{\textcolor{black}{Qualitative comparison with nine state-of-the-art methods. The signal maps provide the differences of pixel intensity with  ground truth.}}
	\label{fig:gmefusion2}
\end{figure*}
\begin{figure*}[!htb]
	\centering
	\setlength{\tabcolsep}{1pt} 
	\begin{tabular}{c}		
		\includegraphics[width=0.98\textwidth]{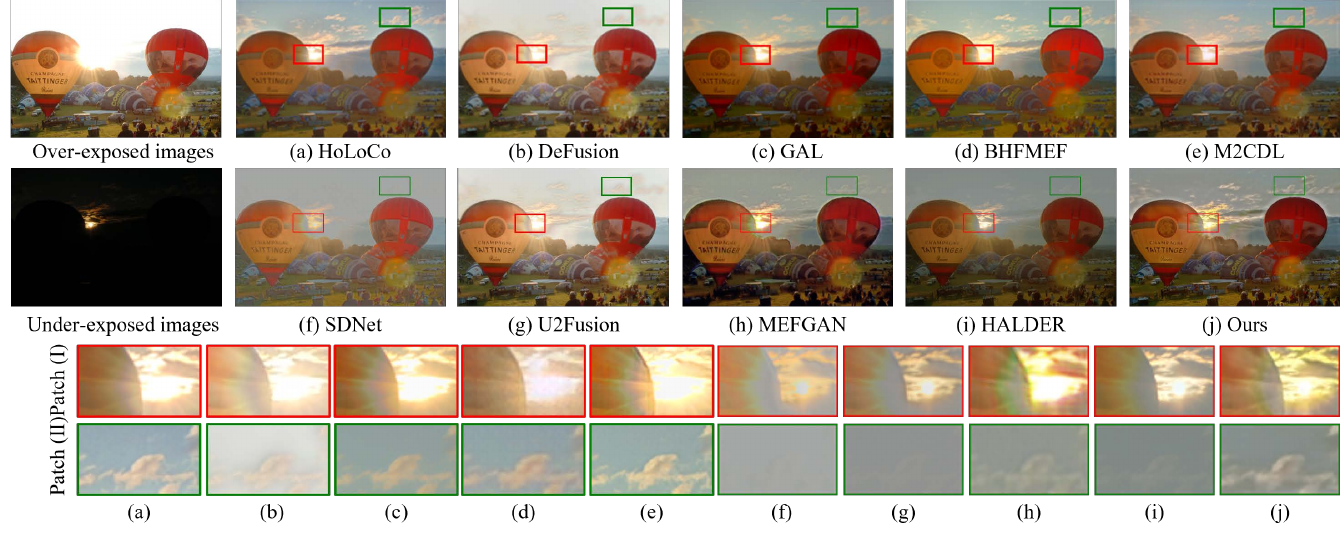}\\
		\includegraphics[width=0.98\textwidth]{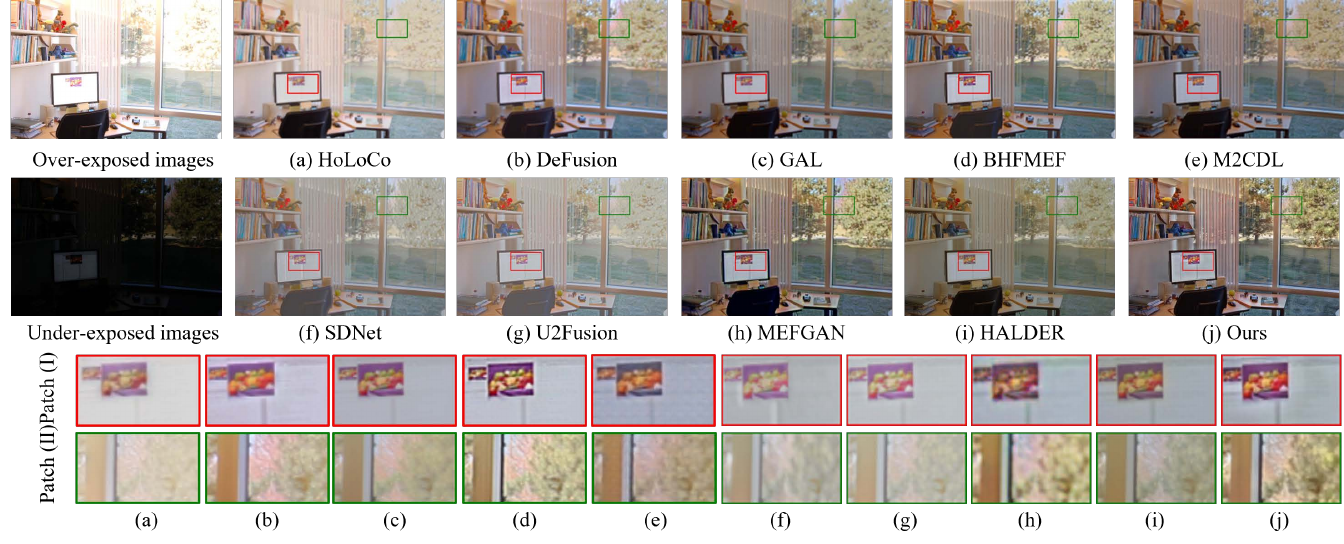}\\
	\end{tabular}
	\vspace{-0.3cm}  
	\caption{\textcolor{black}{Qualitative comparison with nine state-of-the-art methods on the dataset~\cite{ma2017robust} without ground truth.}}
	\label{fig:numcpr}
\end{figure*}

\section{Experimental Results}
In this section, we first introduce the detailed configurations of the architecture search and training procedure. Then we conduct the subjective and objective comparisons on general multi-exposure image fusion and misaligned multi-exposure image fusion with eleven  methods, which demonstrates the remarkable performances and robust generalization ability of the proposed method.
\subsection{Search and Training Configurations}
\subsubsection{Datasets} We used the widely-used SICE dataset~\cite{cai2018learning} to train and evaluate the performance of our network. This dataset contains diverse sequences of scenes with varying exposure ratios. Each sequence has a well-exposed ground truth. For our general multi-exposure image fusion task, we randomly selected 258 pixel-level registered pairs for training and 100 pairs for testing with a significant exposure difference from each sequence. For the misaligned multi-exposure image fusion task, we selected 100 pairs with noticeable unregistered pixels to create a dataset for misaligned scenarios. Additionally, we introduced a dataset~\cite{ma2017robust} without ground truth to verify the generalization ability of the network.

\subsubsection{Architecture Search}
Specifically, in contrast to the original weight-sharing approach, each search block in our method has unique architecture weights, integrating operations from diverse sub-search spaces. \textcolor{black}{We construct a look-up table for each operation $O$, calculating the inference time of operations with diverse feature channels (including 16, 32, and 64) and diverse scales with a batch size of 16. We test each operation 1000 times and leverage the average running time as the latency.} Firstly, the whole super-net is pre-trained with 10 epochs to obtain well-initialized $\bm{\omega}$. Then we conduct the differentiable architecture search with 300 epochs.
SGD optimizer and cosine delay schedule with initial learning rate $3e^{-4}$ are introduced to optimize the neural parameters.  Adam optimizer is introduced to update the architecture with a learning rate $1e^{-4}$.  The $\eta$ at Eq.~\eqref{eq:hs} is empirically set to $0.5$ to balance the performance and inference time (denoted as ``Ours'').
The faster version is also provided based on the constraint $\eta = 1$ and denoted as  ``Ours$^{*}$''.
Moreover, 80 unregistered pairs 
are utilized to search for the specific network for the misaligned scenarios. $\ell_\mathtt{Int}$ is defined as the training and validation loss for architecture search. 
\subsubsection{Network Training} \textcolor{black}{The $\beta_{1}$ and $\beta_{2}$ of Eq.\eqref{eq:totloss} are set to 0.75 and 0.05 respectively, which are selected by grid search.} Data augmentation, such as random crop, horizontal and vertical flipping, and rotating are utilized for the training procedure. With a patch size of $128 \times 128$, we train the network 2000 epochs. This network is trained with Adam optimizer
and introduce the cosine annealing strategy to delay the learning rate from $1e^{-4}$ to $1e^{-10}$ progressively.

We introduce two categories of metrics to measure the visual quality of generated results, including the reference-based measurements (PSNR and SSIM) and visual perception metrics (LPIPS~\cite{zhang2018unreasonable} and FSIM~\cite{zhang2011fsim}). PSNR measures the differences in pixel intensity between outputs and ground truths. SSIM can provide similar measurements from the luminance, contrast, and structure aspects. LPIPS measures perceptual similarity based on deep features aligned with human visual perception, while FSIM evaluates salient low-level features using phase congruency and gradient magnitude.
%% PSNR, SSIM, LPIPS, MEF-SSIM, VIF and FSIM

\subsection{General Multi-Exposure Image Fusion}
\textcolor{black}{To demonstrate the effectiveness and remarkable advantages of the proposed methods, we comprehensively compare our methods with ten competitors including DeFusion~\cite{liang2022fusion}, BHFMEF~\cite{mu2023little}, GAL~\cite{lei2023galfusion}, HoLoCo~\cite{liu2023holoco},  M2CDL~\cite{deng2023deepm2cdl}, SDNet~\cite{zhang2021sdnet}, U2Fusion~\cite{xu2020u2fusion}, MEFGAN~\cite{xu2020mef},  HALDER~\cite{liu2021halder} and DPEMEF~\cite{han2022multi}. }
%In detail, MESPD, FMMEF,  MEFSIFT and DEMEF are representative numerical computation methods with effective ghost removal abilities based on structural patch decomposition and multi-scale fusion.
%DeepFuse and IFCNN are two landmarks for multi-exposure fusion.
% Deepfuse directly merges the luminance information supervised by MEF-SSIM.
%IFCNN utilizes the element-wise summation to integrate the features and reconstruct them by two convolutions. SDNet decomposes this task as pixel intensity and gradient preservation. U2Fusion utilizes the feature-level measurement
%to constrain the similarity. MEF-GAN utilizes the non-local generator and adversarial network for the training.  HALDER proposes a hierarchical attention module and utilizes the refine module to obtain the fused results. DPEMEF leverages two U-Net modules to realize detail enhancement and aesthetic perseveration jointly.
Then we evaluate the proposed scheme with these methods
from three aspects, \textit{i.e.,} the objective, subjective comparison, and computation complexity.
\begin{table*}[htb]
	\renewcommand{\arraystretch}{1.3}
	\caption{\textcolor{black}{Qualitative comparison of proposed methods with a series of multi-exposure fusion schemes.}}
	\label{tab:general_mef}
	\centering
	\setlength{\tabcolsep}{1.5mm}{
		\begin{tabular}{|c| c| c| c| c| c| c| c| c| c| c| c | c|}
			\hline
			Metrics   & DeFusion &BHFMEF & GAL&  HoLoCo  & M2CDL  & SDNet & U2Fusion& MEFGAN   	& HALDER  & DPEMEF & Ours & Ours$^{*}$ \\\hline
			\cellcolor{gray!15}	PSNR $\uparrow$ &12.73                                                        & 19.47                                                 & 19.32                                                  & 20.09                                                    & 18.82                                                    & 17.42& 17.67  & 19.71&19.91& 19.23&\textcolor{red}{\textbf{20.71}}$_{\uparrow 3.09\%}$& \textcolor{blue}{\textbf{20.54}}\\\hline
			\cellcolor{gray!15}	SSIM $\uparrow$& 0.689                                                        & 0.802                                                 & 0.770                                                  & 0.819                                                    & 0.741                                                         & 0.753 &0.718 & 0.757 & 0.763&\textcolor{red}{\textbf{0.844}}&\textcolor{blue}{\textbf{0.825}}&0.822\\\hline
			\cellcolor{gray!15}	LPIPS $\downarrow$&0.299                                                        & 0.194                                                 & 0.207                                                  & 0.167                                                    & 0.242                                                      &0.248&0.224&0.273&0.175&0.143&\textcolor{red}{\textbf{0.132}}$_{\downarrow 7.69\%}$&\textcolor{blue}{\textbf{0.138}}\\\hline
			\cellcolor{gray!15}	FSIM $\uparrow$&0.837                                                        & 0.887                                                 & 0.876                                                  & 0.923                                                    & 0.876                                                    &0.829&0.853&0.906&0.921&0.886&\textcolor{red}{\textbf{0.924}}$_{\uparrow 0.11\%}$&\textcolor{red}{\textbf{0.924}}\\\hline
			%			MEF-SSIM & 0.859&0.823&0.849&0.862&0.856&0.849&0.862&0.851&0.870&0.870&0.871&{0.872}&\textbf{0.880}\\
			%			\hline
			
		\end{tabular}	
	}
\end{table*}
\begin{figure*}[htb]
	\centering
	\includegraphics[width=0.98\textwidth]{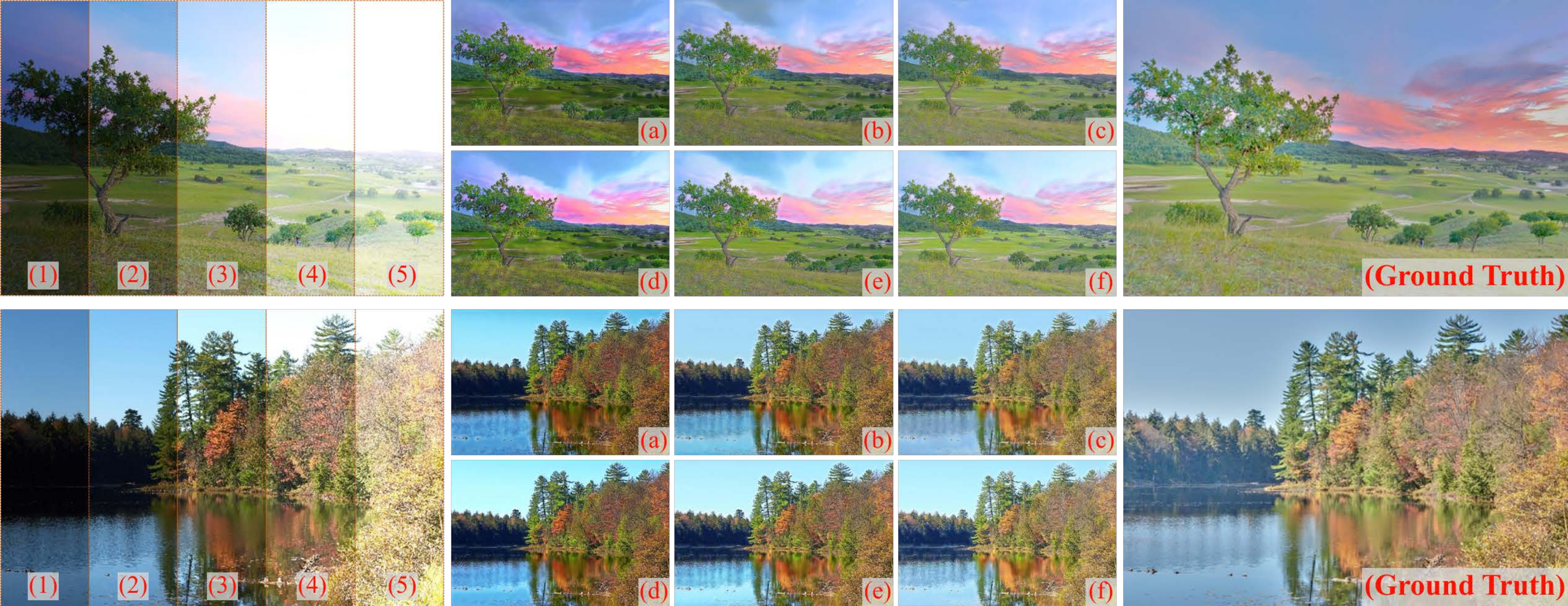}
	\caption{{Visual results about the source image sequences with different exposure ratios. (1) and (2) are under-exposed images, (3), (4) and (5) are over-exposed images.  (a)-(c) are fused by inputs of (1) and (3)-(5).  (d)-(f) are fused by  inputs of (2) and (3)-(5).}}
	
	\label{fig:multiseq}
\end{figure*}
\subsubsection{Subjective Comparison} We select two representative pairs to demonstrate the superiority of our proposed method, which are shown at Fig.~\ref{fig:gmefusion} and Fig.~\ref{fig:gmefusion2}. In the below part of each image, we also show signal maps related to the marked line in blue, compared with the ground truth, to highlight the noticeable differences.
%Since these image pairs have a challenging distinct exposure gap, it is untoward to design the multi-exposure fusion scheme to preserve the suitable brightness, structural details, and abundant color distribution. From these noticeable comparisons, we can conclude our methods contain three significant advantages. 

\textcolor{black}{Firstly, our scheme effectively handles the highly-extreme area and accurate pixel intensity distribution, which are without the information distortion. DeFusion and GAL cannot realize the pixel consistency with the ground truth, shown at the corresponding signal maps.
	Meanwhile, the sky is degraded by extreme-low exposure time, as shown in the local under-exposed regions. In contrast, our method can recover the promising brightness with normal illumination. Secondly, current learning-based schemes are easily trended to color distortion. For instance, SDNet, U2Fusion, and HALDER methods cannot realize the vivid color details, including the bushes in the first scene and the flowers in the second scene.  DPEMEF and BHFMEF cannot maintain the textural details, affected by the strong illumination of over-exposure images. This illustration is also reflected in the corresponding signal maps. These methods cannot achieve large signal changes and are with a moderate reflection of pixel intensity. Our method and MEFGAN
	can preserve the promising color distribution with remarkable improvement.
	Our results are visual-friendly, which is an incline with the human vision system.}

On the other hand, we also provide another comparison based on the dataset~\cite{ma2017robust} without ground truth in Fig.~\ref{fig:numcpr}. We select two pairs with extreme exposure variance to illustrate the effectiveness of our scheme with nine state-of-the-art multi-exposed image fusions. As shown in the first sequence, the information (\textit{e.g.,} cloud layer) under low exposure cannot be recovered clearly.  \textcolor{black}{Thus, these details are hard to highlight and recover from under-exposed images (\textit{e.g.,}  DeFusion, SDNet, and U2Fusion marked by the green boxes). We can clearly observe that attention-based methods (HoLoCo and Ours) can achieve abundant detail preservation.} Especially, our network can effectively promote visual perception to render sufficient details. Furthermore, our method can accomplish vivid color enhancement. Most of the results appear in the local over-exposure region. In contrast, our method achieves abundant texture details (\textit{e.g.,} bushes) and consistent color distribution (\textit{e.g.,} computer screen). Most learning-based schemes cannot address the global consistency of illumination.
\subsubsection{Objective Comparison}
To demonstrate the superiority of the proposed scheme, we utilize four different metrics, including PSNR, SSIM, LPIPS, and FSIM to measure the visual quality of diverse methods. The whole numerical results are reported in Table.~\ref{tab:general_mef}. We introduce two versions to conduct the comparison, which are named ``Ours'' and ``Ours$^{*}$'' respectively. The difference between both versions is utilizing diverse latency constraints, where ``Ours$^{*}$'' focuses more on the inference time.
Our scheme achieves consistent performance improvement in terms of these metrics. \textcolor{black}{Compared with 
	representative supervised learning-based schemes M2CDL and HoLoCo, our scheme promotes 1.89 dB and 0.62 dB drastically. On the other hand, it can be clearly seen that we obtain the second-best numerical results. However, the patch-based fusion scheme obtains the suboptimal numerical performance, which indicates our scheme can effectively preserve sufficient textural details and structural information.  We also utilize the LPIPS to measure the distortion at feature levels. Our scheme can reduce almost 45.5\% of LPIPS compared with M2CDL, which demonstrates better visual quality in line with the human perception system.  Obviously, our scheme also achieves the best results.}

\begin{figure*}[thb]
	\centering \begin{tabular}{c} 
		\includegraphics[width=0.99\textwidth]{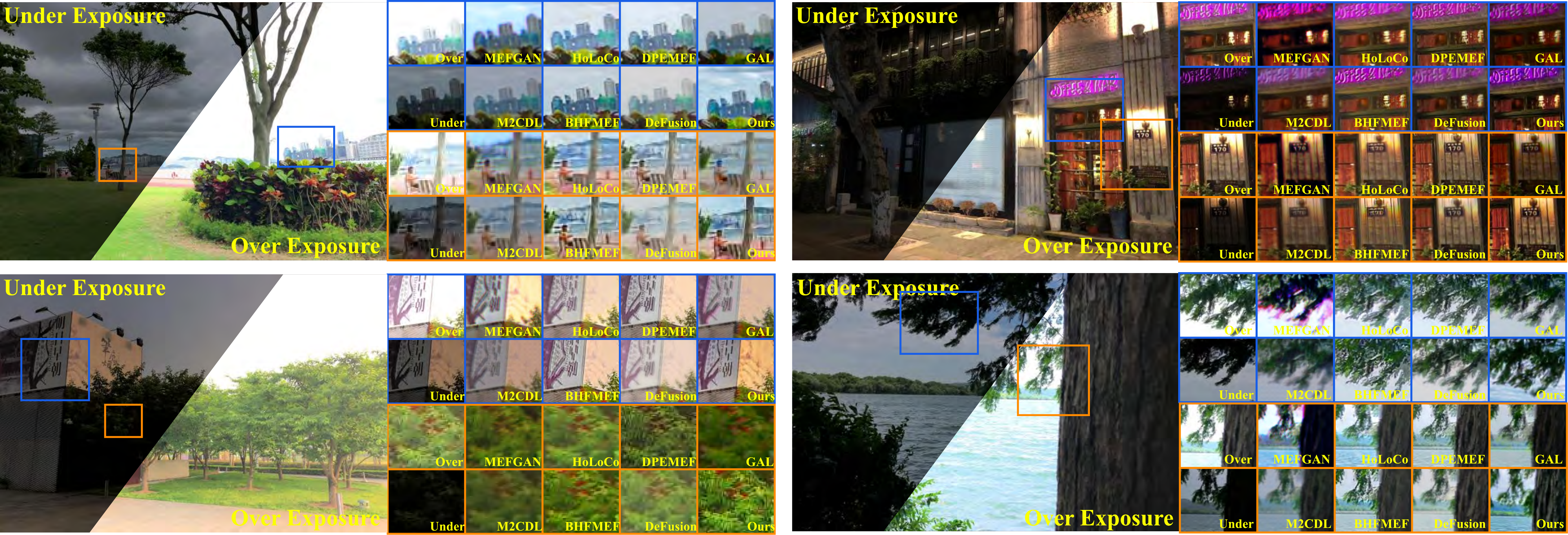} \\
	\end{tabular}
	\caption{\textcolor{black}{Qualitative results compared with the state-of-the-arts on the misaligned multi-exposure fusion.  Other methods are aligned by AGFlow.} }
	\label{fig:misalignment_MEF}
\end{figure*}
%\begin{table*}[!htb]
%	\centering
%	\caption{Trade-off $\eta$ for hardware-sensitive analysis.}
%	\renewcommand\arraystretch{1.3} \footnotesize
%	\setlength{\tabcolsep}{3.5mm}
%\begin{tabular}{|c|c|c|c|c|c|c|c|c|c|}
%	\hline
%Metrics	& IFCNN  &  DeepFuse & SDNet & U2Fusion  & MEFGAN  & HALDER  & DPEMEF  & Ours  & Ours$^{*}$ \\ \hline
%Parameters (M)	&  &  &  &  &3.157  & 4.711 & 51.93 & 1.879 & 0.684 \\ \hline
%%FLOPs (G)	&  &  &  &  &  &  &  &  &  \\ \hline
%%Runtime(s)	&  &  &  &  &  &  &  &  &  \\ \hline
%\end{tabular}
%	\label{tab:flops}	
%\end{table*}
%\begin{table}[htb]
%	\centering
%	\renewcommand{\arraystretch}{1.3}
%	\caption{Quantitative comparisons of effectiveness with SRSM.}
%	\label{tab:nogt}
%	\setlength{\tabcolsep}{3mm}{
	%		\begin{tabular}{|c| c |c| c| c|}
		%			\hline
		%			Methods & NIQE   & Cascade-1 & SF& AG  \\\hline
		%			
		%			DEMEF&0.611&0.819&\textbf{0.825}&\textbf{0.825}\\\hline
		%			FMMEF&{0.407}&0.150&\textbf{0.132}&{0.136}\\\hline
		%			MEFDSIFT&{0.787}&\textbf{0.932}&0.924&{0.929}\\\hline
		%			MEFSPD&{0.787}&\textbf{0.932}&0.924&{0.929}\\\hline
		%			IFCNN&{0.787}&\textbf{0.932}&0.924&{0.929}\\\hline
		%			HALDER&{0.787}&\textbf{0.932}&0.924&{0.929}\\\hline
		%			SDNet & 9.57&20.15&\textbf{20.71}&{20.56}\\
		%			\hline	
		%			U2Fusion & 9.57&20.15&\textbf{20.71}&{20.56}\\
		%			\hline
		%			Ours&{0.787}&\textbf{0.932}&0.924&{0.929}\\
		%			\hline
		%		\end{tabular}	
	%	}
%	
%\end{table}
\subsubsection{Computation Efficiency Analyses} We also conduct a comparison under computation efficiency, which is a critical point for real-world deployment. The concert numerical results among these competitors under the metrics of parameters and runtime on the SICE dataset are reported in Table.~\ref{tab:hardware}. \textcolor{black}{Obviously, though precise visual results are obtained, these methods  suffer from the slow inference time due to the huge model parameters. More importantly, both our methods realize the faster inference time.   Compared with the latest HoLoCo scheme, the fastest version (ours$^{*}$)significantly reduces  99.3\% parameters and accelerates 79.4\% of inference time., which demonstrates high efficiency with visual-pleasant fused results.}
\begin{table*}[htb]
	\renewcommand{\arraystretch}{1.3}
	\caption{\textcolor{black}{Computation efficiency comparison including parameters and averaged runtime on the SICE dataset.}}
	\label{tab:hardware}
	\centering
	\setlength{\tabcolsep}{0.8mm}{
		\begin{tabular}{|c| c| c| c| c| c| c| c| c| c| c| c | c|}
			\hline
			Metrics & DeFusion &BHFMEF & GAL  &  HoLoCo  & M2CDL  & SDNet & U2Fusion & MEFGAN 	& HALDER   & DPEMEF   & Ours & Ours$^{*}$ \\\hline
			Platform &  Pytorch                                                        & Pytorch                                                 & Pytorch                                                 & Pytorch                                                  & Pytorch                                                    & Tensorflow & Tensorflow  &Tensorflow  & Pytorch &Pytorch&Pytorch&Pytorch\\\hline
			Device& 	GPU & GPU & GPU & GPU & GPU& GPU &GPU & GPU & GPU&GPU&GPU&GPU\\\hline
			\cellcolor{gray!15}	Parameters (M) $\downarrow$ & 7.874                                                  & 1.001                                                & 1.597                                                  & 17.39                                                   & 425.1                                                   &\textcolor{red}{\textbf{0.067}}& \textcolor{blue}{\textbf{0.659}}  &3.157 & 4.711 & 51.93 &1.879&{0.684}\\\hline
			\cellcolor{gray!15}	Runtime (S) $\downarrow$ & 0.188                                                        & 0.769                                                 & 0.082                                                  &0.102                                                & 3.771                                         & 0.885& 0.305 &0.862&0.141& 0.068&\textcolor{blue}{\textbf{0.047}}&\textcolor{red}{\textbf{0.021}}\\\hline
			
		\end{tabular}	
	}
\end{table*}
\subsubsection{Fusion with Arbitrary Exposure Ratios}
In order to verify the generalization ability to address the inputs with arbitrary exposure ratios, we select two  representative sequences,  which is shown in Fig.~\ref{fig:multiseq}. Since we utilize the pairs with the largest exposure ratios for training and testing, the proposed scheme is robust enough to handle different exposure ratios.  These sequences contain two under-exposed images and three over-exposed images. We can clearly observe that the scheme can obtain the consistent visual-pleasant fused results with natural color correction, generated by the image pairs with diverse exposure ranges. Several characteristics can be found from the fused result of this sequence. Firstly, the proposed scheme is with large capacity for wide exposure difference to preserve the textural details and maintain
suitable illumination distribution. Secondly, though without the training procedure of small exposure difference, the fused images generated by (a) and (b) contain sufficient scene details, \textit{e.g.,} grasses and sunset glow. Finally,  our result (\textit{i.e.,} (c)) fused by the large difference of exposure is close to the ground
truth. Several fused images (\textit{e.g.,} (b) and (d)) have more vivid scene representation compared with ground truth.

\subsection{Misaligned Multi-Exposure Image Fusion}
Misaligned multi-exposure image fusion is a challenging scenario due to the camera movement and device shaking, whereas current methods for MEF are easy to generate blurs without the consideration of pixel registration. We also illustrate the performance by numerical and visual comparisons.

\subsubsection{Visual Comparison}
We select four misaligned pairs to demonstrate the effectiveness of the proposed framework in Fig.~\ref{fig:misalignment_MEF}. \textcolor{black}{We leverage the AGFlow to previously align these methods for the comparison. The last two rows illustrate the scene with large pixel movements.  Most methods cannot preserve sufficient details, generated with obvious artifacts. M2CDL and DeFusion cannot recover accurate illuminations with blurred scenes. More importantly, our scheme successfully realizes the uniform promotion of pixel alignment and visual enhancement, which can effectively address diverse levels of pixel alignment with abundant details (\textit{e.g.,} buildings and sign boards at the first row).
}

%The first two rows illustrate the scene with large pixel movements.  Most methods cannot preserve sufficient details. The results generated by  MEF-GAN and FMMEF methods have obvious artifacts and cannot preserve the normal color distribution. More importantly, our scheme successfully realizes the uniform promotion of pixel alignment and visual enhancement, which can effectively address diverse levels of pixel alignment.
\subsubsection{Numerical Comparison} \textcolor{black}{In Table.~\ref{tab:misalign}, we report the numerical results compared with various representative multi-exposure image fusion methods. We also utilize the remarkable optical flow techniques (including RAFT~\cite{teed2020raft}, SKflow~\cite{sun2022skflow} and AGflow~\cite{luo2022learning})  to align existing MEF methods for a  fair comparison. Moreover, we also utilize five typical metrics to measure the 
	visual quality of fused images.  Since existing methods often assume the image pairs are well-registered, these methods cannot obtain promising quantitative results.  We can clearly observe that though advanced alignment is leveraged for these MEF methods, our method still performs the remarkable performance among all three metrics. Compared with M2CDL, our method improved 12.8\% of PSNR, 47.7\% of SSIM, and reduced 45.5\%  error of LPIPS.}

%\begin{table*}[htb]
%	\renewcommand{\arraystretch}{1.3}
%	\caption{Numerical results with representative methods on \textit{TNO} and \textit{RoadScene} datasets.}
%	\label{tab:align_reference}
%	\centering
%	\setlength{\tabcolsep}{1.2mm}{
	%		\begin{tabular}{|c c c c c c c c c c c c c c |}
		%			\hline
		%			Metrics  & CFNet & AGAL & CUNet& SDNet & U2Fusion & MESPD & MEFGAN &  DEMEF &FMMEF & MEFCNN  & MEFDSIFT	& HALDER & Ours \\\hline
		%
		%			NIQE& 
		%			20.31& 19.29& 15.11& 17.42& 17.67 &12.53 & 19.71 & 13.24& 14.50& 12.17&15.10&19.91&\textbf{20.71}\\
		%			BRISQUE& 
		%			0.872& 0.856& 0.524& 0.718& 0.753 &0.718 & 0.757 & 0.745& 0.763& 0.726&0.766&0.808&{0.825}\\
		%			DE& 0.393&0.471&0.374&0.434&0.464&0.423&0.464&0.435&0.454&0.388&0.436&0.495&\textbf{0.681}\\
		%		
		%			ARISMC& 10.09&14.25&16.10&16.22&2.271&17.04&16.34&14.79&15.11&13.73&13.40&16.71&22.04\\
		%			\hline
		%			
		%		\end{tabular}	
	%	}
%\end{table*}
\begin{table*}[htb]
	
	\renewcommand{\arraystretch}{1.1}
	\caption{\textcolor{black}{Numerical results compared with representative methods for misaligned multi-exposure image fusion.}}
	\label{tab:misalign}
	\centering
	\setlength{\tabcolsep}{1.5mm}{
		\begin{tabular}{|c|c|c|c|c|c|c|c|c|c|c|c|c|}
			\hline
			Alignment                 & Metrics & U2Fusion & SDNet & MEFGAN & HoLoCo & HALDER & DPEMEF & GAL & M2CDL & BHFMEF & DeFusion & Ours \\ \hline
			\multirow{3}{*}{AGFlow}   & 	\cellcolor{gray!15} PSNR $\uparrow$    &     \textcolor{blue}{\textbf{17.46}}    &    16.57   &    15.98    & 14.98       & 16.21       &  14.63      & 16.84    &  17.37     &  16.05      &    13.07      & \textcolor{red}{\textbf{22.04}}  $_{\uparrow 26.2\%}$    \\ \cline{2-13} 
			& 	\cellcolor{gray!15}SSIM $\uparrow$    &       0.457   &   0.432    &    0.438    &    0.439    &     \textcolor{blue}{\textbf{0.462 }}  &     0.416   & 0.451    &   0.452    &     0.421   &  0.408        & \textcolor{red}{\textbf{0.681}} $_{\uparrow 47.4\%}$    \\ \cline{2-13} 
			& 	\cellcolor{gray!15}LPIPS $\downarrow$   &       0.341   &    0.327   &     0.410   &       0.311 &     0.305   &   \textcolor{blue}{\textbf{ 0.288 }}   &   0.327  & 0.376      &     0.306   &   0.380       & \textcolor{red}{\textbf{0.187}}   $_{\downarrow 35.1\%}$    \\ \hline
			\multirow{3}{*}{RAFT} & 	\cellcolor{gray!15}PSNR $\uparrow$    &  \textcolor{blue}{\textbf{17.84}}        & 16.98      & 16.68       & 15.34       & 16.68       &  14.90      &  17.26   &    17.77   &    16.64    &    16.64      &  \textcolor{red}{\textbf{22.04}}    $_{\uparrow 23.5\%}$   \\ \cline{2-13} 
			&	\cellcolor{gray!15} SSIM  $\uparrow$   &    0.463      & 0.432      &   0.480     &   0.456     &    \textcolor{blue}{\textbf{0.480}}   &     0.416   &  0.456   &   0.451    & 0.445       &     0.445     &    \textcolor{red}{\textbf{0.681}}  $_{\uparrow 41.9\%}$ \\ \cline{2-13} 
			& 	\cellcolor{gray!15}LPIPS  $\downarrow$  &       0.358   &   0.327    &  0.317      &      0.318  &      0.317  &    \textcolor{blue}{\textbf{ 0.281}}  &   0.318  & 0.420      &   0.306     &     0.306     &  \textcolor{red}{\textbf{0.187}}  $_{\downarrow 33.5\%}$    \\ \hline
			\multirow{3}{*}{SKFLow} & 	\cellcolor{gray!15} PSNR $\uparrow$    & 17.36         & 16.47      & 16.13       &  14.92      &  16.13      &   14.56     &  16.76   &  \textcolor{blue}{\textbf{ 17.77 }}   &     15.98   &    15.99      &   \textcolor{red}{\textbf{22.04}}     $_{\uparrow 24.0\%}$    \\ \cline{2-13} 
			& 	\cellcolor{gray!15}SSIM  $\uparrow$   &     0.456     &    0.432   &  0.437      &    0.439    &   \textcolor{blue}{\textbf{0.462}}     &     0.417   & 0.450    &  0.461     &    0.421    &       0.421   &  \textcolor{red}{\textbf{0.681}}  $_{\uparrow 47.4\%}$   \\ \cline{2-13} 
			& 	\cellcolor{gray!15} LPIPS $\downarrow$   &       0.343   &     0.329  &   0.412     &    0.311    &    0.306    &   \textcolor{blue}{\textbf{0.291}}     & 0.328    &   0.420    &    0.308    &   0.308       &   \textcolor{red}{\textbf{0.187}}  $_{\downarrow 35.7\%}$   \\ \hline
	\end{tabular}}
\end{table*}

\begin{figure}[htb]
	\centering
	\begin{tabular}{c} 
		\includegraphics[width=0.48\textwidth]{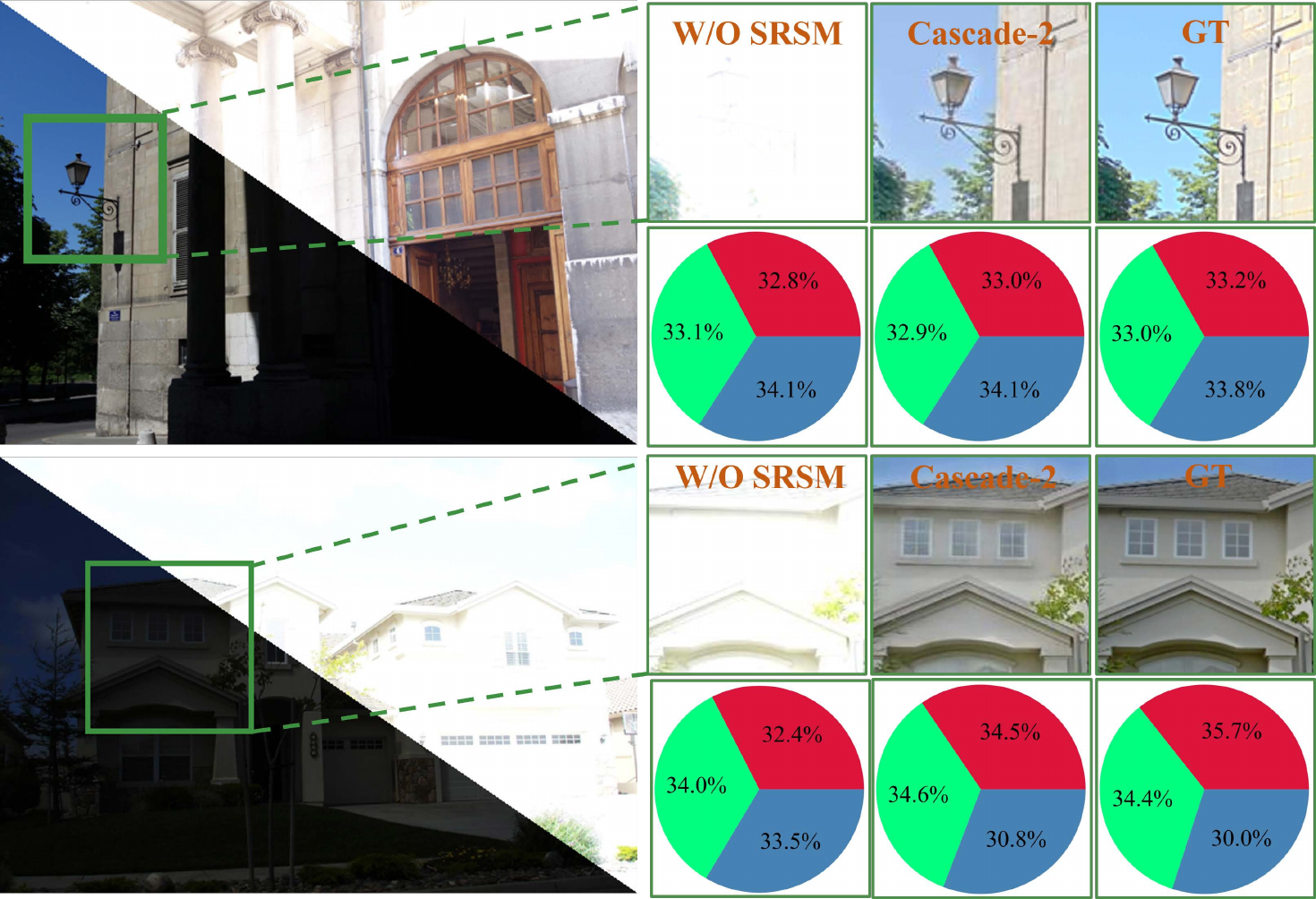} \\
		
	\end{tabular}
	
	\caption{Visual comparison of effectiveness for SRSM. }
	\label{fig:ablation_att}
	
\end{figure}

\section{Ablation  Study}
In this section, we conduct sufficient experiments with numerical and visual evaluations to verify the effectiveness of proposed modules, loss functions and architecture search.

\subsection{Effectiveness of Scene Relighting.} In this part, we first validate the effectiveness of proposed SRSM and validate the suitable cascaded numbers  for MEF task. The role of scene relighting is to gradually preserve the scene information and constrain the level of illumination for following feature aggregation. The ablation experiment about the cascaded number of SRSM is conducted, where the quantitative and qualitative comparisons are shown at Table.~\ref{tab:ablation_att} and Fig.~\ref{fig:ablation_att} respectively. Firstly, we illustrate the necessary of proposed SRSM. The version without SRSM only concatenate the inputs to feed into the DRM, without the procedure of illumination adjustment. We can clearly observe that, directly processing compromises the image quality, which reduce the numerical performance drastically. As shown in Fig.~\ref{fig:ablation_att}, we can obtain the output image is over-exposed, cannot render sufficient detail and preserve the normal light distribution. Compared by pie charts, which depicts  the proportion of RGB channels. The version w/o SRSM cannot restore the normal color distribution, which leads to the distortion of color and details. Then we evaluate the cascaded number of SRSM. By introducing the cascaded SRSMs, we propose the recurrent attention mechanisms to 
extract the sufficient features. Cascading two modules can achieve the best numerical performance. Increasing the number of SRSM obtain the moderate improvement.

\begin{table}[htb]
	\centering
	\renewcommand{\arraystretch}{1.3}
	\caption{Quantitative comparisons of effectiveness with SRSM.}
	\label{tab:ablation_att}
	\setlength{\tabcolsep}{1.8mm}{
		\begin{tabular}{|c| c |c| c| c|}
			\hline
			Number & w/o SRSM   & Cascade-1 & Cascade-2&Cascaded-3  \\\hline
			\cellcolor{gray!15}			PSNR $\uparrow$ & 9.57&20.15&\textbf{20.71}&{20.56}\\
			\hline
			\cellcolor{gray!15}			SSIM $\uparrow$&0.611&0.819&\textbf{0.825}&\textbf{0.825}\\\hline
			\cellcolor{gray!15}			LPIPS $\downarrow$&{0.407}&0.150&\textbf{0.132}&{0.136}\\\hline
			\cellcolor{gray!15}	 		FSIM $\uparrow$ &{0.787}&\textbf{0.932}&0.924&{0.929}\\
			%			LPIPS& & &0.434 &\\
			
			\hline
		\end{tabular}	
	}
	
\end{table}

\subsection{Effectiveness of Deformable Alignment}
\begin{figure}[htb]
	\centering
	\begin{tabular}{c} 
		\includegraphics[width=0.48\textwidth]{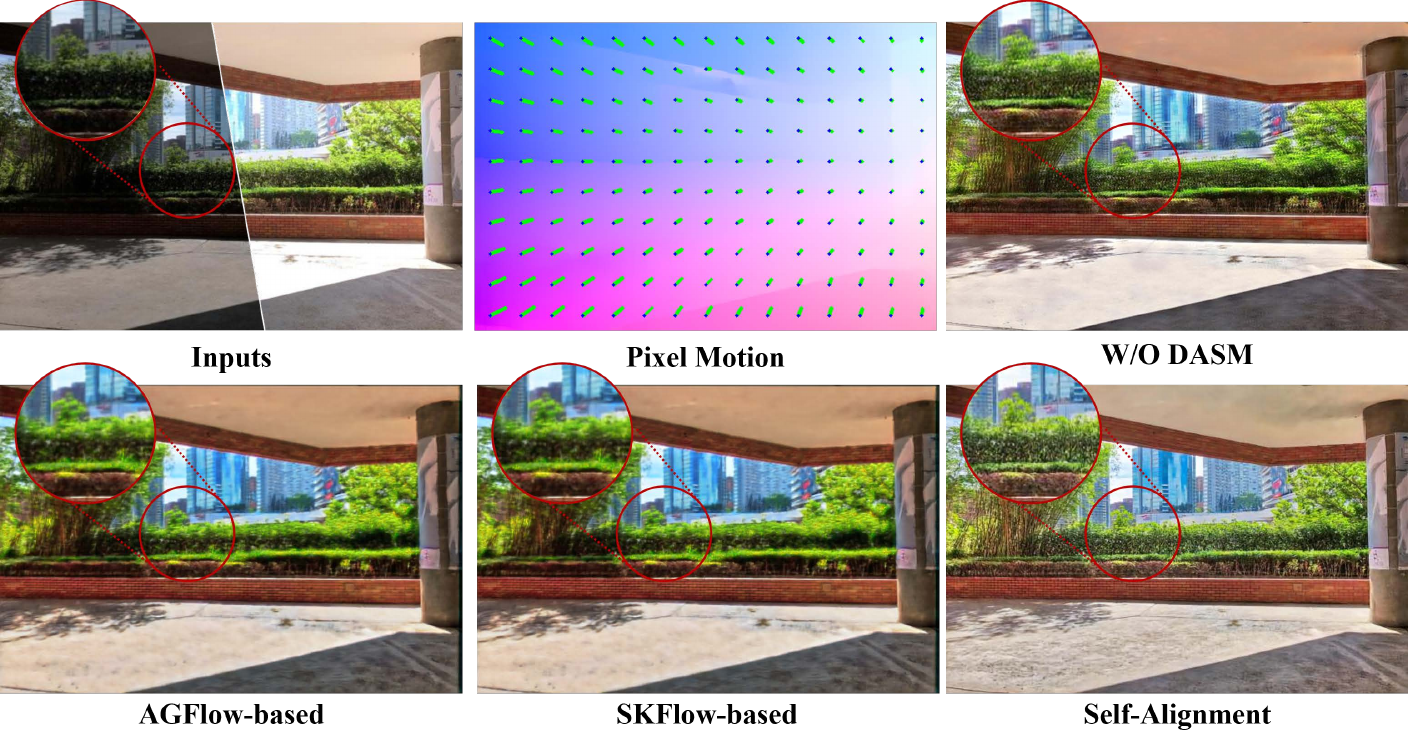} \\
		
	\end{tabular}
	
	\caption{\textcolor{black}{Visual comparison with diverse optical flow alignment strategies.}}
	\label{fig:ablation_align}
	
\end{figure}

We further evaluate the advantages of self-alignment mechanism. \textcolor{black}{Six variants of comparison are conducted,  including ``w/o DASM'', flow-based alignment (such as AGFlow, SKFlow, and RAFT), and changing the position before relighting. Self-alignment module targets to align the unregistered pixels of image pairs. In the architecture construction, we put DASM after the SRSM. Optical flow-based schemes are to utilize diverse alignment techniques to align image pairs. Numerical results and visual comparisons are depicted in Table.~\ref{tab:ablation_alignment} and Fig.~\ref{fig:ablation_align} respectively. From the numerical results, we can observe the effectiveness of the proposed mechanisms. Compared with RAFT-based methods, our scheme improves 14.9\% of PSNR, 26.1\% of SSIM, and 43.5\% of LPIPS. Our method effectively solves the pixel alignment under moderate offsets.  Designing the MEF-oriented optical flow method is a potential direction to address the artifacts caused by the large motion.}

\textcolor{black}{  From the visual comparison, the fused result of ``w/o DASM'' cannot preserve enough texture details, such as the grasses. On the other hand, we can see that the optical flow-based schemes cannot improve the quantitative results due to the inaccurate motion estimation caused by different illumination. The fused images contain more obvious artifacts. Our method effectively solves the pixel alignment under moderate offsets.  Designing the MEF-oriented flow method is a potential direction to address the artifacts with the large motion.}

\begin{table}[htb]
	\centering
	\renewcommand{\arraystretch}{1.3}
	\caption{\textcolor{black}{Quantitative comparisons of effectiveness with DASM.}}
	\label{tab:ablation_alignment}
	\setlength{\tabcolsep}{1.5mm}{
		\begin{tabular}{|c| c|c| c |c| c| c|}
			\hline
			Metric & Model-1   & Model-2 &Model-3&Model-4&Model-5 & Ours\\\hline
			\cellcolor{gray!15}		PSNR $\uparrow$ & 19.39&18.68&18.61&19.19&21.73&\textbf{22.04}\\
			\hline
			\cellcolor{gray!15}		SSIM $\uparrow$&0.561&0.544&0.544&0.540&0.675&\textbf{0.681}\\\hline
			\cellcolor{gray!15}		LPIPS $\downarrow$ &{0.325}&0.350&0.350&0.331&0.237&\textbf{0.187}\\\hline
			%			VIF&{0.692}&0.657&0.840&\textbf{0.855}\\\hline
			\cellcolor{gray!15}	 	FSIM  $\uparrow$ &{0.852}&0.837&0.836&0.838&0.907&\textbf{0.913}\\\hline	
		\end{tabular}	
	}
	
\end{table}
%\subsection{Validation of Detail Repletion}
%\begin{figure}[!htb]
%	\centering
%	\setlength{\tabcolsep}{1pt} 
%	\begin{tabular}{c}		
	%		\includegraphics[width=0.48\textwidth]{pics/DRM.pdf}\\
	%		
	%		%				\includegraphics[width=0.98\textwidth]{pics/244.pdf}\\
	%	\end{tabular}
%
%	\caption{\textcolor{blue}{Effectiveness of DRM compared by the visual effects and the difference with ground truth.}}
%	\label{fig:DRM}
%\end{figure}
\subsection{Training Losses Analyses}  
\begin{figure}[htb]
	\centering
	\begin{tabular}{c@{\extracolsep{0.3em}}c@{\extracolsep{0.3em}}c@{\extracolsep{0.3em}}c} 
		\includegraphics[width=0.109\textwidth]{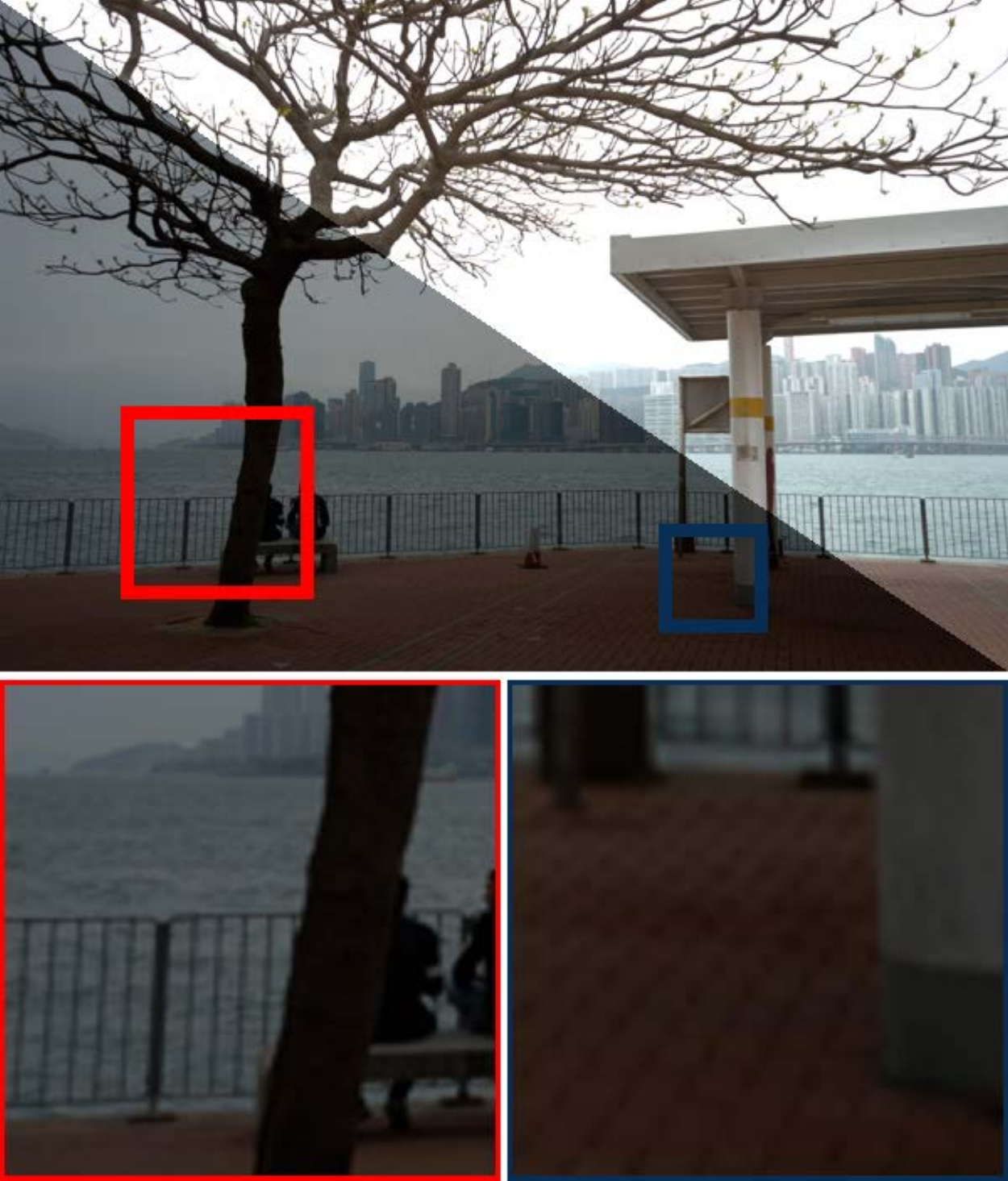}&
		\includegraphics[width=0.109\textwidth]{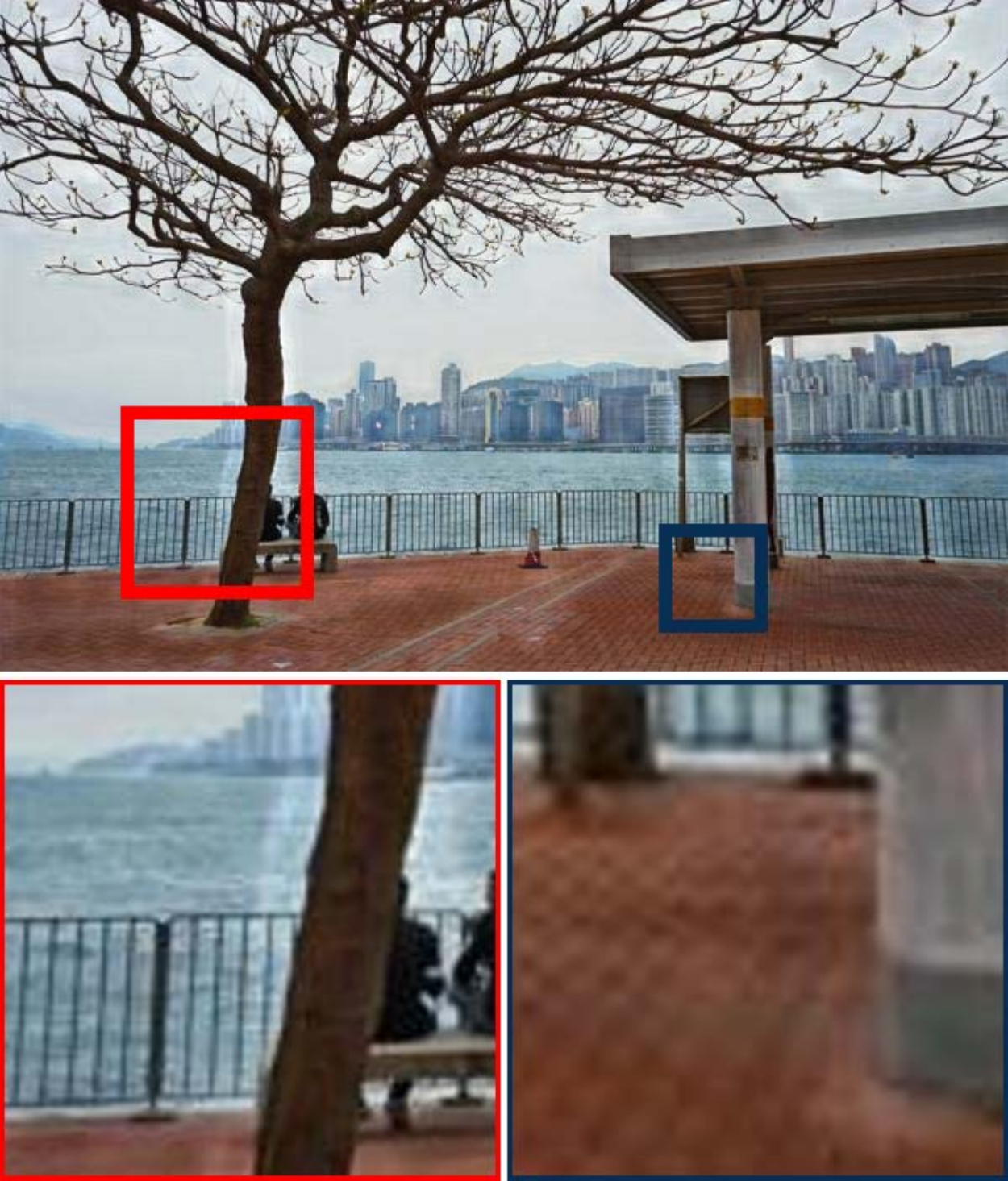}&
		\includegraphics[width=0.109\textwidth]{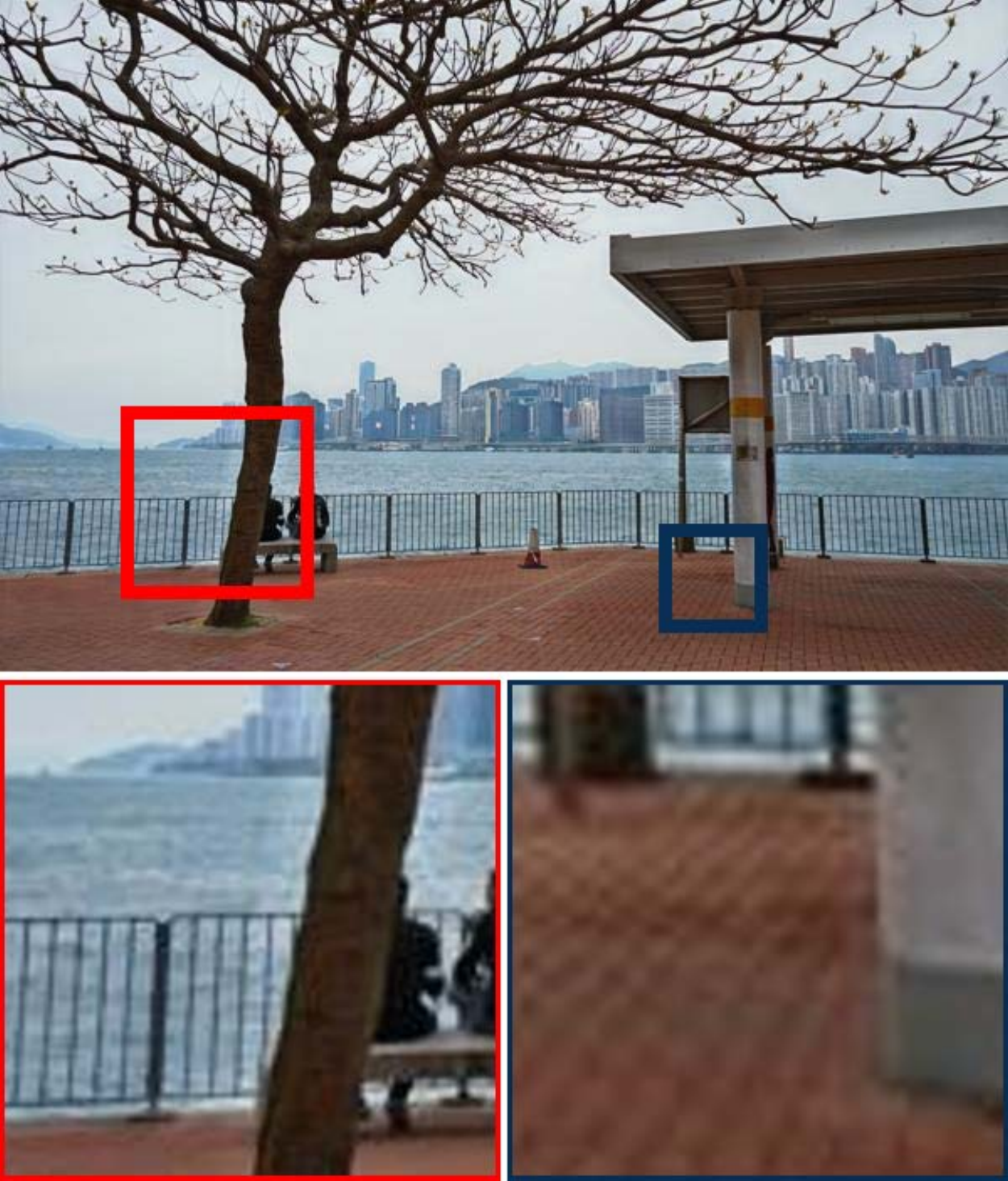}&
		\includegraphics[width=0.109\textwidth]{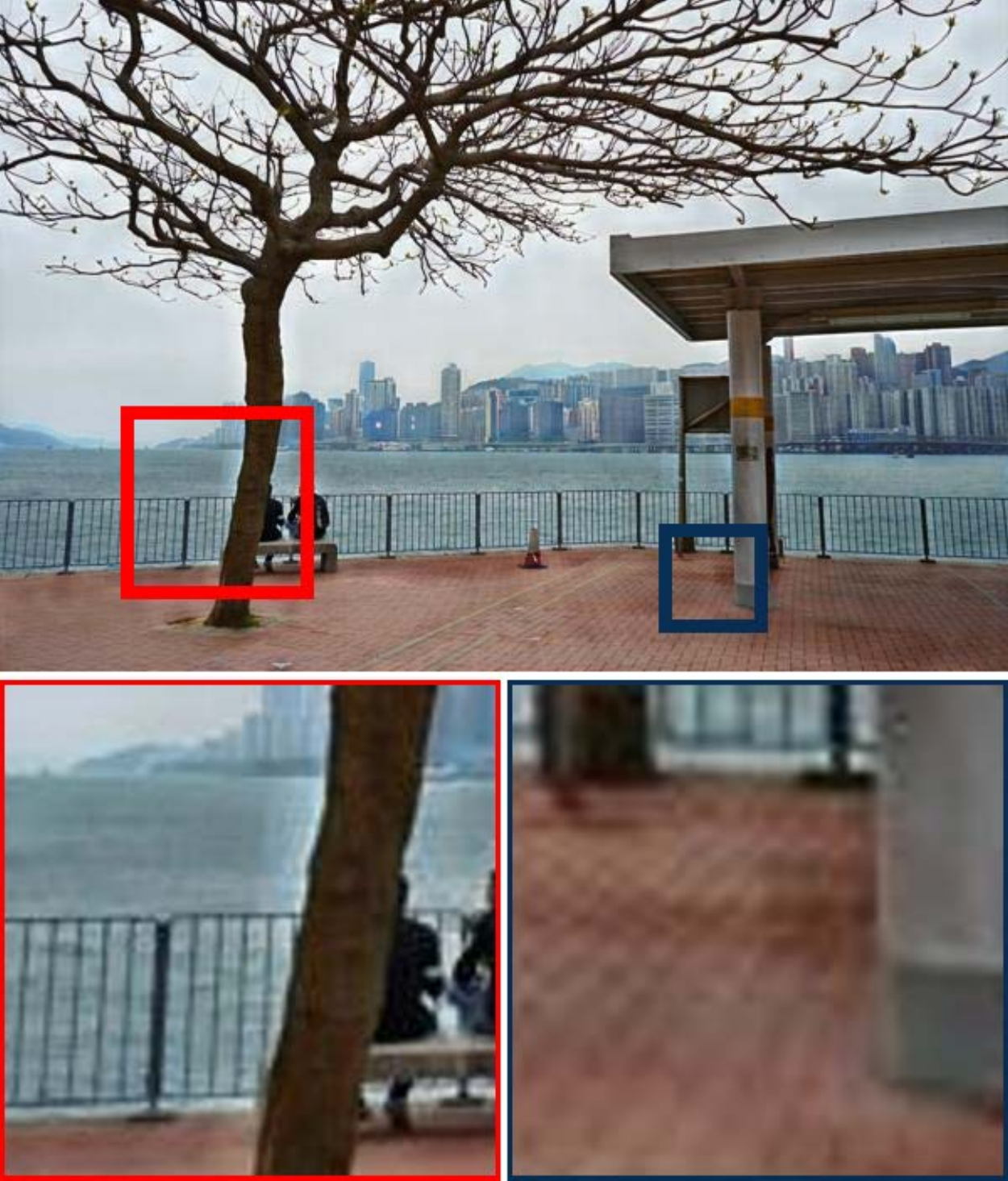}\\
		\includegraphics[width=0.109\textwidth]{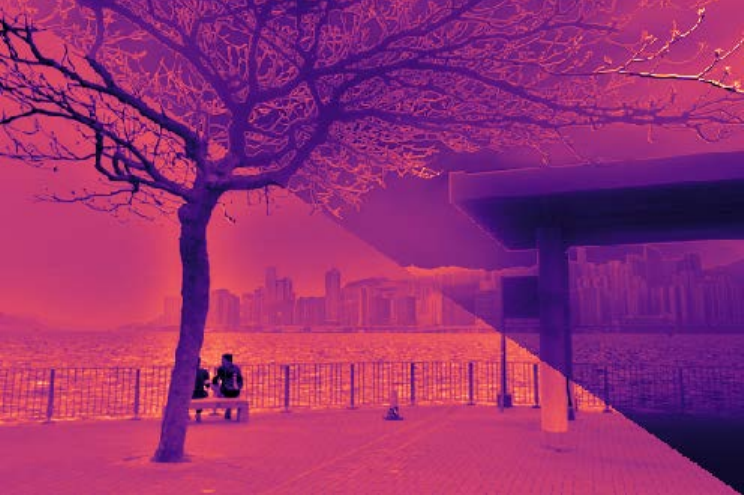}&
		\includegraphics[width=0.109\textwidth]{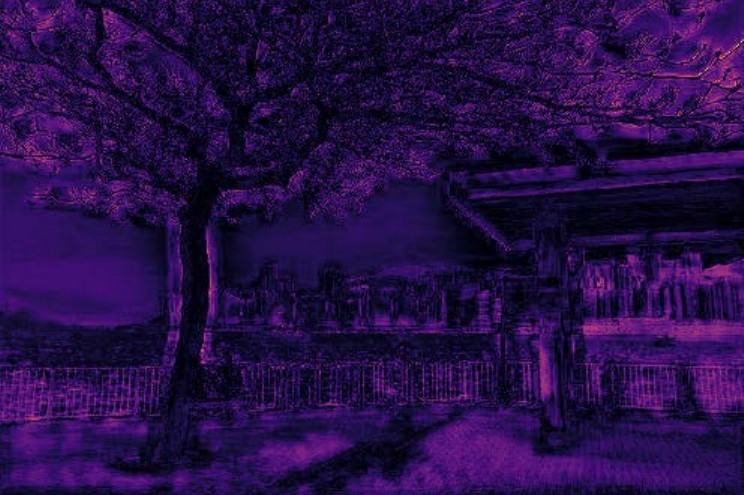}&
		\includegraphics[width=0.109\textwidth]{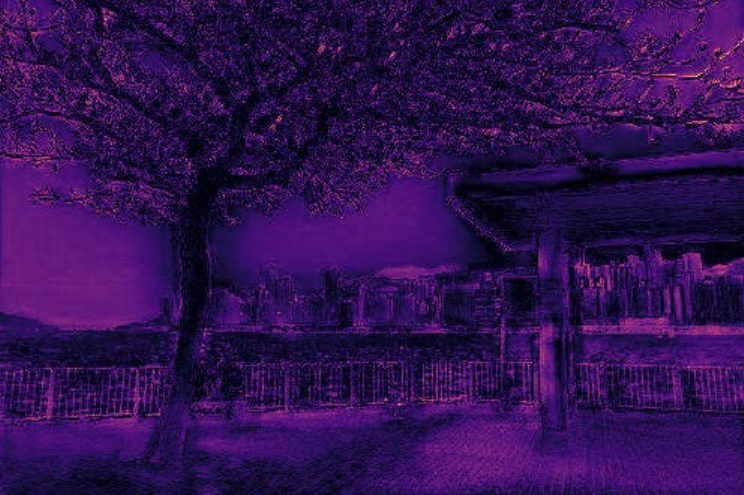}&
		\includegraphics[width=0.109\textwidth]{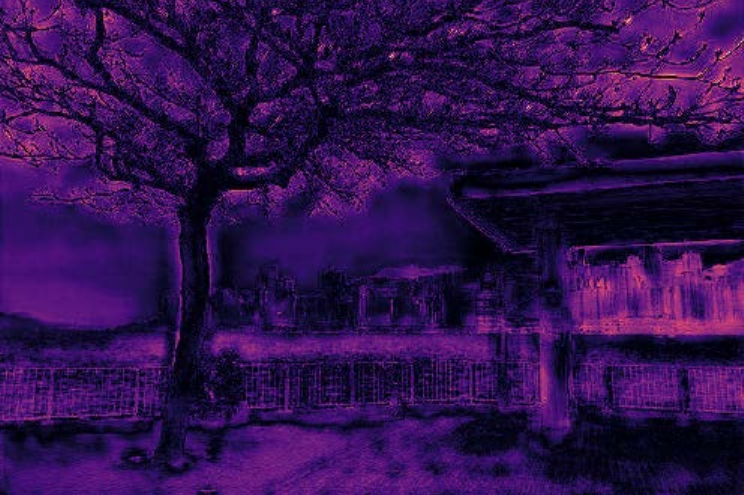}\\
		\includegraphics[width=0.109\textwidth]{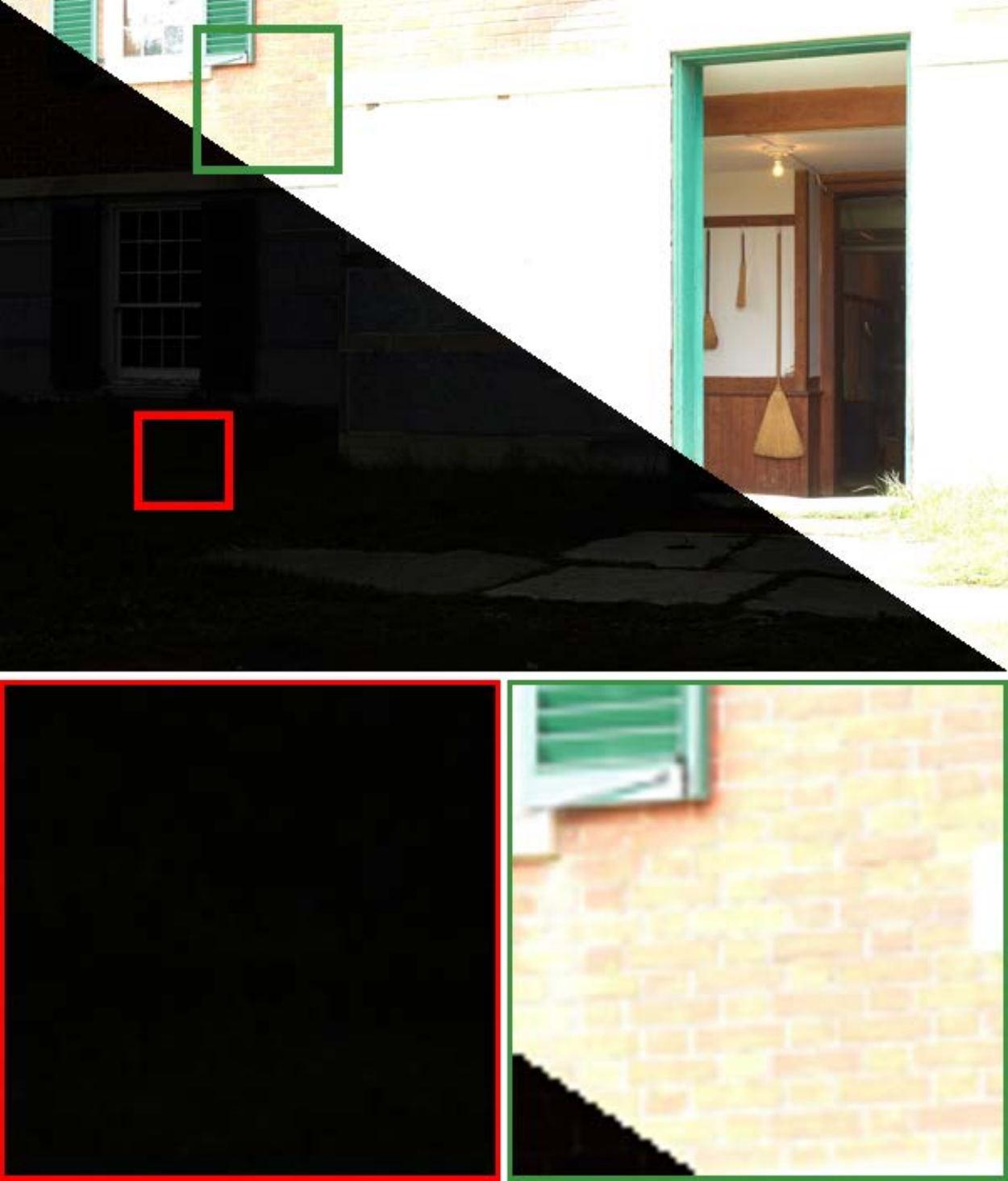}&
		\includegraphics[width=0.109\textwidth]{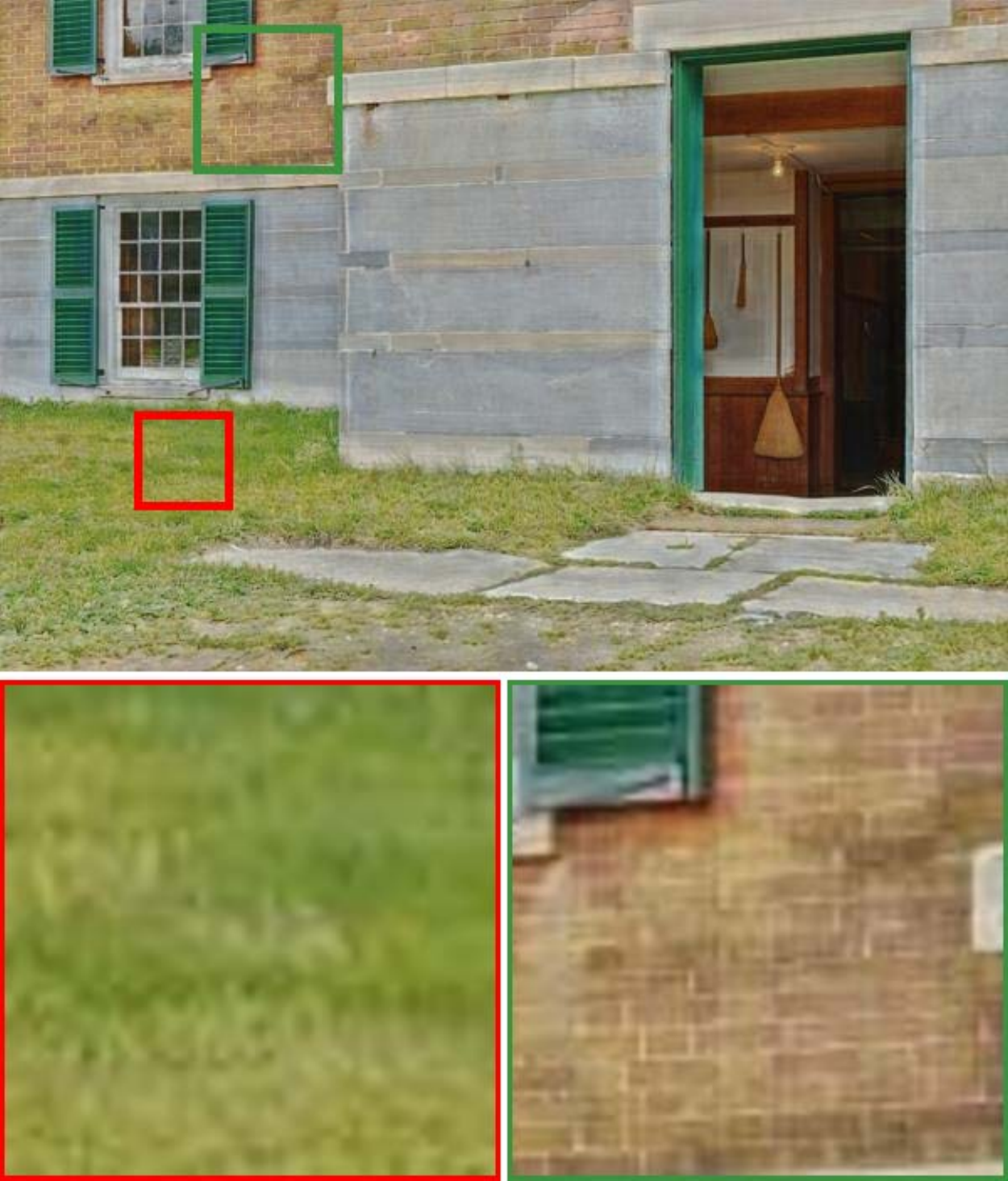}&
		\includegraphics[width=0.109\textwidth]{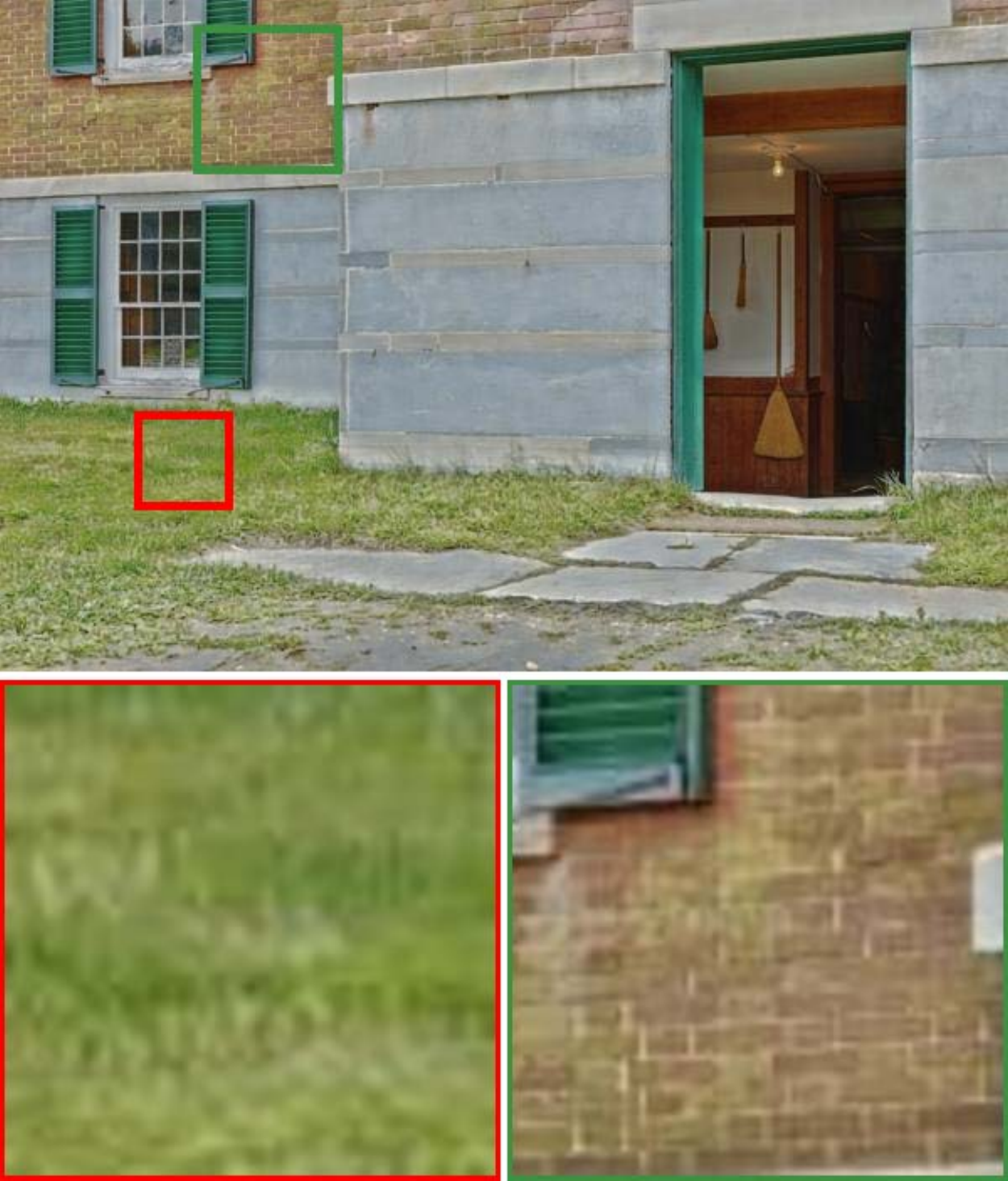}&
		\includegraphics[width=0.109\textwidth]{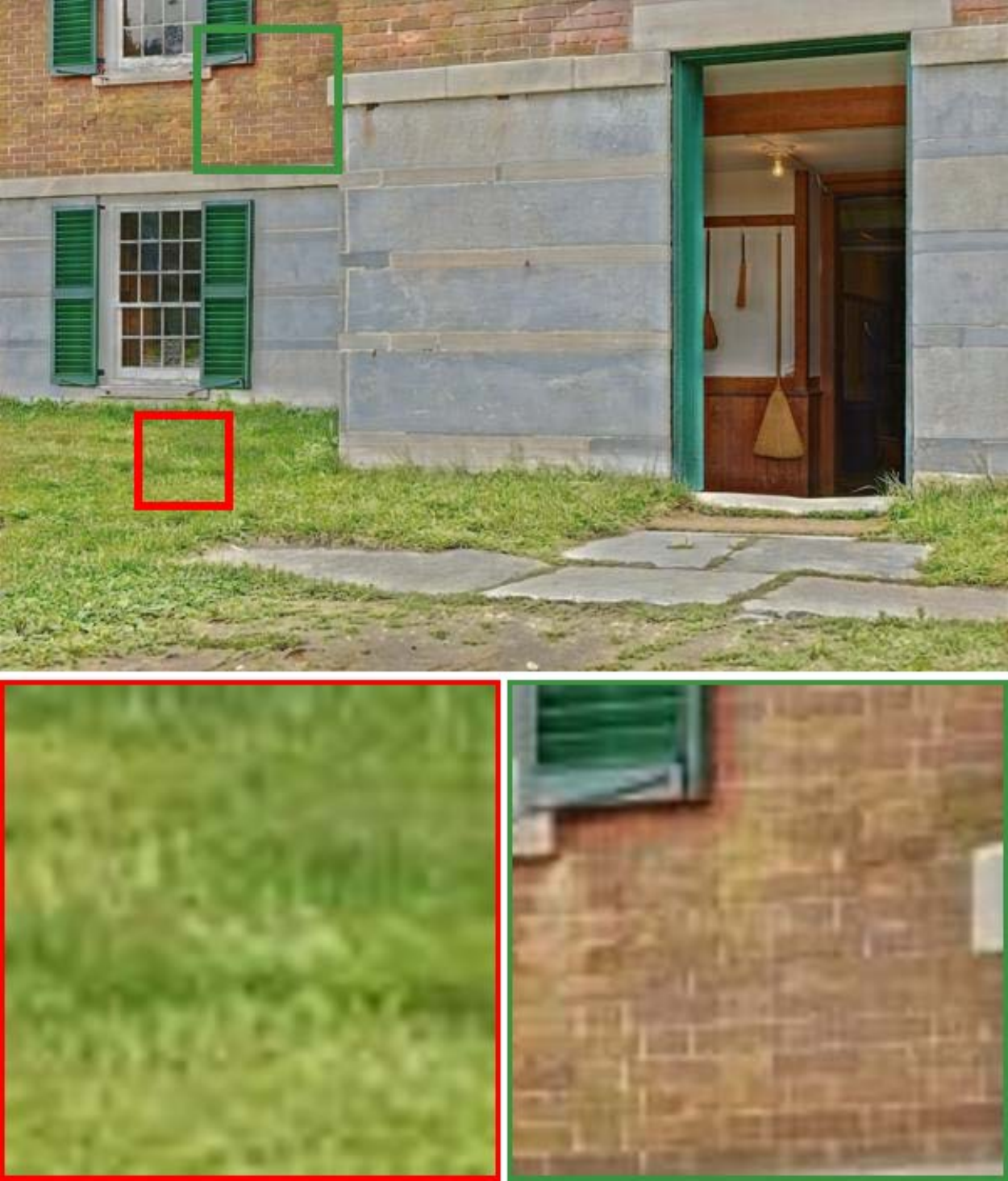}\\
		\includegraphics[width=0.109\textwidth]{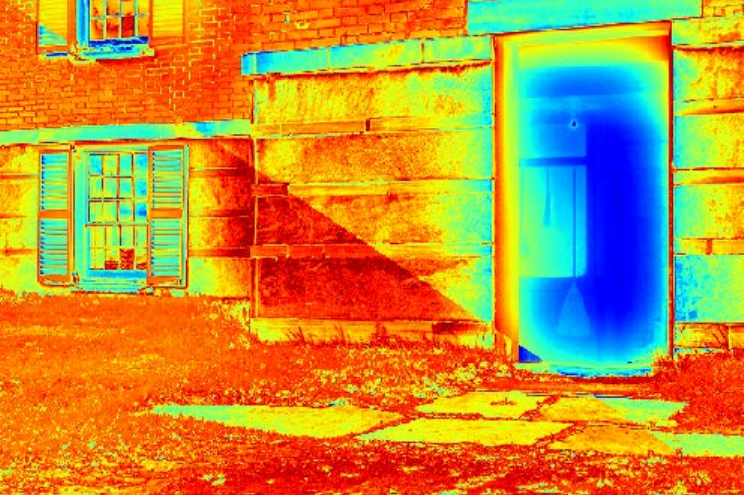}&
		\includegraphics[width=0.109\textwidth]{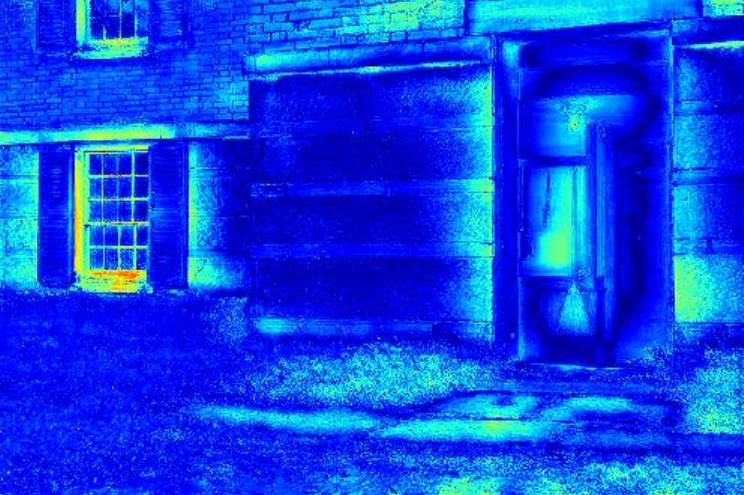}&
		\includegraphics[width=0.109\textwidth]{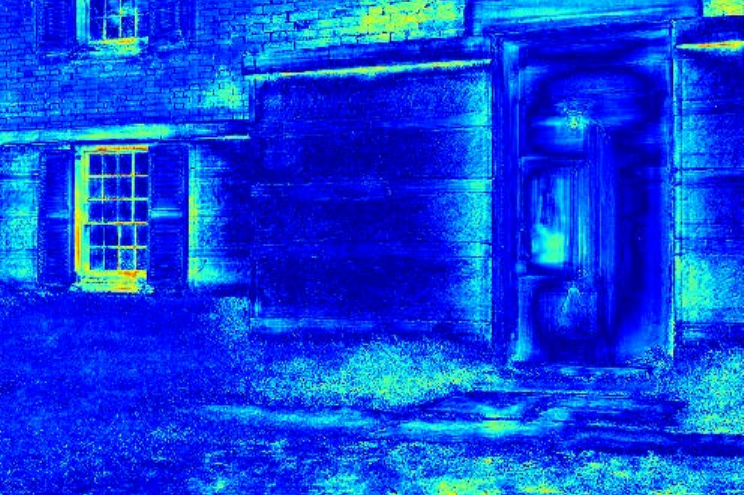}&
		\includegraphics[width=0.109\textwidth]{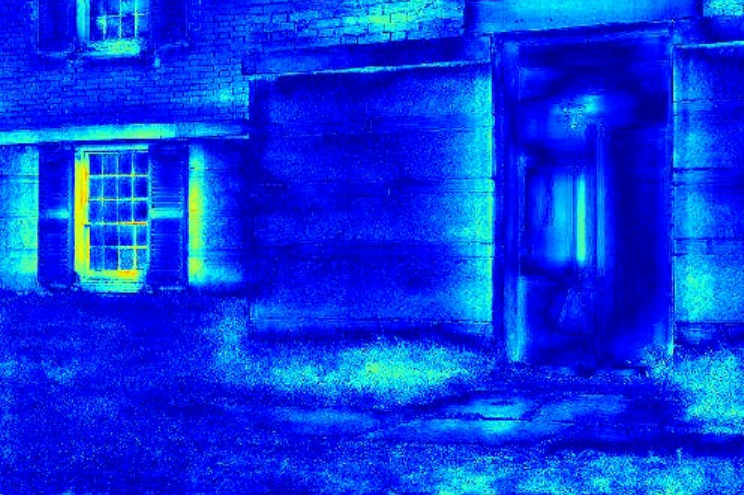}\\
		\footnotesize	Inputs & \footnotesize $\ell_\mathtt{Int}$ & \footnotesize $\ell_\mathtt{Int}$+$\ell_\mathtt{Gra}$& \footnotesize $\ell_\mathtt{Total}$\\
	\end{tabular}
	
	\caption{Visual results and error maps obtained by the different  loss functions. }
	\label{fig:ablation_loss}
	
\end{figure}

We also  perform the detailed analyses to evaluate the effectiveness of diverse training strategies (\textit{i.e.,} combinations of loss functions). In this part, we gradually introduce the loss functions to composited three schemes, including ``w/ $\ell_\mathtt{Int}$'', ``w/ $\ell_\mathtt{Int}$ + $\ell_\mathtt{Gra}$'' and our scheme. Related visual results are plotted at Fig.~\ref{fig:ablation_loss}. From the objective comparison,  $\ell_\mathtt{Gra}$ can effectively preserve the edge information, which reflects on the structural measurement SSIM and feature-level metric LPIPS. As shown in Fig.~\ref{fig:ablation_loss}, visual results scheme  `` w/ $\ell_\mathtt{Int}$+$\ell_\mathtt{Gra}$'' provide flourishing textural details, \textit{e.g.,} the details of grasses and floors. Meanwhile, introducing $\ell_\mathtt{Gra}$ can effectively remove the artifact such as the shape of tree on the first row. Our scheme is combined with three categories of losses, \textit{i.e.,} $\ell_\mathtt{Int}$ for pixel intensity, $\ell_\mathtt{Gra}$ for structural detail and $\ell_\mathtt{Dis}$ for color distribution.
Thus, our scheme can further improve the visual quality, obtaining with the highest numerical results. For instance, our final scheme obtain the vivid color distribution, without any color distortion (\textit{e.g.,} the color of wall), shown at the second row in Fig.~\ref{fig:ablation_loss}.
\begin{table}[htb]
	\centering
	\renewcommand{\arraystretch}{1.3}
	\caption{Analysing the effectiveness of loss functions.}
	\label{tab:ablation_loss}
	\setlength{\tabcolsep}{2.5mm}{
		\begin{tabular}{|c| c | c|c|c|}
			\hline
			Metric & w/ $\ell_\mathtt{Int}$   & w/ $\ell_\mathtt{Int}$+$\ell_\mathtt{Gra}$ & w/ $\ell_\mathtt{Int}$+$\ell_\mathtt{Dis}$ &  $\ell_\mathtt{Total}$\\\hline
			\cellcolor{gray!15}	PSNR $\uparrow$  & 20.62&20.63&20.66&\textbf{20.71}\\
			\hline
			\cellcolor{gray!15}	SSIM $\uparrow$ &0.799&0.823&0.821&\textbf{0.825}\\\hline
			\cellcolor{gray!15}	LPIPS $\downarrow$ &{0.145}&0.134&0.136&\textbf{0.132}\\\hline
			\cellcolor{gray!15}	FSIM $\uparrow$ &\textbf{0.932}&{0.913}&0.927&{0.924}\\
			%			LPIPS& & &0.434 &\\
			
			\hline
		\end{tabular}	
	}
	
\end{table}
\begin{table}[thb]
	\centering
	\renewcommand{\arraystretch}{1.3}
	\caption{Quantitative comparision of search space.}
	\label{tab:ablation_searchspace}
	\setlength{\tabcolsep}{1.3mm}{
		\begin{tabular}{|c| c |c| c| c|c|}
			\hline
			\cellcolor{gray!15} 	Operator & \cellcolor{gray!15} PSNR  $\uparrow$  & \cellcolor{gray!15} SSIM  $\uparrow$ & \cellcolor{gray!15} LPIPS  $\downarrow$ & \cellcolor{gray!15} FSIM   $\uparrow$&  \cellcolor{gray!15} Runtime (s)  \\\hline
			%			Super-Net& 28.076&22.833&28.614&\textbf{29.122}\\
			%			\hline
			3-C & 20.66&0.824&\textcolor{blue}{\textbf{0.132}}&0.926&\textcolor{red}{\textbf{0.042}}\\
			\hline
			3-DC&20.68&\textcolor{red}{\textbf{0.825}}&\textcolor{red}{\textbf{0.131}}&\textcolor{blue}{\textbf{0.931}}&\textcolor{blue}{\textbf{0.044}}\\\hline
			5-C&20.58&0.822&0.134&0.930&0.124\\\hline
			5-DC&20.49&0.818&0.135&0.930&0.088\\
			\hline
			7-C &20.52&0.820&0.135&0.928&0.195\\
			\hline
			7-DC&\textcolor{red}{\textbf{20.80}}&0.823&0.140&\textcolor{red}{\textbf{0.933}}&0.153\\\hline
			Ours&\textcolor{blue}{\textbf{20.71}}&\textcolor{red}{\textbf{0.825}}&\textcolor{blue}{\textbf{0.132}}&{0.924}&0.047\\\hline
		\end{tabular}	
	}
	
\end{table}
\subsection{Search Space Analyses} We also verify the basic properties of search space, where the concrete performances are reported at Table.~\ref{tab:ablation_searchspace}. From the fusion performance, we can observe that dilated convolutions (3-DC and 7-DC)  have higher quantitative results (\textit{e.g.,} PSNR, LPIPS, and FSIM) compared with normal convolutions. On the other hand, we can directly observe that $3\times3$ convolution has a fast inference speed but has sub-optimal statistical results. Under the constraint of hardware latency, our search scheme actually achieves the balance between visual quality and inference speed.

\subsection{Hardware-sensitive Analyses} We also analyze the influence of trade-off parameter $\eta$, which controls the influences of hardware-sensitive latency constraint. The numerical results are reported in Table.~\ref{tab:hardware2}. ``C32'' and ``C64'' denote the version with 32 and 64 channels. When $\eta = 0.5$, the balance between fusion quality and inference requirement can be guaranteed simultaneously.
The concrete architectures under diverse $\eta$ are plotted in Fig.~\ref{fig:cacasede}. The previous three layers illustrate the architecture of SAM. The last four layers show the structure of DRM with residual connection. In detail, without the constraint of latency, the NAS scheme chooses an operator with a large receptive field to better capture the large features. Moreover, we also can conclude that $3\times 3$ convolution can effectively extract features with high efficiency, which is widely leveraged for SRSM under the latency constraint. As for DRM, \{3-C, 1-C\} with skip connection is a low-weight combination for the detail compensation, as shown in the subfigure (c) and (d).
\begin{table}[!htb]
	\centering
	\caption{Trade-off $\eta$ for hardware-sensitive analysis.}
	\renewcommand\arraystretch{1.3} \footnotesize
	\setlength{\tabcolsep}{1.5mm}
	\begin{tabular}{|c|cc|cc|cc|}
		\hline
		\multirow{2}{*}{Trade-off $\eta$} & \multicolumn{2}{c|}{ $\eta = 0$}    & \multicolumn{2}{c|}{ $\eta = 0.5$}     & \multicolumn{2}{c|}{ $\eta = 1\&=5$}    \\ \cline{2-7} 
		& \multicolumn{1}{c|}{C32} &  C64 & \multicolumn{1}{c|}{C32} &  C64  & \multicolumn{1}{c|}{C32} & C64\\ \hline
		\cellcolor{gray!15}		PSNR $\uparrow$  & \multicolumn{1}{c|}{20.65} &20.61  & \multicolumn{1}{c|}{20.60} & 20.71 & \multicolumn{1}{c|}{20.54} & 20.56 \\ \hline
		\cellcolor{gray!15}		Parameters (M)   $\downarrow$           & \multicolumn{1}{c|}{2.716} &4.631  & \multicolumn{1}{c|}{0.701} &1.879    & \multicolumn{1}{c|}{0.684} &1.617  \\ \hline
		\cellcolor{gray!15}		FLOPs (G) $\downarrow$  & \multicolumn{1}{c|}{215.3} &  675.0& \multicolumn{1}{c|}{89.49} &368.3  & \multicolumn{1}{c|}{81.56} &305.4  \\ \hline
		\cellcolor{gray!15}		Runtime (s)  $\downarrow$ & \multicolumn{1}{c|}{0.064} & 0.085 & \multicolumn{1}{c|}{0.017} & 0.047  & \multicolumn{1}{c|}{0.021} &  0.039 \\ \hline
		
	\end{tabular}
	\label{tab:hardware2}	
\end{table}
\begin{figure}[!htb]
	\centering
	\setlength{\tabcolsep}{1pt}
	\begin{tabular}{cc}
		
		\includegraphics[width=0.23\textwidth]{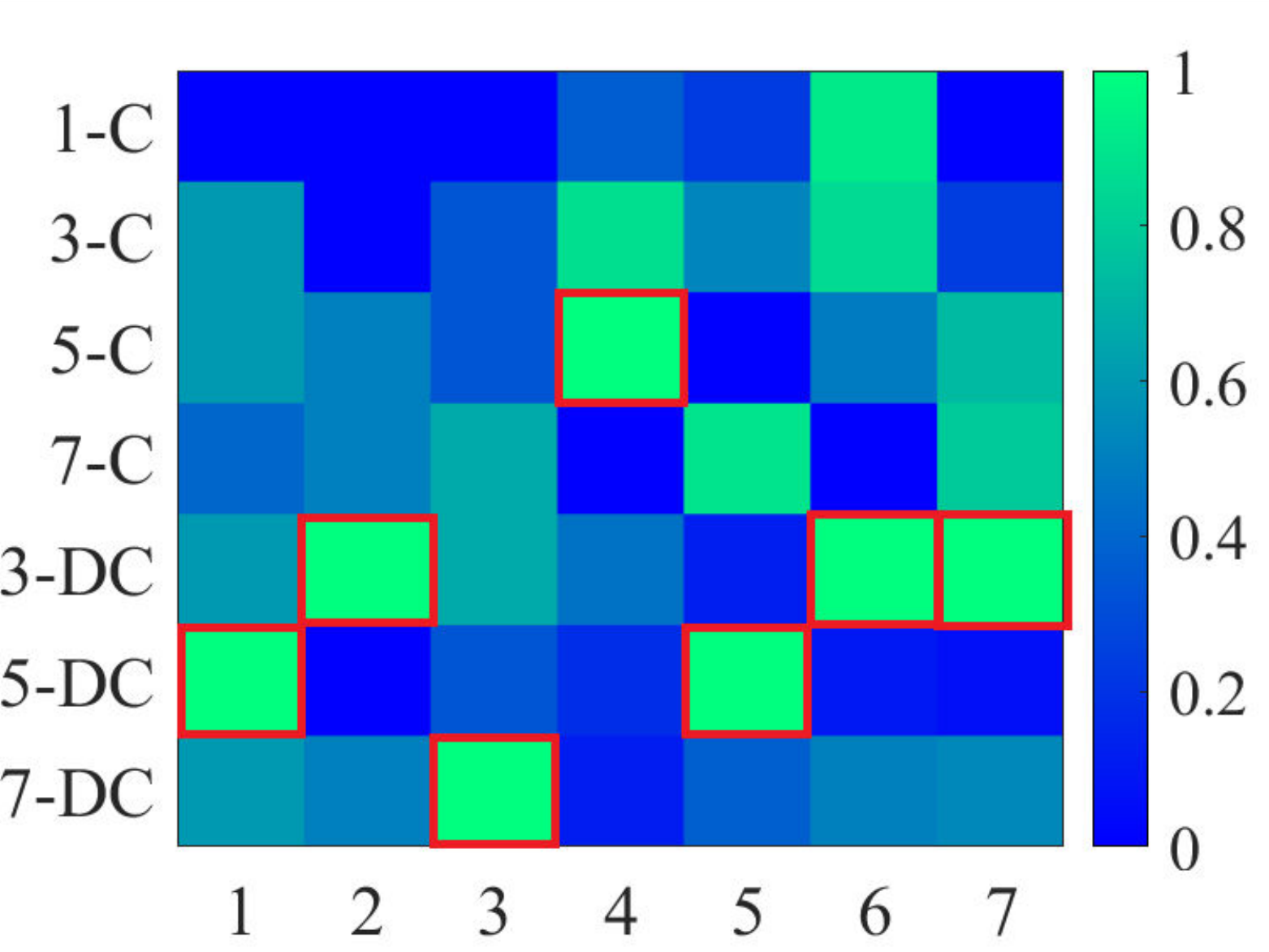}
		&\includegraphics[width=0.23\textwidth]{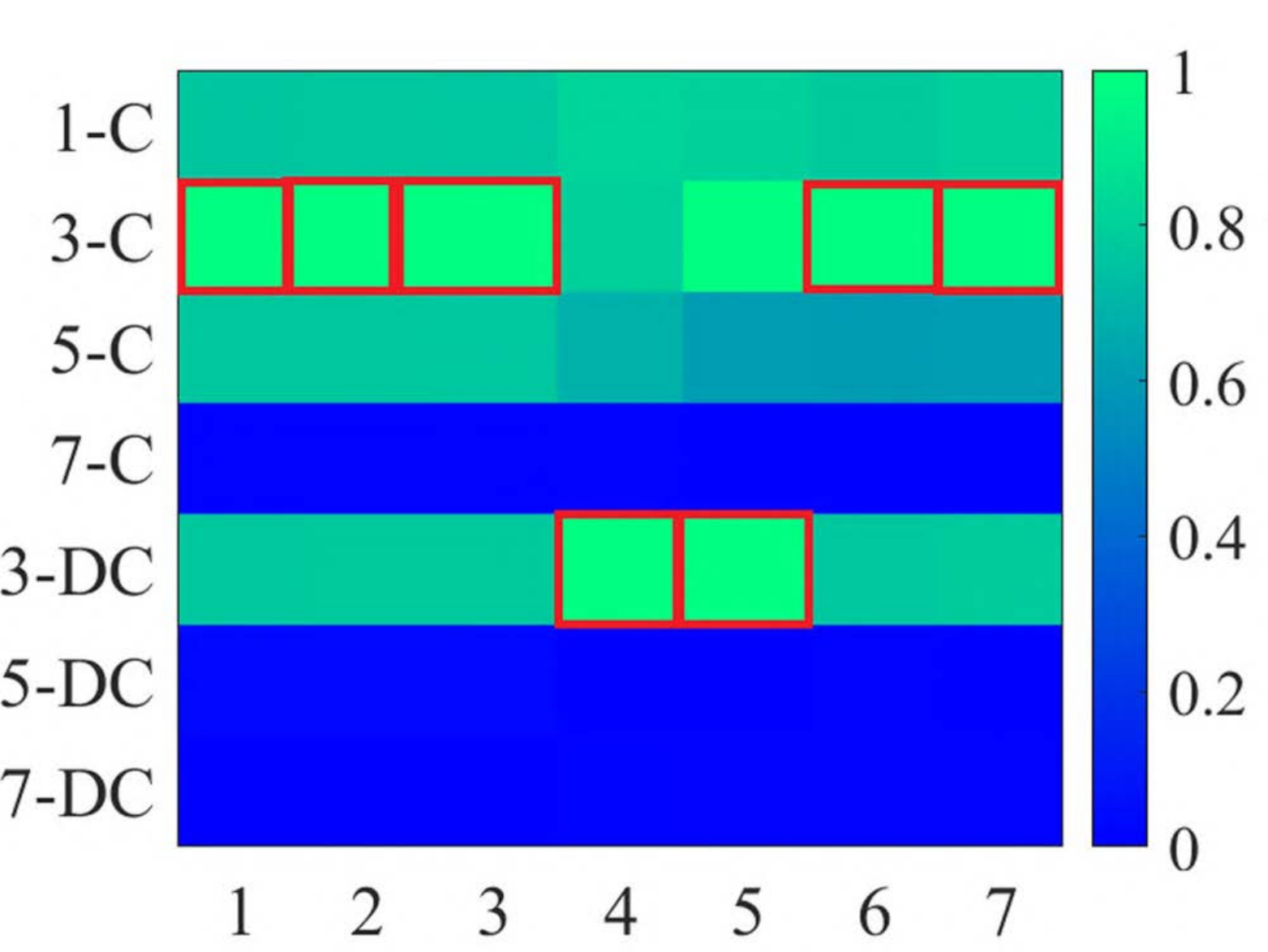}	
		\\
		\footnotesize (a) $\eta = 0$ &\footnotesize (b)	$\eta = 0.5$			\\	
		\includegraphics[width=0.23\textwidth]{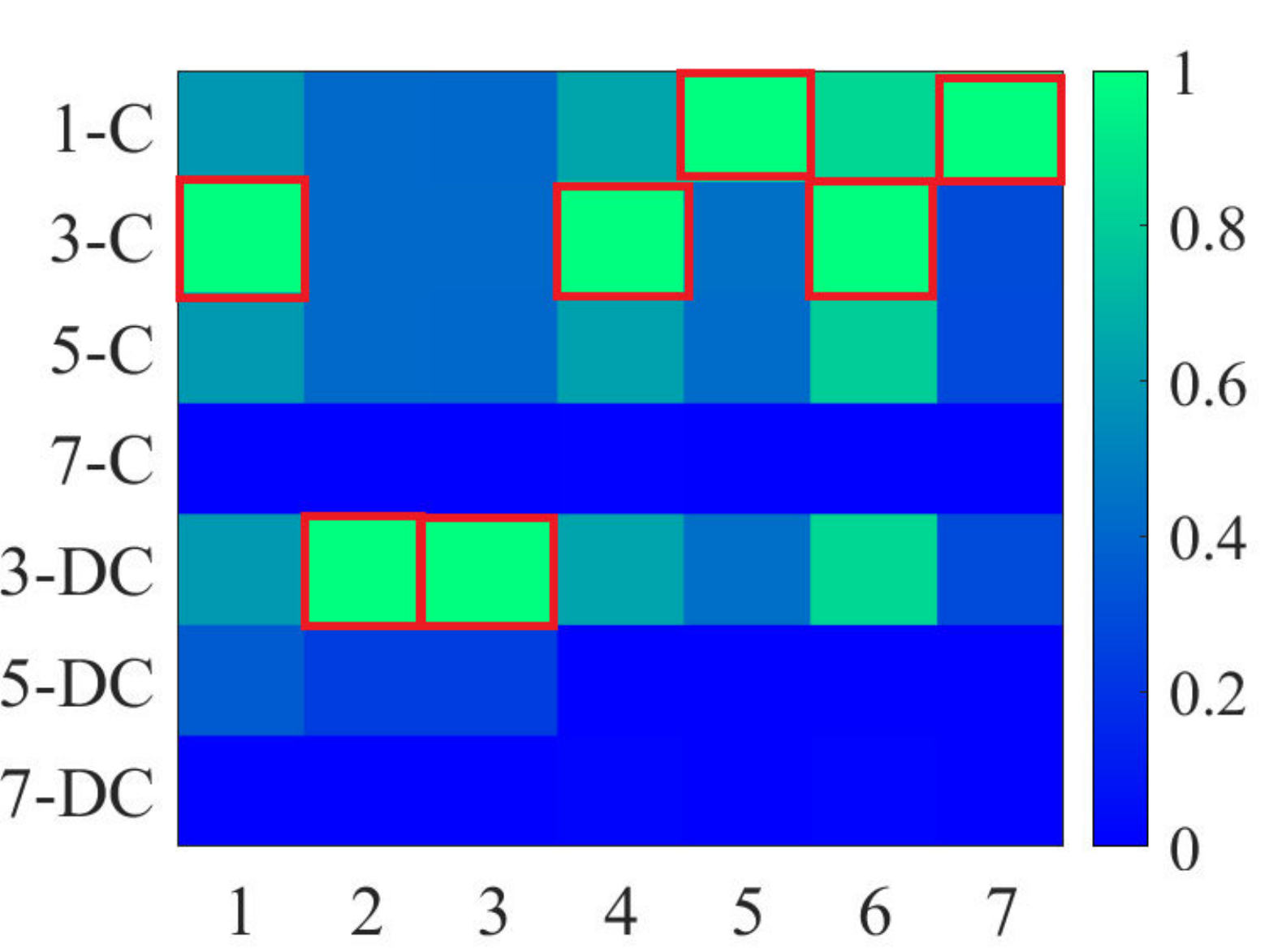}
		&\includegraphics[width=0.23\textwidth]{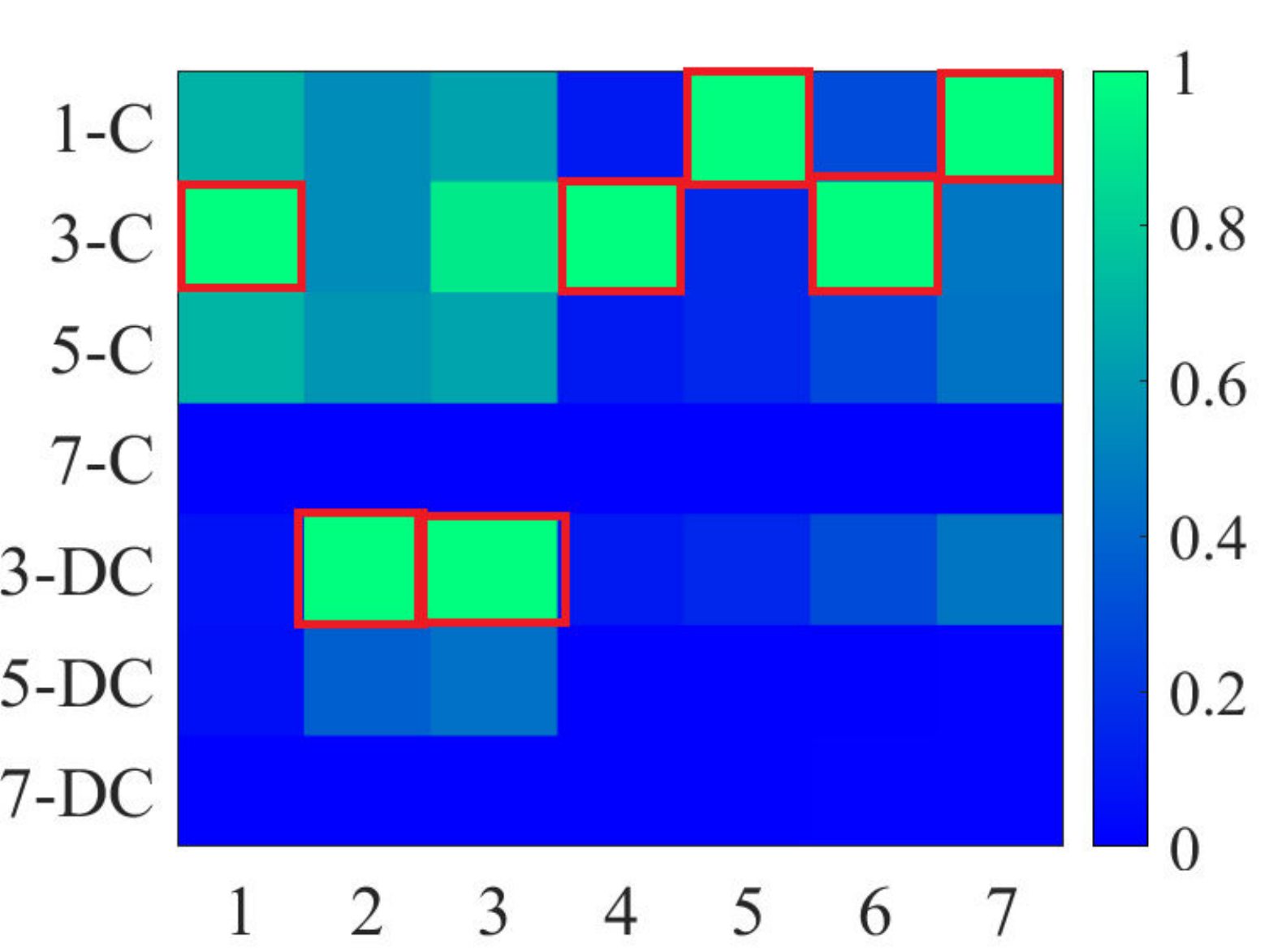}	
		\\
		\footnotesize 	(c) $\eta = 1$ & \footnotesize (d)	$\eta = 5$		
	\end{tabular}
	
	\caption{Heatmaps of the searched architectures based on different trade-off parameter $\eta$. The selected operators are marked by red boxes.}
	\label{fig:cacasede}
\end{figure}

\section{Concluding Remarks}
In this paper, we proposed a robust multi-exposure image fusion framework to address various 
scenarios, including the aligned and misaligned image pairs. We divided the fusion procedure into two parts: self-alignment for feature-wise alignment and detail repletion to enhance texture details visually. By utilizing a hardware-friendly architecture search strategy and incorporating a task-oriented search space, we  discovered a highly efficient and compact architecture for MEF. Furthermore, we conducted comprehensive subjective and objective comparisons to demonstrate the outstanding performance of our method compared to various state-of-the-arts.

\bibliographystyle{IEEEtran}
\bibliography{reference}

\end{document}